%% file: main.tex
\documentclass[11pt]{article}

\usepackage[preprint]{acl}

\usepackage{times}
\usepackage{latexsym}
\usepackage{amsmath}
\usepackage{amssymb}
\usepackage{amsfonts}

\usepackage[T1]{fontenc}
\usepackage[utf8]{inputenc}

\usepackage{microtype}

\usepackage{inconsolata}

\usepackage{graphicx}

\usepackage{xcolor}
\usepackage{listings}
\usepackage[capitalise]{cleveref}
\usepackage{booktabs}
\usepackage{multirow}
\usepackage{enumitem}
\usepackage{adjustbox}
\usepackage{placeins}
\usepackage{algorithm}
\usepackage{algpseudocode}

\title{Ranking Reasoning LLMs under Test-Time Scaling}

\author{
\textbf{Mohsen Hariri\textsuperscript{1}},
 \textbf{Michael Hinczewski\textsuperscript{2}},
 \textbf{Jing Ma\textsuperscript{1}},
 \textbf{Vipin Chaudhary\textsuperscript{1}},
\\
  \textsuperscript{1}Department of Computer and Data Sciences,
  \textsuperscript{2}Department of Physics \\
Case Western Reserve University, Cleveland, OH, USA
\\
  \small{
   \href{mailto:mohsen.hariri@case.edu}{mohsen.hariri@case.edu}
 }
}

\usepackage{xcolor}

\newif\ifcomments
\commentsfalse

\newif\ifdraft
\draftfalse

\newcommand{\dc}[1][]{%
  \ifdraft
    \PackageWarning{draftrefs}{Missing reference on page \thepage}%
    \textcolor{red}{\textbf{[REF\ifx\relax#1\relax\else: #1\fi]}}%
  \else
    \cite{#1}%
  \fi
}

\usepackage{comment}

\newcommand{\icon}[1]{\adjustbox{valign=m}{\includegraphics[width=2ex,height=2ex,keepaspectratio]{#1}}}

\newcommand{\skyicon}{\icon{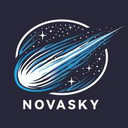}}
\newcommand{\qwenicon}{\icon{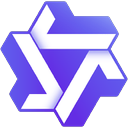}}
\newcommand{\dsicon}{\icon{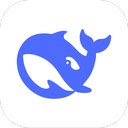}}
\newcommand{\gpticon}{\icon{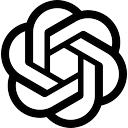}}
\newcommand{\gairicon}{\icon{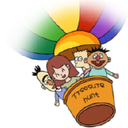}}
\newcommand{\lgicon}{\icon{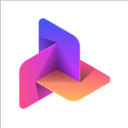}}
\newcommand{\nvidiaicon}{\icon{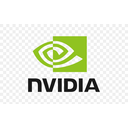}}
\newcommand{\openicon}{\icon{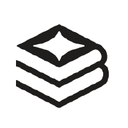}}
\newcommand{\compassicon}{\icon{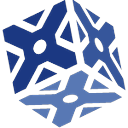}}
\newcommand{\microsoft}{\icon{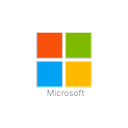}}
\newcommand{\fuse}{\icon{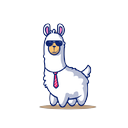}}
\newcommand{\openr}{\icon{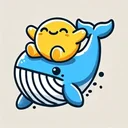}}
\newcommand{\qihoo}{\icon{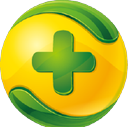}}
\newcommand{\bespoke}{\icon{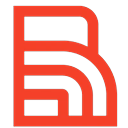}}

\newcommand{\bayesg}[1]{\ensuremath{\mathrm{Bayes}_{\mathbf{R}_0}@#1}}

\newcommand{\bayes}[1]{Bayes@$\!#1$}
\newcommand{\pass}[1]{Pass@$\!#1$}
\newcommand{\avg}[1]{avg@$\!#1$}
\newcommand{\aimefour}{AIME'24}
\newcommand{\aimefive}{AIME'25}

\newcommand{\hmmt}{HMMT'25}
\newcommand{\brumo}{BrUMO'25}
\newcommand{\scorio}{\texttt{Scorio}}

\newcommand{\bayesu}[1]{\ensuremath{\mathrm{Bayes}_{\mathcal{U}}@#1}}

\begin{document}
\maketitle
\input{main/abstract.tex}
\input{main/intro.tex}
\input{main/method.tex}
\input{main/experiments.tex}
\input{main/related_workds.tex}
\input{main/conclusion.tex}

\FloatBarrier
\clearpage
\input{main/limit.tex}
\input{main/acknowledge.tex}
\bibliography{custom}

\clearpage

\begingroup
\tableofcontents
\endgroup

\clearpage

\appendix

\crefalias{section}{appendix}
\crefname{appendix}{Appendix}{Appendices}
\Crefname{appendix}{Appendix}{Appendices}
\crefname{lstlisting}{Listing}{Listings}
\Crefname{lstlisting}{Listing}{Listings}

\input{appendix/notations.tex}

\input{appendix/accuracy.tex}
\input{appendix/gold_standard.tex}
\input{appendix/stability_one.tex}
\input{appendix/priors.tex}
\input{appendix/cat.tex}

\input{appendix/related_work.tex}

\input{appendix/experiment_setup.tex}
\input{appendix/scorio.tex}
\end{document}

%% file: main/abstract.tex
\begin{abstract}
Test-time scaling evaluates reasoning LLMs by sampling multiple outputs per prompt, but ranking models in this regime remains underexplored. We formalize dense benchmark ranking under test-time scaling and introduce \scorio{}, a library that implements statistical ranking methods such as paired-comparison models, item response theory (IRT) models, voting rules, and graph- and spectral-based methods. Across $20$ reasoning models on four Olympiad-style math benchmarks (\aimefour{}, \aimefive{}, \hmmt{}, and \brumo{}; up to $N=80$ trials), most full-trial rankings agree closely with the Bayesian gold standard \bayesu{80} (mean Kendall's $\tau_b = 0.93$--$0.95$), and $19$--$34$ methods recover exactly the same ordering. In the single-trial regime, the best methods reach $\tau_b \approx 0.86$. Using greedy decoding as an empirical prior (\bayesg{N}) reduces variance at $N=1$ by $16$--$52\%$, but can bias rankings when greedy and stochastic sampling disagree. These results identify reliable ranking methods for both high- and low-budget test-time scaling. We release \scorio{} as an open-source library at \href{https://github.com/mohsenhariri/scorio}{\adjustbox{valign=m}{\includegraphics[height=2.5ex]{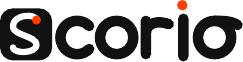}}}\footnote{\url{https://github.com/mohsenhariri/scorio}. See Appendix~\ref{sec:app:scorio} for API documentation.}.
\end{abstract}

%% file: main/intro.tex
\section{Introduction}\label{sec:intro}

\begin{figure*}[t]
    \centering
    \includegraphics[width=\textwidth]{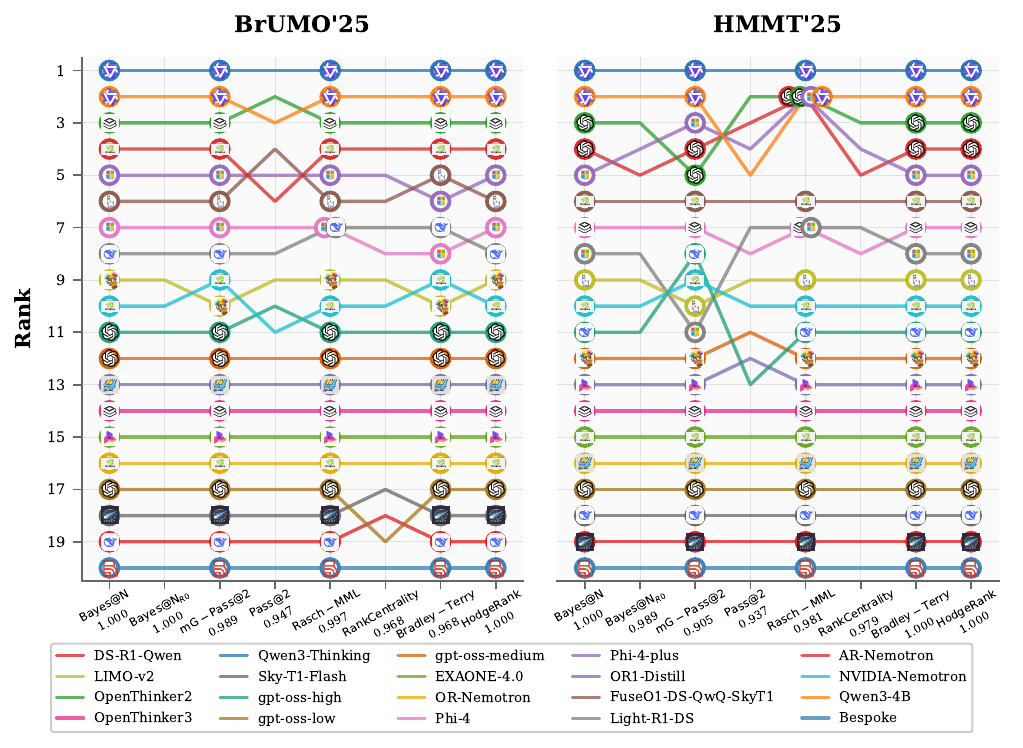}
    \caption{Agreement between each method's full-trial ranking and the gold standard. Kendall's $\tau_b$ is computed between each method's ranking (at $N=80$ trials) and \bayesu{80} on an easier benchmark (\brumo{}, left) and the hardest benchmark (\hmmt{}, right). On \brumo{}, multiple methods achieve near-perfect or perfect agreement: \bayesg{N} and HodgeRank reach $\tau_b = 1.0$, while Rasch MML achieves $0.997$. On \hmmt{}, Bradley--Terry and HodgeRank maintain perfect agreement ($\tau_b = 1.0$), but \bayesg{N} drops to $0.989$ and \pass{2} falls to $0.937$. This divergence is consistent with the lower greedy--sampling alignment observed on harder benchmarks (\cref{ssec:priors}).}
    \label{fig:showcase}
\end{figure*}

Large language models (LLMs) are increasingly used as general-purpose reasoning systems for tasks such as programming and mathematical problem solving \cite{chen2021evaluating,Wang2022SelfConsistency}. Reliable evaluation is therefore essential. In many settings, what matters is not only an absolute score but also a \emph{ranking} that supports model selection, deployment, and scientific comparison. This need is amplified by \emph{test-time scaling}, which allocates additional inference compute by sampling multiple outputs per prompt and aggregating them, turning evaluation into a repeated-sampling problem \cite{Wang2022SelfConsistency,Snell2024TestTimeCompute,Zeng2025Revisiting}.

Statistical ranking methods underpin two common LLM workflows. First, preference-based learning and alignment pipelines rely on human or model preferences over alternative responses, where the primitive observations are paired comparisons and downstream optimization depends on how those preferences are modeled and aggregated \cite{christiano2017deep,rafailov2023direct}. Second, model comparisons are often communicated through leaderboards. Crowdsourced paired-comparison platforms such as Chatbot Arena collect head-to-head judgments and fit rating or paired-comparison models to produce public rankings \cite{chiang2024chatbotarena}, while benchmark-style evaluations rank models by task performance metrics such as \pass{k} \cite{chen2021evaluating}. Recent work has revisited the statistical foundations of LLM ranking in both preference-based settings \cite{ameli2024statisticalframework} and benchmark settings, including IRT-style benchmarking \cite{zhou2025lost}. Different ranking methods can produce noticeably different model orderings, and their agreement can vary with benchmark difficulty (\cref{fig:showcase}).

A key practical distinction between these regimes is the \emph{representation} of the data used for ranking. Preference-based evaluation typically yields a sparse and evolving comparison graph because only a subset of model pairs are compared and the model pool changes over time \cite{chiang2024chatbotarena}. In contrast, benchmark evaluations produce dense outcomes for every model--question pair. For a fixed set of $L$ models and $M$ questions, we observe an outcome for every pair. Under test-time scaling, each model--question pair is evaluated with $N$ independent trials, producing a response tensor $\mathbf{R}\in\{0,1\}^{L\times M\times N}$. This dense repeated-trial setting raises new methodological questions: Which ranking rule should be used when $N$ is small? How quickly do different ranking methods stabilize as $N$ grows? How do priors and uncertainty estimates affect ranking robustness?

In this work, we study performance-based ranking under test-time scaling. We formalize the dense benchmark setting through the response tensor $\mathbf{R}$, evaluate ranking methods by their low-budget stability and convergence as the test-time budget increases, and implement the studied methods in \scorio{}.

We summarize our contributions as follows:
\begin{itemize}[leftmargin=*]
  \item We formalize dense benchmark ranking under test-time scaling via $\mathbf{R}\in\{0,1\}^{L\times M\times N}$ and connect common ranking families through pointwise, pairwise, and setwise transformations of $\mathbf{R}$.
  \item We propose an evaluation protocol based on low-budget stability (agreement between rankings computed from subsampled trials and reference rankings) and convergence with increasing numbers of trials.
  \item We compare a broad suite of ranking methods across $20$ reasoning models and four Olympiad-style math benchmarks (up to $N=80$ trials), characterizing where method families agree and where they diverge.
  \item We analyze Bayesian and uncertainty-aware ranking choices, including priors and conservative (quantile-based) scoring, and quantify their bias--variance trade-offs in low-trial regimes.
  \item We release \scorio{}, a library implementing the ranking methods and Bayesian options.
\end{itemize}

%% file: main/method.tex
\section{Ranking Problem and Test-time Scaling}\label{sec:framework}

In classical statistical settings, there is no canonical theoretical ground truth or empirical gold standard against which competing ranking rules can be judged. Choosing among methods therefore usually requires additional modeling assumptions. Test-time scaling offers a useful alternative: because each model--question pair can be sampled repeatedly, it lets us evaluate ranking methods by how stable they are in low-budget settings and how quickly they converge as more trials are observed.

Statistical ranking methods are widely used in domains such as sports competitions (e.g., paired-comparison models and rating systems for head-to-head games) \cite{BradleyTerry1952,Elo1978,Glickman1999} and voting or collective decision-making \cite{Borda1781,Condorcet1785,Arrow1951}. In such settings, there are $L$ entities to be ranked (e.g., players, items, or models) over $M$ tasks (e.g., matches, questions, or instances). Test-time scaling adds a third dimension: $N$, the number of i.i.d.\ samples generated for a fixed question $m\in\{1,\dots,M\}$. Repeated sampling lets us study two complementary properties. First, \emph{low-budget stability} asks whether a ranking computed from a small number of trials agrees with a high-budget reference ranking. In our experiments, the low-budget case is $N=1$: we subsample one trial per question, compute the ranking, repeat this over the available single-trial draws, and compare each ranking either with an empirical gold standard or with the same method's full-trial ranking. Second, \emph{convergence} asks how quickly rankings computed from $n$ trials approach the full-trial ordering as $n$ increases from $1$ to $N$.

\subsection{Gold Standard Rankings}\label{ssec:gold_standard}

Evaluation metrics widely used in test-time scaling, such as \pass{k} and \bayes{N}, can be analyzed through statistical properties such as bias. For instance, \citet{chen2021evaluating} derive an unbiased estimator for \pass{k}. As the number of trials $N$ grows, empirical estimates of these metrics concentrate around their population values, making metric-based rankings increasingly stable. In particular, for binary outcomes, \bayesu{N} is order-equivalent to mean accuracy \avg{N} \cite{hariri2025dontpasskbayesianframework}, which motivates our use of the full-trial \bayesu{N} ranking as an empirical accuracy-based gold standard.

This reasoning does not extend automatically to all ranking methods. Even as the number of questions $M$ or trials $N$ increases, different ranking methods need not converge to a unique limiting ordering, such as the one induced by average accuracy (Appendix \ref{app:ssub:nonconvergence}). Unlike evaluation metrics, ranking algorithms can emphasize different aspects of performance across tasks, players, or items. In \cref{sec:exp}, we show that rankings induced by probabilistic models (e.g., Bradley--Terry) can differ from those induced by expected-performance metrics (e.g., mean accuracy or Bayesian estimates).

Given the absence of a universal gold standard for ranking methods, we use two target rankings for comparison. First, we define an empirical gold standard based on average performance over all trials with a large sample size (e.g., $N=80$). This target captures aggregate performance across tasks and trials while allowing ties. This choice is justified for several reasons: (a) the ranking induced by average performance is order-equivalent to the ranking induced by Bayesian estimation with a uniform prior (\bayesu{N}); (b) when $N$ is large, average performance is among the most stable ranking rules relative to the alternatives (\cref{ssec:exp:gold_standard}); and (c) it is easy to interpret, widely used in practice, and yields absolute performance values.

The second target ranking is the ordering produced by a method itself (method@$80$) when all available trials are aggregated. This target lets us assess a method's self-consistency and convergence as more data become available.

\subsection{Representation}\label{ssec:representation}

We consider $L$ models evaluated on a benchmark of $M$ questions under test-time scaling, generating $N$ i.i.d.\ trials per model--question pair. Let $\mathcal{L}=\{1,\dots,L\}$ index models and $\mathcal{Q}=\{1,\dots,M\}$ questions; for each question we observe $N$ independent trials indexed by $n\in\{1,\dots,N\}$. For each $(l,m,n)\in \mathcal{L}\times\mathcal{Q}\times\{1,\dots,N\}$ we observe a binary outcome
\begin{equation}
R_{lmn} \in \{0,1\},
\end{equation}
where $R_{lmn}=1$ if model $l$ solves question $m$ on trial $n$. We collect these outcomes in a response tensor $\mathbf{R}\in\{0,1\}^{L\times M\times N}$. When $N=1$, this reduces to the standard single-run benchmark setting. Unlike crowdsourced paired-comparison datasets (e.g., Chatbot Arena \cite{chiang2024chatbotarena}), where the primitive observations are model--model outcomes on a possibly sparse comparison graph, our benchmark setting produces outcomes for every model--question pair. We therefore take $\mathbf{R}$ as the primitive object; all ranking methods we study use $\mathbf{R}$ as input, but they differ in the representations on which they operate after transforming or aggregating it.

\paragraph{Pointwise (model--question) representation.}
Define the per-question solve rate
\begin{equation}
\widehat{p}_{lm} := \frac{1}{N}\sum_{n=1}^N R_{lmn},
\end{equation}
and the overall mean accuracy $\widehat{p}_{l} := \frac{1}{M}\sum_{m=1}^M \widehat{p}_{lm}$. Pointwise and IRT-style methods operate on the matrix $\widehat{\mathbf{P}}=[\widehat{p}_{lm}]\in[0,1]^{L\times M}$ (or on its row means), optionally reweighting questions (e.g., inverse-difficulty weighting \cite{gotou2020correctiondifficulty}). Classical IRT models infer latent abilities from this representation \cite{rasch1960rasch,birnbaum1968latent}, and have recently been applied to LLM benchmarking \cite{zhou2025lost}. When $N>1$, the trial axis corresponds to repeated Bernoulli observations; likelihood-based models (including IRT) can equivalently work with the sufficient statistic $k_{lm}:=\sum_{n}R_{lmn}$, yielding a binomial-response formulation \cite{mccullaghnelder1989glm,deboeckwilson2004explanatory}. Related repeated-measures and longitudinal IRT extensions are also well studied \cite{verhelst1993dynamicrasch,wang2019longitudinalirt}. Evaluation-metric rankings (e.g., \pass{k} and \bayes{N}) additionally use the per-question trial multiset $\{R_{lm1},\dots,R_{lmN}\}$ (equivalently the count $\sum_n R_{lmn}$) to compute per-question metrics before aggregating across $m$ \cite{chen2021evaluating}.

\paragraph{Pairwise (win/tie) representation.}
Many classical ranking methods reduce $\mathbf{R}$ to pairwise outcomes. For a pair of models $(i,j)\in\mathcal{L}^2$ we define win and tie counts
\begin{align}
W_{ij} &:= \sum_{m=1}^M\sum_{n=1}^N \mathbf{1}\{R_{imn}=1,\; R_{jmn}=0\}, \\
T_{ij} &:= \sum_{m=1}^M\sum_{n=1}^N \mathbf{1}\{R_{imn}=R_{jmn}\},
\end{align}
so that, in our fully observed setting, $W_{ij}+W_{ji}+T_{ij}=MN$ for all $i\neq j$. Equivalently, we can form an undirected comparison graph $G=(V,E)$ with vertex set $V=\mathcal{L}$ and edge set $E=\{\{i,j\}:W_{ij}+W_{ji}+T_{ij}>0\}$, and store $(W_{ij},W_{ji},T_{ij})$ on each edge. In our benchmark setting $E$ is the complete graph (every pair is compared $MN$ times), whereas in interactive evaluation settings $E$ is typically sparse and one assumes $G$ is connected. The matrices $\mathbf{W}=[W_{ij}]$ and $\mathbf{T}=[T_{ij}]$ define a weighted comparison graph over models. Probabilistic paired-comparison models (e.g., Bradley--Terry and tie extensions \cite{BradleyTerry1952,rao1967ties,davidson1970bties}) and voting rules (e.g., Borda and Copeland \cite{Borda1781,brandt2016compsocchoice}) use these aggregated counts; graph- and spectral-based methods (e.g., PageRank, Rank Centrality, HodgeRank, SerialRank, AlphaRank, and Nash-based ranking \cite{page1999pagerank,negahban2017rankcentrality,jiang2009hodgerank,fogel2016serialrank,omidshafiei2019alpharank,balduzzi2019openended}) further transform $(\mathbf{W},\mathbf{T})$ into Markov chains or skew-symmetric edge flows, typically via edge weights based on empirical win rates such as $\widehat{P}_{i\succ j}=(W_{ij}+\tfrac{1}{2}T_{ij})/(W_{ij}+W_{ji}+T_{ij})$. Sequential rating systems (e.g., Elo and TrueSkill \cite{Elo1978,herbrich2006trueskill}) instead process the underlying stream of pairwise ``matches'' induced by each question--trial $(m,n)$.

\paragraph{Listwise or setwise representation.}
For each question--trial $(m,n)$ we define the winning set $U_{mn}:=\{l\in\mathcal{L} : R_{lmn}=1\}$ and the losing set $\mathcal{L}\setminus U_{mn}$, which induces a two-level partial order: all winners tie above all losers. Setwise or listwise models (e.g., Plackett--Luce \cite{plackett1975permutations,luce1959choice} and Davidson--Luce \cite{firth2019davidsonluce}) operate directly on the collection of events $\{(U_{mn},\mathcal{L}\setminus U_{mn})\}_{m,n}$, discarding degenerate events with $U_{mn}=\emptyset$ or $U_{mn}=\mathcal{L}$. In our binary two-level setting, Plackett--Luce likelihoods collapse to functions of pairwise win counts (cf.\ the MM formulation for generalized Bradley--Terry and Plackett--Luce likelihoods \cite{hunter2004mmbradleyterry}), whereas Davidson--Luce explicitly models within-set ties.

\subsection{Bayesian Approaches in Ranking}\label{ssec:bayesian_ranking}

Many ranking methods can be viewed as probabilistic models with latent parameters $\theta$ (e.g., model strength and, optionally, question difficulty). Given observations $\mathbf{R}$ (or derived representations such as pairwise counts; \cref{ssec:representation}), inference reduces to estimating $\theta$ from a likelihood $p(\mathbf{R}\mid\theta)$. We consider maximum likelihood estimation (MLE), maximum a posteriori (MAP), and expected a posteriori (EAP), and discuss how uncertainty can be propagated to rankings \cite{Gelman2013BDA}. Although MLE is not Bayesian, we include it as a standard baseline for likelihood-based ranking models.

\paragraph{Maximum likelihood estimation (MLE).}
The maximum likelihood estimate is
\begin{equation}
\hat{\theta}_{\mathrm{MLE}} \in \arg\max_{\theta} \; p(\mathbf{R} \mid \theta),
\end{equation}
which yields a point estimate without requiring a prior. MLE is attractive for its simplicity, but in paired-comparison and IRT-like models it can be unstable under (near-)separation or weak identification, which motivates priors in MAP and EAP.

\paragraph{Maximum a posteriori (MAP).}
MAP incorporates prior information $p(\theta)$ and estimates the posterior mode:
\begin{equation}
\hat{\theta}_{\mathrm{MAP}} \in \arg\max_{\theta} \; p(\mathbf{R} \mid \theta)\, p(\theta).
\end{equation}
Equivalently, MAP is a penalized MLE in which $-\log p(\theta)$ acts as a regularizer; priors can improve stability in paired-comparison and IRT-style models \cite{CaronDoucet2012,Mislevy1986}. We can also construct \emph{empirical} priors from auxiliary evaluation runs. For example, a prior outcome tensor $\mathbf{R}_0$ (e.g., one greedy decode per question) can be used to regularize stochastic trials (\texttt{EmpiricalPrior} in \scorio{}) \cite{hariri2025dontpasskbayesianframework}.

\paragraph{Expected a posteriori (EAP).}
EAP uses the posterior mean as the point estimate:
\begin{equation}
\hat{\theta}_{\mathrm{EAP}} := \mathbb{E}[\theta \mid \mathbf{R}],
\end{equation}
which is Bayes-optimal under squared-error loss \cite{Gelman2013BDA}. Compared with MAP, EAP accounts for posterior mass beyond the mode and typically requires approximation or sampling. EAP is common in latent-trait settings such as IRT and adaptive testing \cite{ChenHouDodd1998}.

\paragraph{Interval estimates and conservative ranking.}
Bayesian methods naturally yield \emph{credible intervals} (posterior quantiles) for each $\theta_l$, while frequentist analyses can produce approximate \emph{confidence intervals} for $\hat{\theta}_{\mathrm{MLE}}$ via bootstrap resampling of questions or trials. Interval estimates are especially useful because ranking is sensitive to near ties: rather than ranking by point estimates alone, one can rank conservatively using a lower credible or confidence bound (LCB), or report pairwise superiority probabilities $\Pr(\theta_i > \theta_j \mid \mathbf{R})$. Metric-level Bayesian estimators such as \bayes{N} provide both a posterior mean and uncertainty, enabling rankings by posterior mean or by a chosen posterior quantile. \bayes{N} also supports incorporating prior outcomes $\mathbf{R}_0$ (e.g., one greedy decode per question) as pseudo-counts in the posterior, which is complementary to using $\mathbf{R}_0$ to define empirical priors for MAP in parametric ranking models. Our implementation in \scorio{} supports both credible-interval ranking via \bayes{N} and empirical priors via \texttt{EmpiricalPrior} for MAP estimation.

%% file: main/experiments.tex
\section{Experiments}\label{sec:exp}
We evaluate $72$ ranking methods (Appendix \ref{ssec:app:setup:ranking_apis}) on four Olympiad-style math benchmarks: \aimefour{}, \aimefive{}, \hmmt{}, and \brumo{}, each with $M=30$ questions. We use $L=20$ reasoning LLMs (full list in \cref{app:tab:modelid}). For each model--question pair, we collect $N=80$ independent trials via top-$p$ sampling, yielding a response tensor $\mathbf{R}\in\{0,1\}^{20\times 30\times 80}$. We also collect a single greedy-decoding output per question ($\mathbf{R}_0$) to serve as an empirical prior. Detailed generation, sampling, and reproducibility settings appear in \cref{sec:app:setup}; the library API is documented in \cref{sec:app:scorio}.

\subsection{Gold Standard Ranking}\label{ssec:exp:gold_standard}
Following \cref{ssec:gold_standard}, we define the gold-standard ranking as \bayesu{80}, the Bayesian posterior-mean estimator with a uniform prior computed from all $N=80$ trials. This choice is order-equivalent to \avg{80} (mean correctness over all $M$ questions and all $N=80$ trials, with ties allowed) and yields an interpretable accuracy-based target. Empirically, when each of our $72$ ranking methods is computed using all $80$ trials, the resulting orderings agree closely with \bayesu{80} (\cref{tab:gold_standard_agreement}): across benchmarks, the average Kendall's $\tau_b$ between \bayesu{80} and the other methods is $0.93$--$0.95$ (median $0.95$--$0.99$), and $19$--$34$ methods recover exactly the same ordering ($\tau_b=1$). The largest deviations come from a small set of voting rules (e.g., minimax and Nanson variants) and difficulty-weighted baselines, with minimum $\tau_b$ values of $0.68$--$0.79$ depending on the benchmark. Although \bayesu{N} is order-equivalent to \avg{N}, we prefer the Bayesian formulation because it supports priors (e.g., \bayesg{N}) and uncertainty estimates.

\input{Tables/gold_standard_agreement.tex}

\subsection{Ranking-Method Stability}\label{ssec:exp:n1_stability}
To compare ranking methods in the low-budget regime, we set $N=1$ by subsampling one of the $80$ trials per question and recomputing the rankings. For each method, we report Kendall's $\tau_b$ averaged over the $80$ single-trial draws (mean $\pm$ std). Since the \pass{k} family requires at least two trials to differ from mean accuracy, the $N=1$ comparisons below cover the remaining $69$ methods.

\paragraph{Gold-standard agreement.}
We first rank methods by agreement with the empirical gold standard (\bayesu{80}). Across \aimefour{}, \aimefive{}, and \brumo{}, \bayesg{N} performs best, achieving $\tau_b=0.779\pm0.034$, $0.798\pm0.045$, and $0.858\pm0.028$, respectively (\cref{tab:n1_winners}). On \hmmt{}, the hardest benchmark (see \cref{sec:app:accuracy}), the greedy prior no longer helps, and the best score is shared by a $21$-method equivalence class (\bayesu{N} and several graph- and voting-based methods), with $\tau_b=0.790\pm0.053$. When all benchmarks are pooled (Combined), the same $21$-method class attains $\tau_b=0.865\pm0.049$, while \bayesg{N} drops to $\tau_b=0.786\pm0.031$ (\cref{tab:n1_gs_combined}).

\paragraph{Self-consistency and convergence.}
Next, we evaluate each method against its own full-trial ranking (method@$80$), which summarizes convergence from $N=1$ to $N=80$. Rasch MML with LCB scoring is the most self-consistent on \aimefour{}, \aimefive{}, and \hmmt{}, with $\tau_b=0.804\pm0.051$, $0.834\pm0.054$, and $0.810\pm0.056$ (\cref{tab:n1_winners}); \brumo{} again favors \bayesg{N} ($0.858\pm0.028$). On the Combined benchmark, the most self-consistent method is Nanson's rule with tie averaging ($0.892\pm0.050$), followed by Rasch MML (LCB) ($0.883\pm0.037$), whereas several minimax variants are among the least self-consistent (down to $0.765\pm0.045$; \cref{tab:n1_self_combined}). High self-consistency does not imply strong agreement with the gold standard: Nanson (avg ties) ranks first in self-consistency on Combined but has substantially lower gold-standard agreement ($0.807\pm0.036$; \cref{tab:n1_gs_combined}).

\input{Tables/n1_winners.tex}

\subsection{Bootstrapped Model-Pool Robustness}\label{ssec:exp:model_pool}
The preceding $N=1$ results use the full set of $20$ models. To test whether those conclusions depend on the evaluation pool, we repeat the low-budget analysis on bootstrapped model pools of size $5$, $10$, and $15$. For each bootstrap subset, we recompute the full-trial rankings, use the subset-specific \avg{80} ordering as the gold-standard target, and compare each method's $80$ single-trial rankings against two references: (i) the subset-specific \avg{80} ordering and (ii) its own subset-specific full-trial ranking (method@$80$). We aggregate $1000$ bootstrap subsets for each benchmark--size setting.

\paragraph{Easy and medium benchmarks preserve the original winner.}
On \aimefour{}, \aimefive{}, and \brumo{}, \bayesg{N} remains the best representative method under both targets at all three model-pool sizes (\cref{tab:model_pool_bootstrap_winners}). The mean score changes only slightly with pool size: on \aimefour{}, gold-standard agreement moves from $0.769$ to $0.780$ and self-consistency from $0.773$ to $0.785$ as the pool size increases from $5$ to $15$ models; on \aimefive{}, the corresponding ranges are $0.797$--$0.802$ and $0.803$--$0.809$; on \brumo{}, \bayesg{N} stays near $0.854$--$0.858$ for both targets. On \brumo{}, this advantage also becomes more decisive as the pool grows: the fraction of subsets where \bayesg{N} is the top-scoring method rises from about $0.69$ at $k=5$ to $0.98$--$0.99$ at $k=15$.

\paragraph{Harder benchmarks remain tie-rich.}
The harder settings behave differently. On \hmmt{} and on the Combined benchmark, the top score is not unique: for agreement with \avg{80}, an equivalence class of $29$--$30$ methods shares the best mean, while for method@$80$ the tied class still contains $13$--$14$ methods. We report \texttt{avg} (\avg{N}, order-equivalent to \bayesu{N}) as a representative member of these tied classes. The tied optimum is essentially flat across pool size, staying near $0.788$--$0.790$ on \hmmt{} and $0.863$--$0.866$ on Combined. This mirrors the full-model analysis in \cref{ssec:exp:n1_stability}: once the benchmark is difficult or pooled across heterogeneous tasks, many pointwise, voting, and graph-based methods become empirically indistinguishable.

\paragraph{Larger pools mainly reduce between-subset variance.}
The primary effect of increasing the model-pool size is to reduce dispersion across subsets rather than to shift the mean systematically (\cref{tab:model_pool_bootstrap_winners}). For the best method under the \avg{80} target, the across-subset standard deviation falls from $0.209$ to $0.057$ on \aimefour{}, from $0.144$ to $0.038$ on \aimefive{}, from $0.114$ to $0.033$ on \hmmt{}, from $0.136$ to $0.032$ on \brumo{}, and from $0.084$ to $0.023$ on Combined when moving from $5$ to $15$ models. Thus, the qualitative recommendation is stable under moderate changes to the model pool: larger pools mainly make the same conclusion more certain.

\input{Tables/model_pool_bootstrap_winners.tex}

\subsection{Effect of Empirical Priors}\label{ssec:priors}

Empirical priors use auxiliary evaluation signals to stabilize low-budget rankings. In our setting, the signal is a single greedy decode, $\mathbf{R}_0$. We incorporate $\mathbf{R}_0$ into \bayes{N}, yielding \bayesg{N}, and compare it with the uniform-prior variant \bayesu{N}. We evaluate both variants by their agreement with the gold-standard ranking \bayesu{80}. For each $N$, we compute Kendall's $\tau_b$ between the induced model ranking and \bayesu{80} and report the mean and standard deviation over $50$ resampled datasets.

\begin{figure}[t]
    \centering
    \includegraphics[width=\columnwidth]{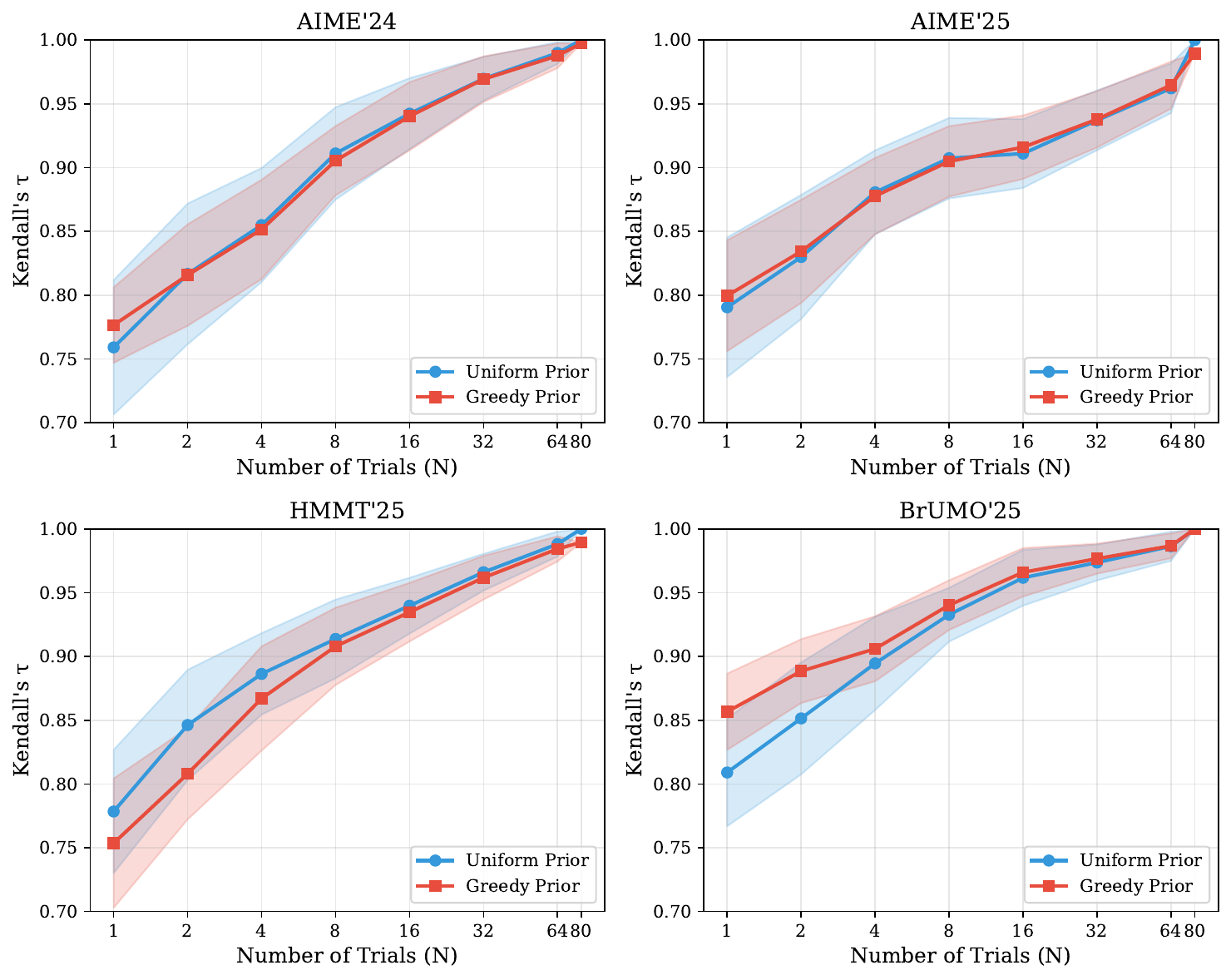}
    \caption{Gold-standard agreement of \bayesu{N} (blue) and \bayesg{N} (red) as a function of $N$ across benchmarks. Shaded regions show $\pm 1$ standard deviation over $50$ resampled datasets.}
    \label{fig:prior_per_dataset}
\end{figure}

\paragraph{Empirical priors reduce variance at low $N$.}
Across all benchmarks, \bayesg{N} yields more stable low-$N$ rankings than \bayesu{N}. At $N=1$, the standard deviation of $\tau_b$ decreases by $16$--$52\%$ depending on the benchmark (\cref{tab:prior_effect,fig:bootstrap_violin}). This advantage shrinks quickly as $N$ increases (\cref{fig:prior_per_dataset}), consistent with the prior contributing only $O(1)$ pseudo-counts per question.

\begin{table}[t]
\centering
\small
\caption{Dataset difficulty (mean accuracy), greedy--sampling alignment ($\tau_{\text{G-S}}$), and the effect of the greedy empirical prior at $N=1$. $\Delta\tau$ is the difference in gold-standard agreement (greedy minus uniform), and Std.\ Red.\ is the relative reduction in the standard deviation of $\tau_b$.}
\label{tab:prior_effect}
\begin{tabular}{lcccc}
\toprule
Benchmark & Difficulty & $\tau_{\text{G-S}}$ & $\Delta\tau$ & Std.\ Red. \\
\midrule
\aimefour & 0.620 & 0.739 & $+0.020$ & 42\% \\
\aimefive & 0.533 & 0.660 & $+0.008$ & 17\% \\
\hmmt & \textbf{0.333} & \textbf{0.635} & $\mathbf{-0.022}$ & 16\% \\
\brumo & 0.588 & \textbf{0.768} & $\mathbf{+0.049}$ & 52\% \\
\bottomrule
\end{tabular}
\end{table}

\paragraph{The mean effect depends on greedy--sampling alignment.}
Variance reduction does not guarantee improved agreement with \bayesu{80}. The greedy prior increases mean $\tau_b$ on \aimefour{}, \aimefive{}, and \brumo{}, but decreases it on \hmmt{} (\cref{tab:prior_effect}). At $N=1$, when all benchmarks are pooled, this negative shift is substantially larger (\cref{tab:n1_gs_combined}), indicating that an empirical prior can introduce systematic bias when greedy and sampling behave differently across datasets.

We summarize this diagnostic via \emph{greedy--sampling alignment} $\tau_{\text{G-S}}$, defined as Kendall's $\tau_b$ between the model rankings induced by greedy decoding and by stochastic sampling at $N=80$. In our results, higher $\tau_{\text{G-S}}$ coincides with a more positive $\Delta\tau$ (\cref{sec:app:prior_diagnostics,fig:difficulty_correlation}), suggesting that the empirical prior is most likely to help when greedy is a faithful proxy for the sampling-induced ordering. While this evidence is limited to four benchmarks, the trend is consistent with \bayesg{N} acting as shrinkage toward the greedy ordering.

\begin{figure}[t]
    \centering
    \includegraphics[width=\columnwidth]{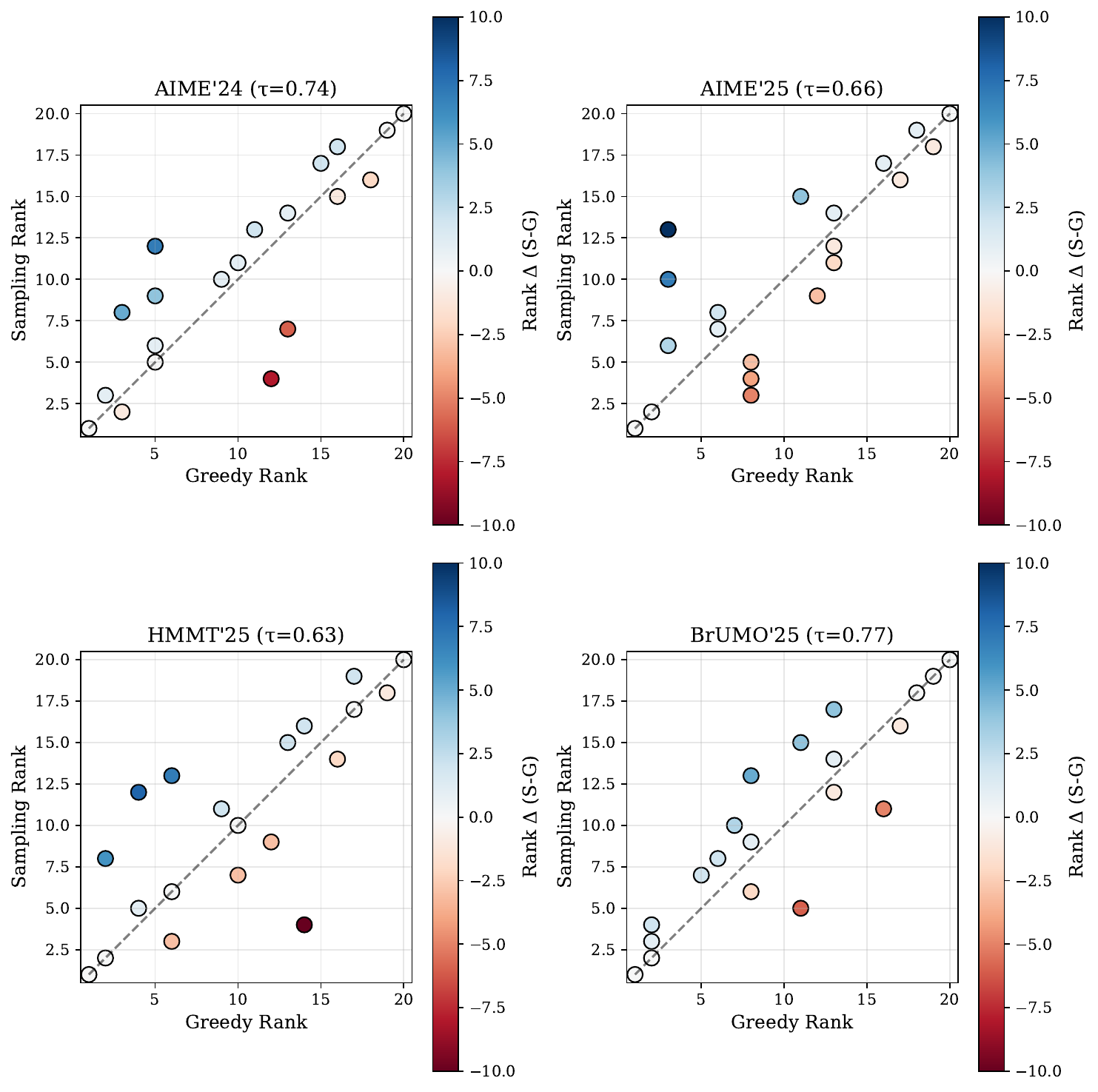}
    \caption{Model-level ranks under greedy decoding versus stochastic sampling ($N=80$) for each benchmark. Points on the diagonal indicate perfect alignment; color shows rank displacement ($\Delta$).}
    \label{fig:rank_scatter}
\end{figure}

\paragraph{Implications.}
\bayesg{N} behaves as a shrinkage estimator toward the greedy ordering: it is helpful when greedy decoding is a faithful proxy for the sampling-induced ranking, and harmful when the two disagree. Because $\mathbf{R}_0$ is generated under a different decoding policy, incorporating it effectively biases the estimate toward greedy behavior. This can be desirable for variance reduction, but it changes the implied evaluation target. A plausible source of disagreement is that greedy decoding may under-explore on hard instances, while stochastic sampling can recover alternative successful reasoning paths. In practice, empirical priors are most attractive when $N$ is very small and greedy--sampling alignment has been checked on a small pilot sample; otherwise, \bayesu{N} provides a safer default.

\paragraph{Bias--variance trade-off.}
\Cref{fig:bootstrap_violin} visualizes the trade-off induced by empirical priors: in our benchmarks, the greedy prior reduces variability (narrower distributions) but can introduce bias (shifted means), with the net effect governed by greedy--sampling alignment.

\subsection{Categorical Ranking}\label{sec:exp:cat}

We extend the Bayesian framework to \emph{categorical outcomes}: each completion is mapped to one of $C+1$ ordered categories based on signals such as answer format (boxed vs.\ unboxed), model confidence (completion bits per token), token efficiency, and external verifier judgments. Each scheme defines a categorical mapping and a utility weight vector $\mathbf{w}=(w_0,\dots,w_C)$; Bayesian estimation then proceeds with a Dirichlet--multinomial model rather than a Beta--binomial model (details and scheme definitions are given in \cref{sec:app:cat}).

We select eight non-redundant representative schemes. Using the $N=1$ subsampling protocol on the Combined benchmark (the first $L=11$ models of \cref{app:tab:modelid}, $M=120$ questions pooled across all four datasets), we measure Kendall's $\tau_b$ against three references (\cref{tab:cat_combined}).

\input{Tables/categorical_combined.tex}

\begin{figure}[t]
    \centering
    \includegraphics[width=\columnwidth]{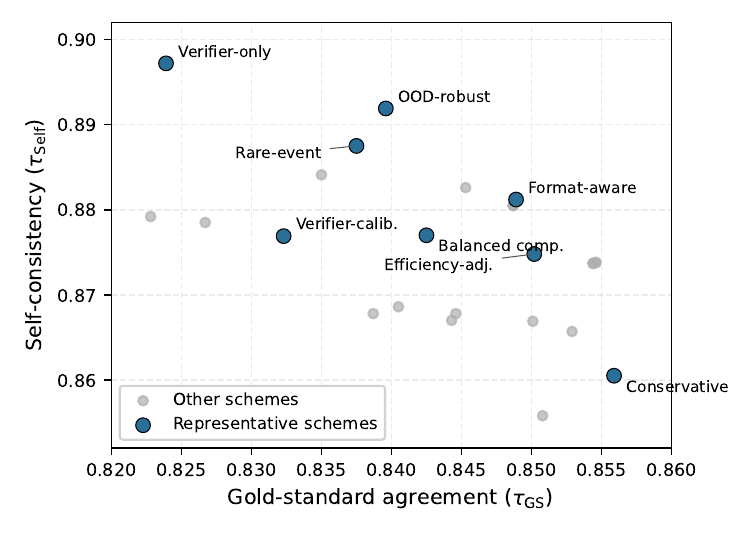}
    \caption{Gold-standard agreement vs.\ self-consistency for $25$ categorical schemes at $N=1$ on the Combined benchmark. Blue markers indicate the $8$ representative schemes; gray markers show the remaining $17$. Schemes in the upper-left are self-consistent but deviate from \bayesu{80}; those in the lower-right closely track the gold standard but are less stable across single-trial draws.}
    \label{fig:cat_tradeoff}
\end{figure}

\paragraph{Self-consistency vs.\ gold-standard trade-off.}
Signal-rich schemes achieve the highest self-consistency: Verifier-only ($\tau_{\text{Self}} = 0.897$) and OOD-robust ($0.892$) rank first and second (\cref{fig:cat_tradeoff}). Yet these schemes have the lowest agreement with the gold standard ($\tau_{\text{GS}} = 0.824$ and $0.840$, respectively), extending the finding from \cref{ssec:exp:n1_stability} that high self-consistency does not imply closeness to the gold standard. The negative correlation between $\tau_{\text{GS}}$ and $\tau_{\text{Self}}$ across schemes (\cref{fig:cat_tradeoff}) suggests that auxiliary signals introduce systematic biases away from the correctness-based ordering while stabilizing single-trial rankings.

\paragraph{Greedy-prior alignment.}
All eight schemes correlate more strongly with \bayesg{80} than with \bayesu{80}; the gap is largest for Verifier-only ($\Delta\tau = +0.046$) and OOD-robust ($+0.031$), consistent with the mechanism in \cref{ssec:priors}: verifier and OOD signals encode information partially aligned with greedy-decoding behavior. Per-dataset results (\cref{sec:app:cat}) show that scheme differentiation widens on harder benchmarks (\hmmt{}, \brumo{}), where Verifier-only drops to $\tau_{\text{GS}}=0.753$ and $0.734$, while correctness-driven schemes remain stable ($\tau_{\text{GS}}\ge 0.80$).

%% file: Tables/gold_standard_agreement.tex
\begin{table}[t]
\centering
\small
\setlength{\tabcolsep}{4pt}
\caption{Agreement between the gold-standard ranking (\bayesu{80}) and each other ranking method, measured by Kendall's $\tau_b$, when all methods are computed from the full $N=80$ trials. Statistics are computed over the other $71$ methods; ``Combined'' pools all benchmarks.}
\label{tab:gold_standard_agreement}
\resizebox{\columnwidth}{!}{%
\begin{tabular}{lccccc}
\toprule
Benchmark & Mean & Median & Min & \#($\tau_b=1$) & \#($\tau_b\ge 0.95$) \\
\midrule
\aimefour & 0.941 & 0.989 & 0.682 & 20 & 40 \\
\aimefive & 0.934 & 0.947 & 0.771 & 19 & 29 \\
\hmmt & 0.950 & 0.989 & 0.758 & 34 & 44 \\
\brumo & 0.954 & 0.968 & 0.789 & 26 & 49 \\
\midrule
Combined & 0.962 & 0.989 & 0.748 & 22 & 53 \\
\bottomrule
\end{tabular}%
}
\end{table}

%% file: Tables/n1_winners.tex
\begin{table*}[t]
\centering
\small
\setlength{\tabcolsep}{4pt}
\caption{Best-performing ranking methods in the low-budget regime ($N=1$) under two targets: (i) agreement with the gold standard (\bayesu{80}) and (ii) self-consistency with each method's own full-trial ranking (method@$80$). Kendall's $\tau_b$ is averaged over $80$ single-trial draws; $\dagger$ denotes a 21-way tie for best gold-standard agreement (see~\cref{tab:n1_gs_combined}). \pass{k} variants are excluded at $N=1$ because they require $N \ge 2$. Method identifiers correspond to the APIs listed in \cref{ssec:app:setup:ranking_apis}.}
\label{tab:n1_winners}
\begin{tabular}{l l c l c}
\toprule
Benchmark & Best vs.\ gold standard & $\tau_b$ & Best self-consistency (vs. \texttt{method@80}) & $\tau_b$ \\
\midrule
\aimefour & \bayesg{1 }  & $0.779 \pm 0.034$ & Rasch MML LCB (\texttt{rasch\_mml\_credible}) & $0.804 \pm 0.051$ \\
\aimefive & \bayesg{1 }  & $0.798 \pm 0.045$ & Rasch MML LCB (\texttt{rasch\_mml\_credible}) & $0.834 \pm 0.054$ \\
\hmmt & \bayes{1 } \textsuperscript{$\dagger$} & $0.790 \pm 0.053$ & Rasch MML LCB (\texttt{rasch\_mml\_credible}) & $0.810 \pm 0.056$ \\
\brumo & \bayesg{1 }  & $0.858 \pm 0.028$ & \bayesg{1 }  & $0.858 \pm 0.028$ \\
Combined & \bayes{1 } \textsuperscript{$\dagger$} & $0.865 \pm 0.049$ & Nanson avg ties (\texttt{nanson\_rank\_ties\_average}) & $0.892 \pm 0.050$ \\
\bottomrule
\end{tabular}
\end{table*}

%% file: Tables/model_pool_bootstrap_winners.tex
\begin{table}[t]
\centering
\small
\setlength{\tabcolsep}{1.5pt}
\caption{Bootstrapped model-pool results in the low-budget regime ($N=1$). For each model-pool subset, we compute Kendall's $\tau_b$ over the $80$ single-trial rankings against two targets: the subset-specific gold standard \texttt{avg@80} and each method's own subset-specific full-trial ranking (\texttt{method@80}). The table reports the mean and standard deviation of the subset-level mean score across bootstrap model pools for each subset size.}
\label{tab:model_pool_bootstrap_winners}
\begin{tabular}{l c l c c}
\toprule
\multirow{2}{*}{Benchmark} & \multirow{2}{*}{Pool} & \multirow{2}{*}{Best Method} & \multicolumn{2}{c}{$\tau_b$ vs Target} \\
\cmidrule(lr){4-5}
& & & \texttt{avg@80} & \texttt{method@80} \\
\midrule
\multirow{3}{*}{\aimefour} & 5 & \multirow{3}{*}{\bayesg{1}} & $0.769 \pm {\scriptstyle 0.209}$ & $0.773 \pm {\scriptstyle 0.207}$ \\
& 10 & & $0.776 \pm {\scriptstyle 0.107}$ & $0.781 \pm {\scriptstyle 0.105}$ \\
& 15 & & $0.780 \pm {\scriptstyle 0.057}$ & $0.785 \pm {\scriptstyle 0.057}$ \\
\addlinespace
\multirow{3}{*}{\aimefive} & 5 & \multirow{3}{*}{\bayesg{1}} & $0.802 \pm {\scriptstyle 0.144}$ & $0.809 \pm {\scriptstyle 0.144}$ \\
& 10 & & $0.797 \pm {\scriptstyle 0.071}$ & $0.803 \pm {\scriptstyle 0.073}$ \\
& 15 & & $0.798 \pm {\scriptstyle 0.038}$ & $0.804 \pm {\scriptstyle 0.040}$ \\
\addlinespace
\multirow{3}{*}{\hmmt} & 5 & \multirow{3}{*}{\bayes{1} } & $0.788 \pm {\scriptstyle 0.114}$ & $0.788 \pm {\scriptstyle 0.114}$ \\
& 10 & & $0.789 \pm {\scriptstyle 0.059}$ & $0.789 \pm {\scriptstyle 0.059}$ \\
& 15 & & $0.790 \pm {\scriptstyle 0.033}$ & $0.790 \pm {\scriptstyle 0.033}$ \\
\addlinespace
\multirow{3}{*}{\brumo} & 5 & \multirow{3}{*}{\bayesg{1}} & $0.854 \pm {\scriptstyle 0.136}$ & $0.854 \pm {\scriptstyle 0.136}$ \\
& 10 & & $0.856 \pm {\scriptstyle 0.062}$ & $0.856 \pm {\scriptstyle 0.062}$ \\
& 15 & & $0.858 \pm {\scriptstyle 0.032}$ & $0.858 \pm {\scriptstyle 0.032}$ \\
\addlinespace
\multirow{3}{*}{Combined} & 5 & \multirow{3}{*}{\bayes{1} } & $0.863 \pm {\scriptstyle 0.084}$ & $0.863 \pm {\scriptstyle 0.084}$ \\
& 10 & & $0.866 \pm {\scriptstyle 0.042}$ & $0.866 \pm {\scriptstyle 0.042}$ \\
& 15 & & $0.864 \pm {\scriptstyle 0.023}$ & $0.864 \pm {\scriptstyle 0.023}$ \\
\bottomrule
\end{tabular}
\end{table}

%% file: Tables/categorical_combined.tex
\begin{table}[t]
\centering
\footnotesize
\setlength{\tabcolsep}{2pt}
\caption{Categorical ranking at $N\!=\!1$ on the combined benchmark ($L\!=\!11$, the first 11 models from \cref{app:tab:modelid}, $M\!=\!120$). Eight representative schemes are ordered by agreement with the gold standard ($\tau_{\text{GS}}$, vs.\ \bayesu{80}). Self: $\tau_b$ vs.\ Scheme@$80$; Greedy: $\tau_b$ vs.\ \bayesg{80}. Values are mean $\pm$ std over $80$ draws.}
\label{tab:cat_combined}
\begin{tabular}{@{}lccc@{}}
\toprule
Scheme & $\tau_{\text{GS}}$ & $\tau_{\text{Self}}$ & $\tau_{\text{Greedy}}$ \\
\midrule
Conservative     & $0.856 \pm {\scriptstyle 0.076}$ & $0.861 \pm {\scriptstyle 0.066}$ & $0.858 \pm {\scriptstyle 0.074}$ \\
Efficiency-adj.  & $0.850 \pm {\scriptstyle 0.070}$ & $0.875 \pm {\scriptstyle 0.057}$ & $0.859 \pm {\scriptstyle 0.071}$ \\
Format-aware     & $0.849 \pm {\scriptstyle 0.071}$ & $0.881 \pm {\scriptstyle 0.064}$ & $0.869 \pm {\scriptstyle 0.069}$ \\
Balanced comp.   & $0.843 \pm {\scriptstyle 0.075}$ & $0.877 \pm {\scriptstyle 0.067}$ & $0.862 \pm {\scriptstyle 0.073}$ \\
OOD-robust       & $0.840 \pm {\scriptstyle 0.071}$ & $0.892 \pm {\scriptstyle 0.063}$ & $0.870 \pm {\scriptstyle 0.066}$ \\
Rare-event       & $0.838 \pm {\scriptstyle 0.073}$ & $0.888 \pm {\scriptstyle 0.065}$ & $0.867 \pm {\scriptstyle 0.069}$ \\
Verifier-calib.  & $0.832 \pm {\scriptstyle 0.076}$ & $0.877 \pm {\scriptstyle 0.067}$ & $0.855 \pm {\scriptstyle 0.073}$ \\
Verifier-only    & $0.824 \pm {\scriptstyle 0.071}$ & $0.897 \pm {\scriptstyle 0.068}$ & $0.870 \pm {\scriptstyle 0.071}$ \\
\bottomrule
\end{tabular}
\end{table}

%% file: main/related_workds.tex
\section{Related Work}\label{sec:related_work}

\paragraph{Test-time scaling and stochastic reasoning.}
Test-time scaling samples multiple solutions per prompt and aggregates them \cite{Wang2022SelfConsistency,Snell2024TestTimeCompute,Zeng2025Revisiting}. Because stochastic reasoning varies across runs \cite{liu2024stable_reasoning}, we study how this variability affects \emph{rankings} under different aggregation rules as the test-time budget changes.

\paragraph{Ranking and statistical modeling for LLM evaluation.}
Preference evaluation and alignment learn from paired comparisons \cite{christiano2017deep,rafailov2023direct} and underpin leaderboards such as Chatbot Arena \cite{chiang2024chatbotarena,ameli2024statisticalframework}. Benchmark leaderboards often rank models by task metrics such as \pass{k} \cite{chen2021evaluating}, and recent work adds Bayesian uncertainty and IRT-style modeling \cite{hariri2025dontpasskbayesianframework,zhou2025lost}. We extend this literature to dense repeated-trial benchmarks and compare ranking methods through stability and convergence; \cref{sec:app:extended_related_work} gives additional background.

%% file: main/conclusion.tex
\section{Conclusion \& Future Directions}\label{sec:conclusion}
Test-time scaling turns LLM benchmarking into a repeated-sampling problem, so model rankings must be estimated from stochastic trials rather than from a single run. We formalize this setting and compare a broad collection of ranking methods within a common framework. When many trials are available, most reasonable ranking families induce nearly identical orderings, making \bayesu{N} a simple and interpretable default. The main differences appear in the low-budget regime. There, uncertainty-aware estimators can improve stability, and the greedy prior \bayesg{N} acts as a shrinkage estimator: it reduces variance when greedy and stochastic sampling align, but can bias rankings when they diverge.

In practice, \bayesu{N} is a strong default, whereas \bayesg{N} is best used after checking greedy--sampling alignment on a small pilot sample. Our experiments focus on binary correctness; extending the analysis to partial credit, rubric-based scoring, and other categorical evaluation settings is a natural next step.

%% file: main/limit.tex
\section*{Limitations}

Our experiments focus on mathematical reasoning benchmarks. We do not evaluate partial credit, or open-ended outputs, where outcome categories are less clear and annotation or verification noise may be larger. More generally, when informative priors are used---especially priors derived from auxiliary signals other than greedy decoding---the prior source and specification should be reported explicitly, since the prior can introduce systematic bias if it is misaligned with the stochastic evaluation regime.

%% file: main/acknowledge.tex
\section*{Acknowledgments}
This research was supported in part by NSF awards 2117439 and 2320952.

%% file: appendix/notations.tex
\section{Notation and Definitions}\label{sec:app:notations}

Throughout the paper, we use the following notation.

\subsection{Data and Basic Quantities}

\begin{itemize}[leftmargin=*]
\item $L$: number of models being ranked.
\item $M$: number of questions in a benchmark.
\item $N$: number of independent stochastic trials per model--question pair under test-time scaling.
\item $\mathbf{R}\in\{0,1\}^{L\times M\times N}$: response tensor, where $R_{lmn}=1$ if model $l$ solves question $m$ on trial $n$.
\item $\mathbf{R}_0$: optional prior outcomes used by Bayesian estimators. In this paper, greedy decoding yields a shared prior matrix $\mathbf{R}_0\in\{0,1\}^{M\times D}$ with $D=1$, but the notation can also accommodate model-specific prior tensors.
\item $\widehat{p}_{lm} := \frac{1}{N}\sum_{n=1}^N R_{lmn}$: per-question solve rate for model $l$ on question $m$.
\item $k_{lm} := \sum_{n=1}^N R_{lmn}$: number of successful trials for model $l$ on question $m$.
\end{itemize}

\subsection{Metric Shorthand}

\begin{itemize}[leftmargin=*]
\item \textbf{\bayes{N}}: Bayesian posterior-mean estimate at $N$ trials under a specified prior.
\item \textbf{\bayesu{N}}: Bayesian estimate with a uniform Dirichlet prior, denoted $\mathrm{Bayes}_{\mathcal{U}}@N$.
\item \textbf{\bayesg{N}}: Bayesian estimate with a greedy empirical prior, denoted $\mathrm{Bayes}_{\mathbf{R}_0}@N$.
\item \textbf{\pass{k}}: probability that at least one of $k$ sampled completions is correct.
\item \textbf{\avg{N}}: mean accuracy over all $M$ questions and $N$ trials. For binary outcomes, it is order-equivalent to \bayesu{N}.
\end{itemize}

\subsection{Ranking-Method Families}

\begin{itemize}[leftmargin=*]
\item \textbf{Pointwise methods}: aggregate per-question performance to produce model scores (e.g., mean accuracy, inverse-difficulty weighting).
\item \textbf{Pairwise methods}: transform outcomes into win/tie counts between model pairs and fit paired-comparison models (e.g., Bradley--Terry, Elo, Glicko).
\item \textbf{Listwise, setwise methods}: operate on winner and loser sets for each question--trial (e.g., Plackett--Luce, Davidson--Luce).
\item \textbf{Voting rules}: treat questions as voters that rank models and then aggregate those preferences (e.g., Borda, Copeland, Schulze, Kemeny--Young).
\item \textbf{Graph/spectral methods}: construct comparison graphs and compute centrality- or flow-based scores (e.g., PageRank, Rank Centrality, HodgeRank, $\alpha$-Rank).
\item \textbf{IRT-inspired methods}: estimate latent model abilities and item difficulties (e.g., Rasch, 2PL, 3PL, dynamic IRT).
\end{itemize}

\subsection{Evaluation Criteria}

\begin{itemize}[leftmargin=*]
\item \textbf{Kendall's $\tau_b$}: rank-correlation coefficient that accounts for ties; it ranges from $-1$ (perfect disagreement) to $+1$ (perfect agreement).
\item \textbf{Gold-standard agreement}: agreement between a low-budget ranking and the empirical gold standard, typically \bayesu{80} in this paper.
\item \textbf{Self-consistency}: agreement between a low-budget ranking and the same method's all-trial ranking.
\item \textbf{Convergence}: the rate at which a method's ranking approaches its full-trial ordering as the number of trials increases.
\item \textbf{Greedy--sampling alignment} ($\tau_{\text{G-S}}$): Kendall's $\tau_b$ between the ranking induced by greedy decoding and the ranking induced by stochastic sampling at high budget.
\end{itemize}

\subsection{Inference Terminology}

\begin{itemize}[leftmargin=*]
\item \textbf{MLE} (maximum likelihood estimation): point estimate that maximizes $p(\mathbf{R}\mid\theta)$.
\item \textbf{MAP} (maximum a posteriori): point estimate that maximizes $p(\mathbf{R}\mid\theta)p(\theta)$.
\item \textbf{EAP} (expected a posteriori): posterior mean estimate $\mathbb{E}[\theta\mid\mathbf{R}]$.
\item \textbf{MML} (marginal maximum likelihood): likelihood-based estimation that integrates over a latent population distribution, commonly used in IRT.
\item \textbf{Credible intervals (CrI)}: Bayesian posterior intervals used for uncertainty quantification; we use lower credible bounds (LCBs) for conservative ranking.
\end{itemize}

%% file: appendix/accuracy.tex
\section{Accuracy of Models}\label{sec:app:accuracy}

\begin{figure*}[!ht]
  \centering
  \includegraphics[width=\textwidth]{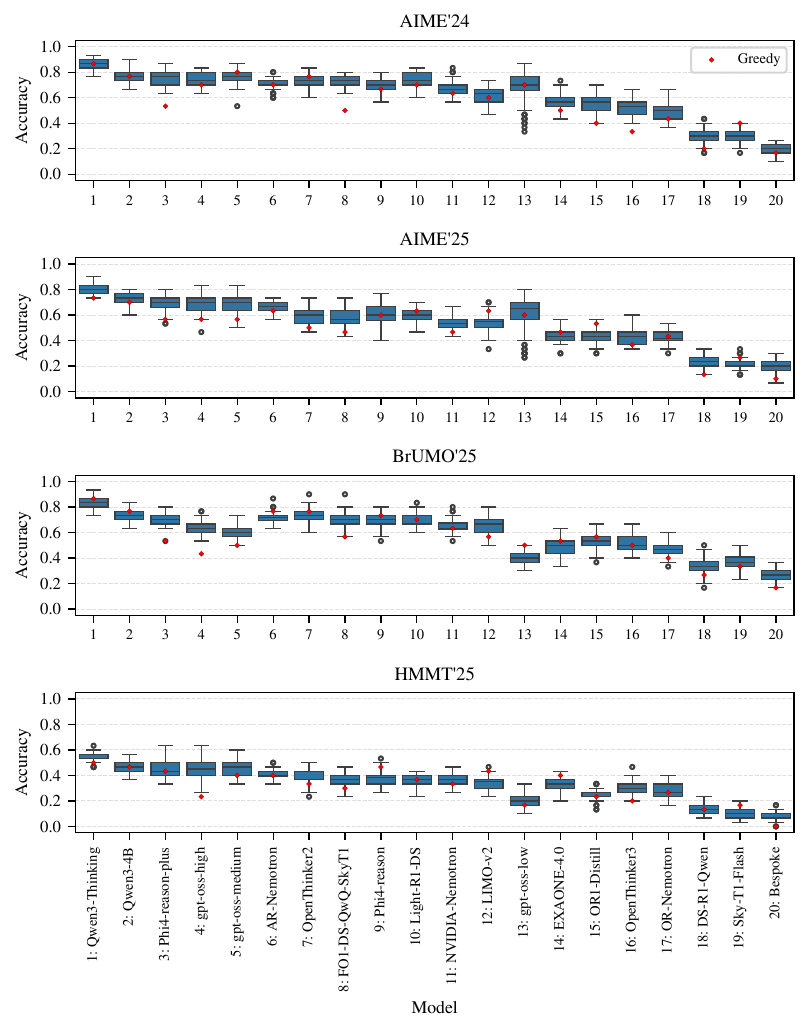}
  \caption{Overview of model accuracies across all four benchmarks. Each panel shows each model's mean accuracy under stochastic sampling (over $N=80$ trials), together with greedy accuracy (markers). Error bars denote one standard deviation across trials and illustrate the variability introduced by test-time scaling. Models are color-coded consistently across benchmarks for ease of comparison. The figure shows substantial heterogeneity in both absolute performance and sampling variance, with \hmmt{} notably harder than the other three benchmarks.}
  \label{app:fig:accuracy_overview}
\end{figure*}

\Cref{tab:aime24,tab:aime25,tab:hmmt_feb_2025,tab:brumo_2025} report detailed accuracy statistics for all $L=20$ models, including greedy accuracy and stochastic-sampling statistics (minimum, mean, maximum, and standard deviation) over $N=80$ trials. \hmmt{} is the hardest benchmark (mean accuracies $0.080$--$0.554$), while \aimefour{} and \brumo{} are relatively easy. \Cref{app:fig:accuracy_overview} visualizes these distributions across benchmarks and highlights the heterogeneity in model performance and sampling variance that motivates our ranking-stability analysis.

\input{Tables/accuracy.tex}

%% file: Tables/accuracy.tex
\begingroup
\setlength{\tabcolsep}{2pt}
\renewcommand{\arraystretch}{1.2}
\newcommand{\accuracytable}[3]{%
  \begin{minipage}[t]{0.491\textwidth}
    \centering
    \captionsetup{type=table}
    \caption{#1}
    \label{#2}
    \vspace{-0.5em}
    \resizebox{\linewidth}{!}{%
    \begin{tabular}{lccccc}
      \toprule
      Model & Greedy & \multicolumn{4}{c}{Top-$p$} \\
      \cmidrule(lr){3-6}
             & Acc. & Min & Mean & Max & Std \\
      \midrule
      #3
      \bottomrule
    \end{tabular}
    }
  \end{minipage}
}

\begin{table*}[t!]
\centering
\accuracytable{Accuracy on AIME'24.}{tab:aime24}{
\dsicon~DS-R1-Qwen & 0.200 & 0.167 & 0.297 & 0.433 & 0.055 \\
\gairicon~LIMO-v2 & 0.600 & 0.467 & 0.619 & 0.733 & 0.059 \\
\openicon~OpenThinker2 & 0.767 & 0.600 & 0.722 & 0.833 & 0.048 \\
\openicon~OpenThinker3 & 0.333 & 0.400 & 0.517 & 0.667 & 0.059 \\
\qwenicon~Qwen3-Thinking & 0.867 & 0.767 & 0.875 & 0.933 & 0.038 \\
\skyicon~Sky-T1-Flash & 0.400 & 0.167 & 0.310 & 0.400 & 0.050 \\
\gpticon~gpt-oss-high & 0.700 & 0.633 & 0.747 & 0.833 & 0.053 \\
\gpticon~gpt-oss-low & 0.700 & 0.333 & 0.675 & 0.867 & 0.130 \\
\gpticon~gpt-oss-medium & 0.800 & 0.533 & 0.755 & 0.867 & 0.054 \\
\lgicon~EXAONE-4.0 & 0.500 & 0.433 & 0.570 & 0.733 & 0.055 \\
\nvidiaicon~OR-Nemotron & 0.433 & 0.367 & 0.490 & 0.667 & 0.064 \\
\microsoft~Phi-4 & 0.667 & 0.567 & 0.705 & 0.800 & 0.050 \\
\microsoft~Phi-4-plus & 0.533 & 0.633 & 0.753 & 0.867 & 0.049 \\
\openr~OR1-Distill & 0.400 & 0.400 & 0.547 & 0.700 & 0.066 \\
\fuse~FuseO1-DS-QwQ-SkyT1 & 0.500 & 0.633 & 0.728 & 0.800 & 0.042 \\
\qihoo~Light-R1-DS & 0.700 & 0.600 & 0.734 & 0.833 & 0.060 \\
\nvidiaicon~AR-Nemotron & 0.700 & 0.600 & 0.709 & 0.800 & 0.043 \\
\nvidiaicon~NVIDIA-Nemotron & 0.633 & 0.567 & 0.676 & 0.833 & 0.059 \\
\qwenicon~Qwen3-4B & 0.767 & 0.667 & 0.772 & 0.900 & 0.052 \\
\bespoke~Bespoke & 0.167 & 0.100 & 0.197 & 0.267 & 0.043 \\
}
\hfill
\accuracytable{Accuracy on HMMT'25.}{tab:hmmt_feb_2025}{
\dsicon~DS-R1-Qwen & 0.133 & 0.067 & 0.135 & 0.233 & 0.040 \\
\gairicon~LIMO-v2 & 0.433 & 0.233 & 0.347 & 0.467 & 0.048 \\
\openicon~OpenThinker2 & 0.333 & 0.233 & 0.382 & 0.500 & 0.057 \\
\openicon~OpenThinker3 & 0.200 & 0.200 & 0.297 & 0.467 & 0.047 \\
\qwenicon~Qwen3-Thinking & 0.500 & 0.467 & 0.554 & 0.633 & 0.037 \\
\skyicon~Sky-T1-Flash & 0.167 & 0.033 & 0.106 & 0.200 & 0.034 \\
\gpticon~gpt-oss-high & 0.233 & 0.267 & 0.449 & 0.633 & 0.069 \\
\gpticon~gpt-oss-low & 0.167 & 0.100 & 0.203 & 0.333 & 0.051 \\
\gpticon~gpt-oss-medium & 0.400 & 0.333 & 0.455 & 0.600 & 0.056 \\
\lgicon~EXAONE-4.0 & 0.400 & 0.200 & 0.335 & 0.433 & 0.060 \\
\nvidiaicon~OR-Nemotron & 0.267 & 0.167 & 0.283 & 0.400 & 0.049 \\
\microsoft~Phi-4 & 0.467 & 0.267 & 0.378 & 0.533 & 0.056 \\
\microsoft~Phi-4-plus & 0.433 & 0.333 & 0.447 & 0.633 & 0.056 \\
\openr~OR1-Distill & 0.233 & 0.133 & 0.251 & 0.333 & 0.042 \\
\fuse~FuseO1-DS-QwQ-SkyT1 & 0.300 & 0.233 & 0.363 & 0.467 & 0.045 \\
\qihoo~Light-R1-DS & 0.367 & 0.233 & 0.356 & 0.433 & 0.045 \\
\nvidiaicon~AR-Nemotron & 0.400 & 0.333 & 0.408 & 0.500 & 0.042 \\
\nvidiaicon~NVIDIA-Nemotron & 0.333 & 0.267 & 0.362 & 0.467 & 0.048 \\
\qwenicon~Qwen3-4B & 0.467 & 0.367 & 0.464 & 0.567 & 0.046 \\
\bespoke~Bespoke & 0.000 & 0.000 & 0.080 & 0.167 & 0.035 \\
}

\vspace{3em}

\accuracytable{Accuracy on AIME'25.}{tab:aime25}{
\dsicon~DS-R1-Qwen & 0.133 & 0.133 & 0.236 & 0.333 & 0.046 \\
\gairicon~LIMO-v2 & 0.633 & 0.333 & 0.541 & 0.700 & 0.068 \\
\openicon~OpenThinker2 & 0.500 & 0.467 & 0.595 & 0.733 & 0.060 \\
\openicon~OpenThinker3 & 0.367 & 0.333 & 0.425 & 0.600 & 0.057 \\
\qwenicon~Qwen3-Thinking & 0.733 & 0.733 & 0.804 & 0.900 & 0.037 \\
\skyicon~Sky-T1-Flash & 0.267 & 0.133 & 0.220 & 0.333 & 0.041 \\
\gpticon~gpt-oss-high & 0.567 & 0.467 & 0.690 & 0.833 & 0.063 \\
\gpticon~gpt-oss-low & 0.600 & 0.267 & 0.598 & 0.800 & 0.145 \\
\gpticon~gpt-oss-medium & 0.567 & 0.500 & 0.689 & 0.833 & 0.065 \\
\lgicon~EXAONE-4.0 & 0.467 & 0.300 & 0.441 & 0.567 & 0.054 \\
\nvidiaicon~OR-Nemotron & 0.433 & 0.300 & 0.425 & 0.533 & 0.054 \\
\microsoft~Phi-4 & 0.600 & 0.400 & 0.599 & 0.767 & 0.072 \\
\microsoft~Phi-4-plus & 0.567 & 0.533 & 0.683 & 0.800 & 0.058 \\
\openr~OR1-Distill & 0.533 & 0.300 & 0.426 & 0.567 & 0.059 \\
\fuse~FuseO1-DS-QwQ-SkyT1 & 0.467 & 0.433 & 0.585 & 0.733 & 0.064 \\
\qihoo~Light-R1-DS & 0.633 & 0.467 & 0.589 & 0.700 & 0.056 \\
\nvidiaicon~AR-Nemotron & 0.633 & 0.567 & 0.651 & 0.733 & 0.045 \\
\nvidiaicon~NVIDIA-Nemotron & 0.467 & 0.433 & 0.546 & 0.667 & 0.050 \\
\qwenicon~Qwen3-4B & 0.700 & 0.600 & 0.729 & 0.800 & 0.044 \\
\bespoke~Bespoke & 0.100 & 0.067 & 0.193 & 0.300 & 0.050 \\
}
\hfill
\accuracytable{Accuracy on BrUMO'25.}{tab:brumo_2025}{
\dsicon~DS-R1-Qwen & 0.267 & 0.167 & 0.344 & 0.500 & 0.062 \\
\gairicon~LIMO-v2 & 0.567 & 0.500 & 0.651 & 0.800 & 0.065 \\
\openicon~OpenThinker2 & 0.767 & 0.600 & 0.738 & 0.900 & 0.061 \\
\openicon~OpenThinker3 & 0.500 & 0.400 & 0.512 & 0.667 & 0.055 \\
\qwenicon~Qwen3-Thinking & 0.867 & 0.733 & 0.838 & 0.933 & 0.038 \\
\skyicon~Sky-T1-Flash & 0.333 & 0.233 & 0.372 & 0.500 & 0.059 \\
\gpticon~gpt-oss-high & 0.433 & 0.533 & 0.628 & 0.767 & 0.053 \\
\gpticon~gpt-oss-low & 0.500 & 0.300 & 0.393 & 0.500 & 0.053 \\
\gpticon~gpt-oss-medium & 0.500 & 0.500 & 0.610 & 0.733 & 0.052 \\
\lgicon~EXAONE-4.0 & 0.533 & 0.333 & 0.484 & 0.633 & 0.059 \\
\nvidiaicon~OR-Nemotron & 0.400 & 0.333 & 0.469 & 0.600 & 0.054 \\
\microsoft~Phi-4 & 0.733 & 0.533 & 0.692 & 0.800 & 0.052 \\
\microsoft~Phi-4-plus & 0.533 & 0.533 & 0.711 & 0.800 & 0.048 \\
\openr~OR1-Distill & 0.567 & 0.367 & 0.538 & 0.667 & 0.057 \\
\fuse~FuseO1-DS-QwQ-SkyT1 & 0.567 & 0.567 & 0.710 & 0.900 & 0.056 \\
\qihoo~Light-R1-DS & 0.700 & 0.600 & 0.690 & 0.833 & 0.049 \\
\nvidiaicon~AR-Nemotron & 0.767 & 0.633 & 0.714 & 0.867 & 0.044 \\
\nvidiaicon~NVIDIA-Nemotron & 0.633 & 0.533 & 0.649 & 0.800 & 0.048 \\
\qwenicon~Qwen3-4B & 0.767 & 0.633 & 0.744 & 0.833 & 0.049 \\
\bespoke~Bespoke & 0.167 & 0.167 & 0.265 & 0.367 & 0.053 \\
}
\end{table*}
\endgroup

%% file: appendix/gold_standard.tex
\section{Gold Standard Agreement}\label{sec:app:gold_standard_agreement}

To justify our use of \bayesu{80} as the gold standard, we compare the full-trial rankings produced by all methods at $N=80$. \cref{tab:gold_standard_agreement} summarizes Kendall's $\tau_b$ between \bayesu{80} and each competing method. The full results show that \bayesu{80} is also a high-consensus ordering: by average agreement with all other methods, it ranks first on \aimefive{}, \hmmt{}, and the Combined benchmark and second on \aimefour{} and \brumo{} within $5\times 10^{-4}$ of the best (\cref{tab:gold_standard_consensus}). Dataset-level consensus tables appear in \cref{tab:aime24_kendalltau,tab:aime25_kendalltau,tab:hmmt_kendalltau,tab:brumo_kendalltau,tab:combined_kendalltau}. Many methods recover the same ordering exactly (\cref{tab:gold_standard_exact_match}), and the remaining disagreement is concentrated in a small low-agreement tail (\cref{tab:gold_standard_low_agreement}).

\input{Tables/gold_standard_consensus.tex}
\input{Tables/gold_standard_low_agreement.tex}
\input{Tables/aime24_kendalltau.tex}

\input{Tables/aime25_kendalltau.tex}
\input{Tables/hmmt_kendalltau.tex}
\input{Tables/brumo_kendalltau.tex}
\input{Tables/combined_kendalltau.tex}
\input{Tables/gold_standard_exact_match.tex}

\subsection{Convergence of Ranking Methods}\label{app:ssub:nonconvergence}

As the number of trials $N$ (or questions $M$) increases, evaluation metrics such as \avg{N}, \bayes{N}, and \pass{k} need not induce the same limiting ordering as ranking methods. The reason is that they target different population quantities.

We make this distinction explicit below for the two canonical choices used throughout the paper: the average-accuracy ranking and the Bradley--Terry (BT) model.

\subsection{Large-Budget Limits: Each Method Converges, but Generally to a Different Target}
\label{app:avg-bt-limits}

To discuss $M\to\infty$ (or $N\to\infty$) formally, we introduce an i.i.d.\ sampling model at the level of question--trial pairs.
Assume $(X_{mn})_{m\in[M],n\in[N]}$ are i.i.d.\ draws from some distribution $P$ on $\{0,1\}^L$.
Let
\begin{align*}
p_\ell &:= \mathbb{P}_{X\sim P}(X_\ell=1) \\
w_{ij} &:= \mathbb{P}_{X\sim P}(X_i=1,\ X_j=0).
\end{align*}
(For clarity: $p_\ell$ depends only on the marginal of model $\ell$, whereas $w_{ij}$ depends on the \emph{joint} distribution of $(X_i,X_j)$.)

\paragraph{Average targets marginal accuracy.}
By the law of large numbers,
\[
\widehat{p}^{\text{avg}}_\ell(R)
\;\xrightarrow[MN\to\infty]{\text{a.s.}}\;
p_\ell.
\]
Likewise, \bayesu{N} converges to the same $p_\ell$; for binary outcomes it differs from $\widehat{p}^{\text{avg}}_\ell$ only by $O((MN)^{-1})$ smoothing.

\paragraph{Bradley--Terry targets a pairwise decisive-win functional.}
The empirical win frequencies converge:
\[
\frac{1}{MN}W_{ij}(R)
\;\xrightarrow[MN\to\infty]{\text{a.s.}}\;
w_{ij}.
\]
Define the BT log-likelihood
\begin{equation}
\ell(\pi;W) := \sum_{i\neq j} W_{ij}\Bigl(\log \pi_i - \log(\pi_i+\pi_j)\Bigr).
\label{eq:bt-loglik}
\end{equation}
Then the BT-ML estimator is an $M$-estimator: maximizing \eqref{eq:bt-loglik} with $W_{ij}$ is equivalent to maximizing the scaled objective $(MN)^{-1}\ell(\pi;W)$.
Under mild regularity and connectivity conditions (ensuring strict concavity in $\log\pi$ and uniqueness up to scale), $\widehat{\pi}$ converges to the unique (up to scale) maximizer of the \emph{population} objective
\begin{equation}
\pi^\star
\in
\arg\max_{\pi>0}
\sum_{i\neq j} w_{ij}\Bigl(\log \pi_i - \log(\pi_i+\pi_j)\Bigr).
\label{eq:bt-pop}
\end{equation}

The limiting objects $(p_\ell)_{\ell=1}^L$ and $\pi^\star$ are generally \emph{not} linked by any monotone transform:
$p_\ell$ depends only on marginal correctness, while $\pi^\star$ depends on the full matrix $(w_{ij})_{i\neq j}$.
Therefore, without additional assumptions on $P$ (e.g., that $P$ is generated by a BT choice model at the level of decisive comparisons), there is no reason to expect the induced orderings to coincide as $MN\to\infty$.
We make this non-equivalence explicit with a counterexample.

\paragraph{A Counterexample: Average accuracy and Bradley--Terry disagree even at infinite budget}
\label{app:avg-bt-counterexample}

We construct a distribution $P$ (equivalently, a finite pattern that can be repeated) for which the average ranking and the BT-ML ranking disagree.
The construction uses $L=3$ models.
For notational convenience, we label them $0,1,2$.

\paragraph{Outcome patterns.}
Consider the following three outcome vectors in $\{0,1\}^3$:
\begin{align*}
\text{Type A: } &(0,1,1), \\
\text{Type B: } &(1,0,0), \\
\text{Type C: } &(1,1,0).
\end{align*}
Let $P$ place mass
\[
\mathbb{P}(\text{A})=\frac{2}{8},\qquad
\mathbb{P}(\text{B})=\frac{3}{8},\qquad
\mathbb{P}(\text{C})=\frac{3}{8}.
\]
Equivalently, one may take a deterministic dataset with $M=8$ questions and $N=1$ trial, containing exactly $2$ questions of Type A, $3$ of Type B, and $3$ of Type C; repeating this block preserves both rankings, as shown below.

The marginal success probabilities are
\[
p_0 = \frac{6}{8}=\frac34,\qquad
p_1 = \frac{5}{8},\qquad
p_2 = \frac{2}{8}=\frac14,
\]
so the average method ranks
\[
0 \;>\; 1 \;>\; 2.
\]

From the three types above, the decisive-win probabilities $w_{ij}=\mathbb{P}(X_i=1,X_j=0)$ are:
\[
\begin{aligned}
w_{01}&=\frac{3}{8},\quad w_{10}=\frac{2}{8},\\
w_{02}&=\frac{6}{8},\quad w_{20}=\frac{2}{8},\\
w_{12}&=\frac{3}{8},\quad w_{21}=0.
\end{aligned}
\]
For the finite $M=8,N=1$ realization, the corresponding win counts are simply $W_{ij}=8w_{ij}$, i.e.,
\begin{equation}
W
=
\begin{pmatrix}
0 & 3 & 6 \\
2 & 0 & 3 \\
2 & 0 & 0
\end{pmatrix}.
\label{eq:W-matrix}
\end{equation}

We now show that BT-ML ranks $1>0>2$ for \eqref{eq:W-matrix}, thereby disagreeing with the average ranking.

A convenient characterization of the BT-ML optimum is the standard first-order condition equating observed wins to model-implied expected wins:
for each $i$,
\begin{equation}
\sum_{j\neq i} W_{ij}
\;=\;
\sum_{j\neq i} (W_{ij}+W_{ji})\cdot \frac{\pi_i}{\pi_i+\pi_j}.
\label{eq:bt-foc}
\end{equation}
(These equations follow by differentiating \eqref{eq:bt-loglik} with respect to $\log\pi_i$.)

Because BT strengths are identifiable only up to a global scale factor, fix $\pi_2=1$ and write $\pi_0=a$, $\pi_1=b$.
Plugging \eqref{eq:W-matrix} into \eqref{eq:bt-foc} yields two independent equations:
\begin{align}
9
&=
5\cdot \frac{a}{a+b} \;+\; 8\cdot \frac{a}{a+1},
\label{eq:foc-a}\\
5
&=
5\cdot \frac{b}{a+b} \;+\; 3\cdot \frac{b}{b+1}.
\label{eq:foc-b}
\end{align}

\paragraph{Step 1: solve $a$ in terms of $b$.}
From \eqref{eq:foc-b},
\[
5 - 3\cdot \frac{b}{b+1} = 5\cdot \frac{b}{a+b}.
\]
The left-hand side simplifies:
\[
5 - 3\cdot\frac{b}{b+1}
=
\frac{5(b+1)-3b}{b+1}
=
\frac{2b+5}{b+1}.
\]
Thus,
\begin{align}
\frac{5b}{a+b} &= \frac{2b+5}{b+1} \\
&\quad\Longrightarrow\quad a+b = \frac{5b(b+1)}{2b+5} \\
&\quad\Longrightarrow\quad a = \frac{3b^2}{2b+5}.
\label{eq:a-in-terms-of-b}
\end{align}

\paragraph{Step 2: determine $b$ from a one-dimensional equation.}
Substitute \eqref{eq:a-in-terms-of-b} into \eqref{eq:foc-a}.
First note that
\[
\frac{a}{a+b}
=
\frac{\frac{3b^2}{2b+5}}{\frac{5b(b+1)}{2b+5}}
=
\frac{3b}{5(b+1)},
\]
so the first term in \eqref{eq:foc-a} becomes $5\cdot \frac{a}{a+b} = \frac{3b}{b+1}$.
Also,
\[
\frac{a}{a+1}
=
\frac{\frac{3b^2}{2b+5}}{\frac{3b^2}{2b+5}+1}
=
\frac{3b^2}{3b^2+2b+5}.
\]
Therefore \eqref{eq:foc-a} is equivalent to
\[
9
=
\frac{3b}{b+1}
+
8\cdot \frac{3b^2}{3b^2+2b+5}.
\]
A short algebraic manipulation gives the cubic equation
\begin{equation}
2b^3 - 5b^2 - 16b - 15 = 0.
\label{eq:cubic}
\end{equation}
Let $f(b)=2b^3 - 5b^2 - 16b - 15$.
We have
\begin{align*}
f(4) &= 128-80-64-15 = -31 < 0, \\
f(5) &= 250-125-80-15 = 30 > 0,
\end{align*}
so there exists a root $b^\star\in(4,5)$.
Moreover,
\[
f'(b)=6b^2-10b-16 = 2(3b^2-5b-8),
\]
whose positive root is $b=\frac{5+\sqrt{121}}{6}=\frac{16}{6}=\frac{8}{3}$.
Hence $f$ is strictly increasing for all $b>\frac{8}{3}$, implying the root $b^\star\in(4,5)$ is unique.
We therefore conclude that the BT-ML solution (under $\pi_2=1$) satisfies $b=b^\star\in(4,5)$ and $a=\frac{3(b^\star)^2}{2b^\star+5}$.

\paragraph{Step 3: show $b>a>1$, hence BT ranks $1>0>2$.}
Using \eqref{eq:a-in-terms-of-b},
\[
\frac{b}{a}
=
\frac{b}{\frac{3b^2}{2b+5}}
=
\frac{2b+5}{3b}.
\]
Thus, $b>a$ holds exactly when $(2b+5)/(3b)>1$, i.e., when $b<5$.
Since $b^\star\in(4,5)$, we have $b^\star>a$.

It remains to show $a>1$.
Suppose for contradiction that $a\le 1$.
We have already established $b>a$, so $b\ge a$.
Then $\frac{a}{a+b}\le \frac{a}{2a}=\frac12$ and $\frac{a}{a+1}\le \frac12$.
Plugging into \eqref{eq:foc-a} gives
\[
9
=
5\cdot \frac{a}{a+b} + 8\cdot \frac{a}{a+1}
\le
5\cdot\frac12 + 8\cdot\frac12
=
\frac{13}{2}
< 9,
\]
a contradiction.
Hence $a>1=\pi_2$.
Putting these inequalities together yields
\[
\pi_1=b^\star \;>\; \pi_0=a \;>\; \pi_2=1,
\]
so BT ranks
\[
1 \;>\; 0 \;>\; 2.
\]
This contradicts the average ranking $0>1>2$, establishing that the two methods can induce different orderings even in the absence of sampling noise.

\paragraph{From a Finite Counterexample to ``No Convergence'' as $M\to\infty$ or $N\to\infty$.}
\label{app:par:avg-bt-replication}

The counterexample already rules out a general theorem forcing average and BT rankings to coincide in the large-budget limit. To connect it directly to $M\to\infty$ or $N\to\infty$, it suffices to note that both methods are invariant under replication.

\paragraph{Replication invariance (deterministic construction).}
Let $R$ be any fixed tensor.
For an integer $k\ge 1$, define:
(i) \emph{question replication} $R^{(k,M)}$ by repeating the $M$ questions $k$ times (so $M' = kM$ and $N' = N$), and
(ii) \emph{trial replication} $R^{(k,N)}$ by repeating the $N$ trials $k$ times (so $M'=M$ and $N' = kN$).
Then:
\begin{enumerate}
\item Average scores are unchanged:
\[
\widehat{p}^{\text{avg}}_\ell(R^{(k,M)})=\widehat{p}^{\text{avg}}_\ell(R^{(k,N)})=\widehat{p}^{\text{avg}}_\ell(R).
\]
\item The decisive-win matrix scales linearly:
\begin{align*}
W(R^{(k,M)}) &= k\,W(R) \\
W(R^{(k,N)}) &= k\,W(R).
\end{align*}
\item The BT-ML maximizer is unchanged, because the log-likelihood scales as
\[
\ell(\pi; kW) = k\,\ell(\pi;W),
\]
and therefore has the same maximizer.
\end{enumerate}
Therefore, if two methods disagree on $R$, they disagree on $R^{(k,M)}$ for arbitrarily large $M$ and on $R^{(k,N)}$ for arbitrarily large $N$.
Applied to the $M=8,N=1$ tensor corresponding to \eqref{eq:W-matrix}, this yields an explicit sequence with $M\to\infty$ (or $N\to\infty$) for which the average and BT rankings remain different at every budget.

\paragraph{Stochastic formulation (i.i.d.\ construction).}
Alternatively, under the i.i.d.\ model of \cref{app:avg-bt-limits}, the same discrepancy appears at the population level. For the distribution $P$ defined above, the limiting average ranking is determined by $(p_\ell)$ and yields $0>1>2$, while the limiting BT ranking is determined by the maximizer of \eqref{eq:bt-pop} and yields $1>0>2$. Thus, even with independent sampling and $MN\to\infty$, the two rankings can converge to different limits.

\subsection{Implications and Support for the Gold-Standard Definition}
\label{app:avg-bt-goldstandard}

The analysis above has a direct implication for benchmarking ranking methods: there is no method-independent guarantee that all reasonable procedures converge to the same ordering as the evaluation budget grows. Different ranking procedures correspond to different statistical targets.

\paragraph{Why this happens.}
Average-based ranking targets the marginal success probabilities $p_\ell=\mathbb{P}(X_\ell=1)$. BT instead targets the latent strengths that best explain the decisive pairwise win rates $w_{ij}=\mathbb{P}(X_i=1,X_j=0)$ through a logistic choice model. These are different summaries of the same joint outcome distribution $P$. The counterexample in \cref{app:avg-bt-counterexample} isolates the mechanism: a model can have higher marginal accuracy while assigning less decisive-win mass against another model, which shifts the BT optimum.

\paragraph{Why a gold standard is needed.}
Because ranking methods need not share a common asymptotic ordering, claims about ``distance to the truth'' require a specified target ordering. Otherwise even statements such as ``method $A$ converges faster than method $B$'' are ambiguous.

\paragraph{Our choice: \bayesu{N}.}
We define the gold-standard ordering as \bayesu{N} (with $N=80$ in our experiments).
This definition is supported by three considerations:
\begin{enumerate}
\item \textbf{Interpretability and decision relevance.}
\bayesu{N} estimates the probability that a model solves a randomly drawn benchmark item under the sampling policy.
This is an accuracy-like quantity with a direct operational meaning.
\item \textbf{Minimal modeling assumptions.}
\bayesu{N} (and \avg{N}) depend only on marginal correctness and do not impose a parametric pairwise-choice model.
Methods such as BT are useful when the pairwise-choice model is appropriate, but their induced ordering is not, in general, a refinement of accuracy.
\item \textbf{Consistency under increasing budget.}
Under i.i.d.\ sampling of $(m,n)$ pairs, \bayesu{N} converges to $p_\ell$ as $MN\to\infty$, making it a natural ``infinite-budget'' reference for accuracy-based evaluation.
\end{enumerate}

\paragraph{Relationship to self-consistency.}
This non-convergence result does \emph{not} argue against BT or other rankers. It instead clarifies that two evaluations are complementary: agreement with an explicit accuracy-based target, and \emph{self-consistency}, i.e., how quickly a method stabilizes toward its own full-budget ordering. The former asks whether a method matches the chosen reference; the latter asks how stable the method itself becomes as trials accumulate. The counterexample shows why these questions are not interchangeable.

\subsection{Minimality of the eight-question construction}
The counterexample in Section~\ref{app:avg-bt-counterexample} uses $M=8$ questions.
Here we record a minimality fact: under the same setting ($L=3$, $N=1$, and BT-ML fit from decisive wins), there is no \emph{strict} disagreement example with fewer than eight questions.

\paragraph{Proposition (minimal $M$ for strict disagreement; verified by exhaustive enumeration).}
Assume $L=3$ and $N=1$.
Assume moreover that the average ranking is strict (all three average scores are distinct), and that BT-ML is well-defined and finite (equivalently, the directed win graph with an edge $i\to j$ whenever $W_{ij}>0$ is strongly connected, which ensures a unique BT-ML maximizer up to global scale).
If BT-ML disagrees with the average ranking, then $M\ge 8$.

\paragraph{Verification.}
With $N=1$, each question produces an outcome pattern in $\{0,1\}^3$.
Hence, up to permutation of questions, any dataset with $M$ questions is determined by the count vector
$c=(c_x)_{x\in\{0,1\}^3}\in\mathbb{N}^8$ with $\sum_x c_x=M$.
For fixed $M$, there are $\binom{M+7}{7}$ such vectors; thus the total number of datasets with $M\le 7$ is
\[
\sum_{M=1}^7 \binom{M+7}{7} \;=\; 6434.
\]
For each such dataset, we compute the induced average ordering and the BT-ML ordering (obtained by maximizing \eqref{eq:bt-loglik}, equivalently solving \eqref{eq:bt-foc}).
Restricting to datasets with (i) strict average ordering and (ii) strong connectivity (so the BT-ML maximizer is unique up to scale), an exhaustive enumeration yields $1506$ instances for $M\le 7$; in all of them the BT-ML ordering agrees with the average ordering.
Therefore, no strict-disagreement example exists for $M\le 7$.

Finally, Section~\ref{app:avg-bt-counterexample} exhibits a strict-disagreement dataset at $M=8$, so $M_{\min}=8$.

%% file: Tables/gold_standard_consensus.tex
\begin{table}[t]
\centering
\small
\setlength{\tabcolsep}{4pt}
\caption{\bayesu{80} as a consensus ranking. ``Consensus rank'' sorts methods by their average Kendall's $\tau_b$ agreement with all other methods (computed at $N=80$; ties broken by lower std).}
\label{tab:gold_standard_consensus}
\resizebox{\columnwidth}{!}{%
\begin{tabular}{lccccc}
\toprule
Benchmark & Mean rank & \bayesu{80} avg. & Best method & Best avg. & Gap \\
\midrule
\aimefour & 2 & 0.9414 & \texttt{rasch\_mml} & 0.9417 & 0.0003 \\
\aimefive & 1 & 0.9344 & \texttt{avg} (tie) & 0.9344 & 0.0000 \\
\hmmt & 1 & 0.9499 & \texttt{avg} (tie) & 0.9499 & 0.0000 \\
\brumo & 2 & 0.9542 & \texttt{rasch\_mml} & 0.9547 & 0.0005 \\
\midrule
Combined & 1 & 0.9616 & \texttt{avg} (tie) & 0.9616 & 0.0000 \\
\bottomrule
\end{tabular}%
}
\end{table}

%% file: Tables/gold_standard_low_agreement.tex
\begin{table}[!t]
\centering
\scriptsize
\setlength{\tabcolsep}{2pt}
\caption{Low-agreement tail: methods whose full-trial rankings have Kendall's $\tau_b < 0.85$ relative to \bayesu{80} (computed at $N=80$).}
\label{tab:gold_standard_low_agreement}
\begin{tabular*}{\columnwidth}{l@{\extracolsep{\fill}}c}
\toprule
Method & $\tau_b$ \\
\midrule
\multicolumn{2}{l}{\textit{\aimefour}} \\
\texttt{minimax\_variant\_margin\_tie\_ignore} & 0.682 \\
\texttt{minimax\_variant\_margin\_tie\_half} & 0.682 \\
\texttt{minimax\_variant\_winning\_votes\_tie\_half} & 0.682 \\
\texttt{minimax\_variant\_winning\_votes\_tie\_ignore} & 0.693 \\
\texttt{nanson\_rank\_ties\_average} & 0.798 \\
\texttt{nanson\_rank\_ties\_max} & 0.802 \\
\texttt{dynamic\_irt\_growth} & 0.821 \\
\texttt{majority\_judgment} & 0.842 \\
\texttt{rasch\_3pl} & 0.842 \\
\texttt{rasch\_3pl\_map} & 0.842 \\
\midrule
\multicolumn{2}{l}{\textit{\aimefive}} \\
\texttt{minimax\_variant\_winning\_votes\_tie\_ignore} & 0.771 \\
\texttt{majority\_judgment} & 0.779 \\
\texttt{minimax\_variant\_margin\_tie\_ignore} & 0.819 \\
\texttt{minimax\_variant\_margin\_tie\_half} & 0.819 \\
\texttt{minimax\_variant\_winning\_votes\_tie\_half} & 0.819 \\
\texttt{nanson\_rank\_ties\_max} & 0.840 \\
\texttt{nanson\_rank\_ties\_average} & 0.849 \\
\midrule
\multicolumn{2}{l}{\textit{\hmmt}} \\
\texttt{nanson\_rank\_ties\_max} & 0.758 \\
\texttt{inverse\_difficulty} & 0.811 \\
\texttt{nanson\_rank\_ties\_average} & 0.818 \\
\texttt{minimax\_variant\_margin\_tie\_ignore} & 0.831 \\
\texttt{minimax\_variant\_margin\_tie\_half} & 0.831 \\
\texttt{minimax\_variant\_winning\_votes\_tie\_half} & 0.831 \\
\texttt{baldwin\_rank\_ties\_max} & 0.850 \\
\midrule
\multicolumn{2}{l}{\textit{\brumo}} \\
\texttt{rasch\_3pl} & 0.789 \\
\texttt{rasch\_3pl\_map} & 0.789 \\
\texttt{minimax\_variant\_margin\_tie\_ignore} & 0.814 \\
\texttt{minimax\_variant\_margin\_tie\_half} & 0.814 \\
\texttt{minimax\_variant\_winning\_votes\_tie\_half} & 0.814 \\
\texttt{inverse\_difficulty} & 0.821 \\
\midrule
\multicolumn{2}{l}{\textit{Combined}} \\
\texttt{minimax\_variant\_winning\_votes\_tie\_ignore} & 0.748 \\
\texttt{minimax\_variant\_margin\_tie\_ignore} & 0.825 \\
\texttt{minimax\_variant\_margin\_tie\_half} & 0.825 \\
\texttt{minimax\_variant\_winning\_votes\_tie\_half} & 0.825 \\
\texttt{nanson\_rank\_ties\_max} & 0.843 \\
\bottomrule
\end{tabular*}
\end{table}

%% file: Tables/aime24_kendalltau.tex
\begin{table*}[t!]
\centering
\scriptsize
\setlength{\tabcolsep}{4pt}
\caption{Consensus ranking on \aimefour{} by average Kendall's $\tau_b$ agreement with all other methods at $N=80$ (higher is better). Method variants with identical (Avg., Std.) are collapsed; we show the top 10 and bottom 5 groups.}
\label{tab:aime24_kendalltau}
\begin{tabular}{r p{0.76\textwidth} cc}
\toprule
Rank & Method(s) & Avg. & Std. \\
\midrule
1 & \texttt{\detokenize{rasch_mml}} & 0.9417 & 0.0799 \\
2 & \texttt{\detokenize{alpharank, bayes, bayes_ci, bradley_terry_davidson, bradley_terry_davidson_map, dynamic_irt_linear, glicko_tie_draw, hodge_rank_binary_decisive, hodge_rank_binary_total, hodge_rank_binary_uniform, avg, nash_advantage_vs_equilibrium, nash_vs_equilibrium, pagerank, rank_centrality_tie_half, rasch, rasch_map, serial_rank_prob_diff, serial_rank_sign, spectral, thompson}} & 0.9414 & 0.0815 \\
3 & \texttt{\detokenize{bayes_greedy}} & 0.9407 & 0.0817 \\
4 & \texttt{\detokenize{bayesian_mcmc, bradley_terry, bradley_terry_map, elo_tie_skip, glicko_tie_correct_draw_only, glicko_tie_skip, mg_pass_at_k_2, pass_hat_k_2, plackett_luce, plackett_luce_map, trueskill}} & 0.9403 & 0.0817 \\
5 & \texttt{\detokenize{hodge_rank_log_odds_decisive, hodge_rank_log_odds_total, hodge_rank_log_odds_uniform, rank_centrality_tie_ignore, rao_kupper, rao_kupper_map}} & 0.9349 & 0.0816 \\
6 & \texttt{\detokenize{borda}} & 0.9179 & 0.0729 \\
7 & \texttt{\detokenize{baldwin_rank_ties_max}} & 0.9163 & 0.0754 \\
8 & \texttt{\detokenize{elo_tie_correct_draw_only, elo_tie_draw}} & 0.9141 & 0.0689 \\
9 & \texttt{\detokenize{copeland}} & 0.9108 & 0.0728 \\
10 & \texttt{\detokenize{schulze_tie_half}} & 0.9045 & 0.0722 \\
\midrule
\multicolumn{4}{c}{\emph{Omitted ranks 11--22.}} \\
\midrule
23 & \texttt{\detokenize{nanson_rank_ties_max}} & 0.7943 & 0.0492 \\
24 & \texttt{\detokenize{nanson_rank_ties_average}} & 0.7904 & 0.0431 \\
25 & \texttt{\detokenize{dynamic_irt_growth}} & 0.7897 & 0.0763 \\
26 & \texttt{\detokenize{minimax_variant_winning_votes_tie_ignore}} & 0.6887 & 0.0691 \\
27 & \texttt{\detokenize{minimax_variant_margin_tie_half, minimax_variant_margin_tie_ignore,} \newline \detokenize{minimax_variant_winning_votes_tie_half}} & 0.6777 & 0.0763 \\
\bottomrule
\end{tabular}

\vspace{5mm}

\caption{Consensus ranking on \aimefive{} by average Kendall's $\tau_b$ agreement with all other methods at $N=80$ (higher is better). Method variants with identical (Avg., Std.) are collapsed; we show the top 10 and bottom 5 groups.}
\label{tab:aime25_kendalltau}
\begin{tabular}{r p{0.76\textwidth} cc}
\toprule
Rank & Method(s) & Avg. & Std. \\
\midrule
1 & \texttt{\detokenize{alpharank, bayes, bayes_ci, bradley_terry_davidson, bradley_terry_davidson_map, dynamic_irt_linear, hodge_rank_binary_decisive, hodge_rank_binary_total, hodge_rank_binary_uniform, avg, nash_advantage_vs_equilibrium, nash_vs_equilibrium, pagerank, rank_centrality_tie_half, rasch, rasch_map, serial_rank_prob_diff, serial_rank_sign, spectral, thompson}} & 0.9344 & 0.0595 \\
2 & \texttt{\detokenize{glicko_tie_draw}} & 0.9331 & 0.0486 \\
3 & \texttt{\detokenize{bayes_greedy}} & 0.9306 & 0.0577 \\
4 & \texttt{\detokenize{rasch_mml}} & 0.9293 & 0.0349 \\
5 & \texttt{\detokenize{hodge_rank_log_odds_decisive, hodge_rank_log_odds_total, hodge_rank_log_odds_uniform, rao_kupper, rao_kupper_map}} & 0.9285 & 0.0541 \\
6 & \texttt{\detokenize{glicko_tie_correct_draw_only}} & 0.9285 & 0.0522 \\
7 & \texttt{\detokenize{rasch_mml_credible}} & 0.9249 & 0.0216 \\
8 & \texttt{\detokenize{rasch_3pl, rasch_3pl_map}} & 0.9172 & 0.0495 \\
9 & \texttt{\detokenize{rasch_2pl, rasch_2pl_map}} & 0.9162 & 0.0507 \\
10 & \texttt{\detokenize{bradley_terry, bradley_terry_map, plackett_luce, plackett_luce_map}} & 0.9156 & 0.0486 \\
\midrule
\multicolumn{4}{c}{\emph{Omitted ranks 11--26.}} \\
\midrule
27 & \texttt{\detokenize{inverse_difficulty}} & 0.8447 & 0.0455 \\
28 & \texttt{\detokenize{nanson_rank_ties_max}} & 0.8353 & 0.0256 \\
29 & \texttt{\detokenize{minimax_variant_margin_tie_half, minimax_variant_margin_tie_ignore,} \newline \detokenize{minimax_variant_winning_votes_tie_half}} & 0.8280 & 0.0377 \\
30 & \texttt{\detokenize{majority_judgment}} & 0.8029 & 0.0322 \\
31 & \texttt{\detokenize{minimax_variant_winning_votes_tie_ignore}} & 0.7923 & 0.0386 \\
\bottomrule
\end{tabular}
\end{table*}

%% file: Tables/hmmt_kendalltau.tex
\begin{table*}[t]
\centering
\scriptsize
\setlength{\tabcolsep}{4pt}
\caption{Consensus ranking on \hmmt{} by average Kendall's $\tau_b$ agreement with all other methods at $N=80$ (higher is better). Method variants with identical (Avg., Std.) are collapsed; we show the top 10 and bottom 5 groups.}
\label{tab:hmmt_kendalltau}
\begin{tabular}{r p{0.76\textwidth} cc}
\toprule
Rank & Method(s) & Avg. & Std. \\
\midrule
1 & \texttt{\detokenize{alpharank, bayes, bayes_ci, bradley_terry, bradley_terry_davidson, bradley_terry_davidson_map, bradley_terry_map, dynamic_irt_linear, elo_tie_correct_draw_only, elo_tie_skip, glicko_tie_correct_draw_only, glicko_tie_draw, glicko_tie_skip, hodge_rank_binary_decisive, hodge_rank_binary_total, hodge_rank_binary_uniform, hodge_rank_log_odds_decisive, hodge_rank_log_odds_total, hodge_rank_log_odds_uniform, avg, nash_advantage_vs_equilibrium, nash_vs_equilibrium, pagerank, plackett_luce, plackett_luce_map, rank_centrality_tie_half, rao_kupper, rao_kupper_map, rasch, rasch_map, serial_rank_prob_diff, serial_rank_sign, spectral, thompson, trueskill}} & 0.9499 & 0.0631 \\
2 & \texttt{\detokenize{rasch_mml}} & 0.9494 & 0.0442 \\
3 & \texttt{\detokenize{bayes_greedy}} & 0.9468 & 0.0556 \\
4 & \texttt{\detokenize{rasch_3pl, rasch_3pl_map}} & 0.9420 & 0.0624 \\
5 & \texttt{\detokenize{rank_centrality_tie_ignore}} & 0.9415 & 0.0511 \\
6 & \texttt{\detokenize{elo_tie_draw}} & 0.9356 & 0.0561 \\
7 & \texttt{\detokenize{bayesian_mcmc}} & 0.9335 & 0.0495 \\
8 & \texttt{\detokenize{dynamic_irt_growth}} & 0.9287 & 0.0434 \\
9 & \texttt{\detokenize{pass_at_k_2}} & 0.9169 & 0.0325 \\
10 & \texttt{\detokenize{rasch_2pl, rasch_2pl_map}} & 0.9161 & 0.0629 \\
\midrule
\multicolumn{4}{c}{\emph{Omitted ranks 11--22.}} \\
\midrule
23 & \texttt{\detokenize{majority_judgment}} & 0.8636 & 0.0276 \\
24 & \texttt{\detokenize{minimax_variant_margin_tie_half, minimax_variant_margin_tie_ignore,} \newline \detokenize{minimax_variant_winning_votes_tie_half}} & 0.8391 & 0.0388 \\
25 & \texttt{\detokenize{inverse_difficulty}} & 0.8319 & 0.0458 \\
26 & \texttt{\detokenize{nanson_rank_ties_average}} & 0.8184 & 0.0156 \\
27 & \texttt{\detokenize{nanson_rank_ties_max}} & 0.7763 & 0.0346 \\
\bottomrule
\end{tabular}
\end{table*}

%% file: Tables/brumo_kendalltau.tex
\begin{table*}[t]
\centering
\scriptsize
\setlength{\tabcolsep}{4pt}
\caption{Consensus ranking on \brumo{} by average Kendall's $\tau_b$ agreement with all other methods at $N=80$ (higher is better). Method variants with identical (Avg., Std.) are collapsed; we show the top 10 and bottom 5 groups.}
\label{tab:brumo_kendalltau}
\begin{tabular}{r p{0.76\textwidth} cc}
\toprule
Rank & Method(s) & Avg. & Std. \\
\midrule
1 & \texttt{\detokenize{rasch_mml}} & 0.9547 & 0.0581 \\
2 & \texttt{\detokenize{alpharank, bayes, bayes_ci, bayes_greedy, bradley_terry_davidson, bradley_terry_davidson_map, dynamic_irt_linear, glicko_tie_draw, hodge_rank_binary_decisive, hodge_rank_binary_total, hodge_rank_binary_uniform, hodge_rank_log_odds_decisive, hodge_rank_log_odds_total, hodge_rank_log_odds_uniform, avg, nash_advantage_vs_equilibrium, nash_vs_equilibrium, pagerank, rank_centrality_tie_half, rao_kupper, rao_kupper_map, rasch, rasch_map, serial_rank_prob_diff, serial_rank_sign, spectral, thompson}} & 0.9542 & 0.0588 \\
3 & \texttt{\detokenize{glicko_tie_correct_draw_only}} & 0.9501 & 0.0575 \\
4 & \texttt{\detokenize{mg_pass_at_k_2, pass_hat_k_2}} & 0.9490 & 0.0582 \\
5 & \texttt{\detokenize{elo_tie_draw}} & 0.9423 & 0.0589 \\
6 & \texttt{\detokenize{borda, win_rate}} & 0.9399 & 0.0470 \\
7 & \texttt{\detokenize{bayesian_mcmc, bradley_terry, bradley_terry_map, elo_tie_correct_draw_only, elo_tie_skip, glicko_tie_skip, plackett_luce, plackett_luce_map, trueskill}} & 0.9382 & 0.0592 \\
8 & \texttt{\detokenize{copeland}} & 0.9376 & 0.0516 \\
9 & \texttt{\detokenize{bradley_terry_luce, bradley_terry_luce_map}} & 0.9350 & 0.0579 \\
10 & \texttt{\detokenize{dynamic_irt_growth}} & 0.9318 & 0.0505 \\
\midrule
\multicolumn{4}{c}{\emph{Omitted ranks 11--19.}} \\
\midrule
20 & \texttt{\detokenize{nanson_rank_ties_average, nanson_rank_ties_max}} & 0.8437 & 0.0294 \\
21 & \texttt{\detokenize{majority_judgment}} & 0.8267 & 0.0393 \\
22 & \texttt{\detokenize{inverse_difficulty}} & 0.8156 & 0.0273 \\
23 & \texttt{\detokenize{minimax_variant_margin_tie_half, minimax_variant_margin_tie_ignore,} \newline \detokenize{minimax_variant_winning_votes_tie_half}} & 0.8113 & 0.0446 \\
24 & \texttt{\detokenize{rasch_3pl, rasch_3pl_map}} & 0.7893 & 0.0396 \\
\bottomrule
\end{tabular}
\end{table*}

%% file: Tables/combined_kendalltau.tex
\begin{table*}[t]
\centering
\scriptsize
\setlength{\tabcolsep}{4pt}
\caption{Consensus ranking on the combined benchmark by average Kendall's $\tau_b$ agreement with all other methods at $N=80$ (higher is better). Method variants with identical (Avg., Std.) are collapsed; we show the top 10 and bottom 5 groups.}
\label{tab:combined_kendalltau}
\begin{tabular}{r p{0.76\textwidth} cc}
\toprule
Rank & Method(s) & Avg. & Std. \\
\midrule
1 & \texttt{\detokenize{alpharank, bayes, bayes_ci, bradley_terry_davidson, bradley_terry_davidson_map, dynamic_irt_linear, glicko_tie_draw, hodge_rank_binary_decisive, hodge_rank_binary_total, hodge_rank_binary_uniform, kemeny_young_tie_half, kemeny_young_tie_ignore, avg, nash_advantage_vs_equilibrium, nash_vs_equilibrium, pagerank, rank_centrality_tie_half, rasch, rasch_map, serial_rank_prob_diff, serial_rank_sign, spectral, thompson}} & 0.9616 & 0.0559 \\
2 & \texttt{\detokenize{copeland, ranked_pairs_strength_margin_tie_half, ranked_pairs_strength_margin_tie_ignore, ranked_pairs_strength_winning_votes_tie_half, ranked_pairs_strength_winning_votes_tie_ignore, schulze_tie_half, schulze_tie_ignore}} & 0.9602 & 0.0546 \\
3 & \texttt{\detokenize{hodge_rank_log_odds_decisive, hodge_rank_log_odds_total, hodge_rank_log_odds_uniform, rao_kupper, rao_kupper_map}} & 0.9567 & 0.0516 \\
4 & \texttt{\detokenize{glicko_tie_correct_draw_only}} & 0.9566 & 0.0522 \\
5 & \texttt{\detokenize{baldwin_rank_ties_average}} & 0.9558 & 0.0495 \\
6 & \texttt{\detokenize{borda}} & 0.9557 & 0.0516 \\
7 & \texttt{\detokenize{rank_centrality_tie_ignore}} & 0.9554 & 0.0505 \\
8 & \texttt{\detokenize{bayesian_mcmc}} & 0.9512 & 0.0493 \\
9 & \texttt{\detokenize{bayes_greedy}} & 0.9508 & 0.0473 \\
10 & \texttt{\detokenize{rasch_2pl}} & 0.9502 & 0.0486 \\
\midrule
\multicolumn{4}{c}{\emph{Omitted ranks 11--26.}} \\
\midrule
27 & \texttt{\detokenize{nanson_rank_ties_average}} & 0.8562 & 0.0263 \\
28 & \texttt{\detokenize{elo_tie_draw}} & 0.8552 & 0.0239 \\
29 & \texttt{\detokenize{minimax_variant_margin_tie_half, minimax_variant_margin_tie_ignore,} \newline \detokenize{minimax_variant_winning_votes_tie_half}} & 0.8339 & 0.0324 \\
30 & \texttt{\detokenize{nanson_rank_ties_max}} & 0.8333 & 0.0302 \\
31 & \texttt{\detokenize{minimax_variant_winning_votes_tie_ignore}} & 0.7665 & 0.0390 \\
\bottomrule
\end{tabular}
\end{table*}

%% file: Tables/gold_standard_exact_match.tex
\begin{table*}[t]
\centering
\scriptsize
\setlength{\tabcolsep}{4pt}
\caption{Methods that induce \emph{exactly} the same ranking as \bayesu{80} ($\tau_b=1$) when computed on the full $N=80$ trials (excluding avg itself).}
\label{tab:gold_standard_exact_match}
\begin{tabular}{lcp{0.76\textwidth}}
\toprule
Benchmark & Count & Methods \\
\midrule
\aimefour & 20 & {\ttfamily alpharank, bayes, bayes\_ci, bradley\_terry\_davidson, bradley\_terry\_davidson\_map, dynamic\_irt\_linear, glicko\_tie\_draw, hodge\_rank\_binary\_decisive, hodge\_rank\_binary\_total, hodge\_rank\_binary\_uniform, nash\_advantage\_vs\_equilibrium, nash\_vs\_equilibrium, pagerank, rank\_centrality\_tie\_half, rasch, rasch\_map, serial\_rank\_prob\_diff, serial\_rank\_sign, spectral, thompson} \\
\aimefive & 19 & {\ttfamily alpharank, bayes, bayes\_ci, bradley\_terry\_davidson, bradley\_terry\_davidson\_map, dynamic\_irt\_linear, hodge\_rank\_binary\_decisive, hodge\_rank\_binary\_total, hodge\_rank\_binary\_uniform, nash\_advantage\_vs\_equilibrium, nash\_vs\_equilibrium, pagerank, rank\_centrality\_tie\_half, rasch, rasch\_map, serial\_rank\_prob\_diff, serial\_rank\_sign, spectral, thompson} \\
\hmmt & 34 & {\ttfamily alpharank, bayes, bayes\_ci, bradley\_terry, bradley\_terry\_davidson, bradley\_terry\_davidson\_map, bradley\_terry\_map, dynamic\_irt\_linear, elo\_tie\_correct\_draw\_only, elo\_tie\_skip, glicko\_tie\_correct\_draw\_only, glicko\_tie\_draw, glicko\_tie\_skip, hodge\_rank\_binary\_decisive, hodge\_rank\_binary\_total, hodge\_rank\_binary\_uniform, hodge\_rank\_log\_odds\_decisive, hodge\_rank\_log\_odds\_total, hodge\_rank\_log\_odds\_uniform, nash\_advantage\_vs\_equilibrium, nash\_vs\_equilibrium, pagerank, plackett\_luce, plackett\_luce\_map, rank\_centrality\_tie\_half, rao\_kupper, rao\_kupper\_map, rasch, rasch\_map, serial\_rank\_prob\_diff, serial\_rank\_sign, spectral, thompson, trueskill} \\
\brumo & 26 & {\ttfamily alpharank, bayes, bayes\_ci, bayes\_greedy, bradley\_terry\_davidson, bradley\_terry\_davidson\_map, dynamic\_irt\_linear, glicko\_tie\_draw, hodge\_rank\_binary\_decisive, hodge\_rank\_binary\_total, hodge\_rank\_binary\_uniform, hodge\_rank\_log\_odds\_decisive, hodge\_rank\_log\_odds\_total, hodge\_rank\_log\_odds\_uniform, nash\_advantage\_vs\_equilibrium, nash\_vs\_equilibrium, pagerank, rank\_centrality\_tie\_half, rao\_kupper, rao\_kupper\_map, rasch, rasch\_map, serial\_rank\_prob\_diff, serial\_rank\_sign, spectral, thompson} \\
\midrule
Combined & 22 & {\ttfamily alpharank, bayes, bayes\_ci, bradley\_terry\_davidson, bradley\_terry\_davidson\_map, dynamic\_irt\_linear, glicko\_tie\_draw, hodge\_rank\_binary\_decisive, hodge\_rank\_binary\_total, hodge\_rank\_binary\_uniform, kemeny\_young\_tie\_half, kemeny\_young\_tie\_ignore, nash\_advantage\_vs\_equilibrium, nash\_vs\_equilibrium, pagerank, rank\_centrality\_tie\_half, rasch, rasch\_map, serial\_rank\_prob\_diff, serial\_rank\_sign, spectral, thompson} \\
\bottomrule
\end{tabular}
\end{table*}

%% file: appendix/stability_one.tex
\section{Ranking-Method Stability at \texorpdfstring{$N=1$}{N=1}}\label{sec:app:n1_stability}

This appendix provides additional details for the $N=1$ stability analyses in \cref{ssec:exp:n1_stability}. We report method rankings on the Combined benchmark for (i) gold-standard agreement (method@1 vs.\ \bayesu{80}) and (ii) self-consistency (method@1 vs.\ method@80), collapsing method variants with identical mean and standard deviation across the 80 single-trial draws.

\input{Tables/n1_gs_combined.tex}
\input{Tables/n1_self_combined.tex}

%% file: Tables/n1_gs_combined.tex
\begin{table*}[t]
\centering
\scriptsize
\setlength{\tabcolsep}{4pt}
\caption{Gold-standard agreement at $N=1$ on the combined benchmark, measured as Kendall's $\tau_b$ between each method's single-trial ranking and the gold standard (\bayesu{80}). Statistics are computed over $80$ single-trial draws. Methods with identical mean/std. values are collapsed; we show the top 10 and bottom 5 groups.}
\label{tab:n1_gs_combined}
\begin{tabular}{r p{0.76\textwidth} cc}
\toprule
Rank & Method(s) & Mean & Std. \\
\midrule
1 & \texttt{\detokenize{baldwin_rank_ties_average, bayes, bayes_ci, borda, copeland, majority_judgment, avg, minimax_variant_margin_tie_half, minimax_variant_margin_tie_ignore, minimax_variant_winning_votes_tie_half, nash_advantage_vs_equilibrium, nash_vs_equilibrium, pagerank, rank_centrality_tie_half, ranked_pairs_strength_margin_tie_half, ranked_pairs_strength_margin_tie_ignore, ranked_pairs_strength_winning_votes_tie_half, ranked_pairs_strength_winning_votes_tie_ignore, schulze_tie_half, schulze_tie_ignore, spectral}} & 0.8647 & 0.0486 \\
2 & \texttt{\detokenize{alpharank}} & 0.8646 & 0.0486 \\
3 & \texttt{\detokenize{rasch_mml_credible}} & 0.8642 & 0.0351 \\
4 & \texttt{\detokenize{hodge_rank_binary_uniform}} & 0.8623 & 0.0491 \\
5 & \texttt{\detokenize{hodge_rank_binary_decisive}} & 0.8623 & 0.0484 \\
6 & \texttt{\detokenize{hodge_rank_binary_total}} & 0.8616 & 0.0493 \\
7 & \texttt{\detokenize{serial_rank_sign}} & 0.8615 & 0.0503 \\
8 & \texttt{\detokenize{hodge_rank_log_odds_total, hodge_rank_log_odds_uniform}} & 0.8603 & 0.0482 \\
9 & \texttt{\detokenize{rao_kupper_map}} & 0.8603 & 0.0483 \\
10 & \texttt{\detokenize{rao_kupper}} & 0.8601 & 0.0484 \\
\midrule
\multicolumn{4}{c}{\emph{Omitted ranks 11--38.}} \\
\midrule
39 & \texttt{\detokenize{nanson_rank_ties_average}} & 0.8067 & 0.0363 \\
40 & \texttt{\detokenize{bradley_terry_luce_map}} & 0.8064 & 0.0556 \\
41 & \texttt{\detokenize{bradley_terry_luce}} & 0.8058 & 0.0554 \\
42 & \texttt{\detokenize{bayes_greedy}} & 0.7856 & 0.0309 \\
43 & \texttt{\detokenize{nanson_rank_ties_max}} & 0.7825 & 0.0394 \\
\bottomrule
\end{tabular}
\end{table*}

%% file: Tables/n1_self_combined.tex
\begin{table*}[t]
\centering
\scriptsize
\setlength{\tabcolsep}{4pt}
\caption{Self-consistency at $N=1$ on the combined benchmark, measured as Kendall's $\tau_b$ between each method's single-trial ranking and its own full-trial ranking (method@80). Statistics are computed over 80 single-trial draws. Methods with identical (Mean, Std.) are collapsed; we show the top 10 and bottom 5 groups.}
\label{tab:n1_self_combined}
\begin{tabular}{r p{0.76\textwidth} cc}
\toprule
Rank & Method(s) & Mean & Std. \\
\midrule
1 & \texttt{\detokenize{nanson_rank_ties_average}} & 0.8925 & 0.0497 \\
2 & \texttt{\detokenize{rasch_mml_credible}} & 0.8831 & 0.0370 \\
3 & \texttt{\detokenize{nanson_rank_ties_max}} & 0.8669 & 0.0589 \\
4 & \texttt{\detokenize{baldwin_rank_ties_average}} & 0.8664 & 0.0492 \\
5 & \texttt{\detokenize{copeland, ranked_pairs_strength_margin_tie_half, ranked_pairs_strength_margin_tie_ignore, ranked_pairs_strength_winning_votes_tie_half, ranked_pairs_strength_winning_votes_tie_ignore, schulze_tie_half, schulze_tie_ignore}} & 0.8654 & 0.0489 \\
6 & \texttt{\detokenize{rasch_mml}} & 0.8648 & 0.0417 \\
7 & \texttt{\detokenize{bayes, bayes_ci, avg, nash_advantage_vs_equilibrium, nash_vs_equilibrium, pagerank, rank_centrality_tie_half, spectral}} & 0.8647 & 0.0486 \\
8 & \texttt{\detokenize{alpharank}} & 0.8646 & 0.0486 \\
9 & \texttt{\detokenize{borda}} & 0.8646 & 0.0499 \\
10 & \texttt{\detokenize{hodge_rank_binary_uniform}} & 0.8623 & 0.0491 \\
\midrule
\multicolumn{4}{c}{\emph{Omitted ranks 11--44.}} \\
\midrule
45 & \texttt{\detokenize{elo_tie_correct_draw_only}} & 0.8074 & 0.0507 \\
46 & \texttt{\detokenize{bayes_greedy}} & 0.8064 & 0.0309 \\
47 & \texttt{\detokenize{elo_tie_draw}} & 0.8063 & 0.0507 \\
48 & \texttt{\detokenize{minimax_variant_margin_tie_half, minimax_variant_margin_tie_ignore,} \newline \detokenize{minimax_variant_winning_votes_tie_half}} & 0.7963 & 0.0454 \\
49 & \texttt{\detokenize{minimax_variant_winning_votes_tie_ignore}} & 0.7655 & 0.0455 \\
\bottomrule
\end{tabular}
\end{table*}

%% file: appendix/priors.tex
\section{Additional Prior Diagnostics}\label{sec:app:prior_diagnostics}

This appendix collects supplementary diagnostics for the empirical-prior analysis in \cref{ssec:priors}.

\begin{figure}[t]
    \centering
    \includegraphics[width=\columnwidth]{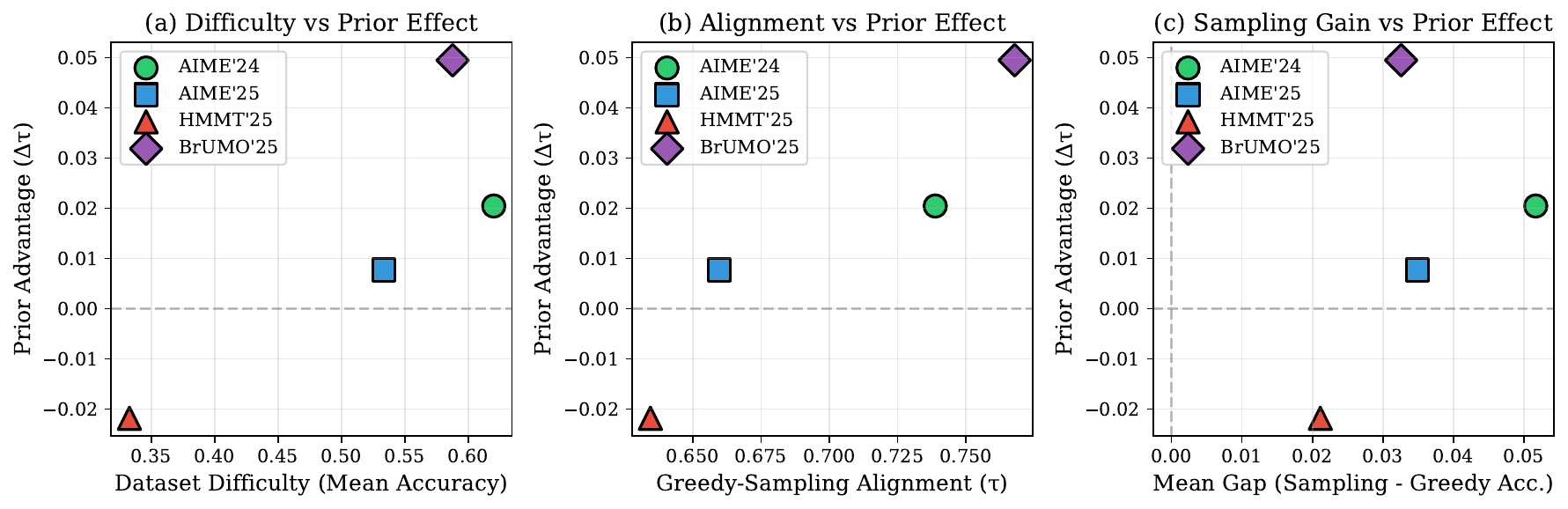}
    \caption{Across our four benchmarks, the prior advantage is not monotonically related to difficulty (a), but it is associated with greedy--sampling alignment (b). The sampling--greedy accuracy gap (c) shows no clear relationship.}
    \label{fig:difficulty_correlation}
\end{figure}

\begin{figure}[t]
    \centering
    \includegraphics[width=\columnwidth]{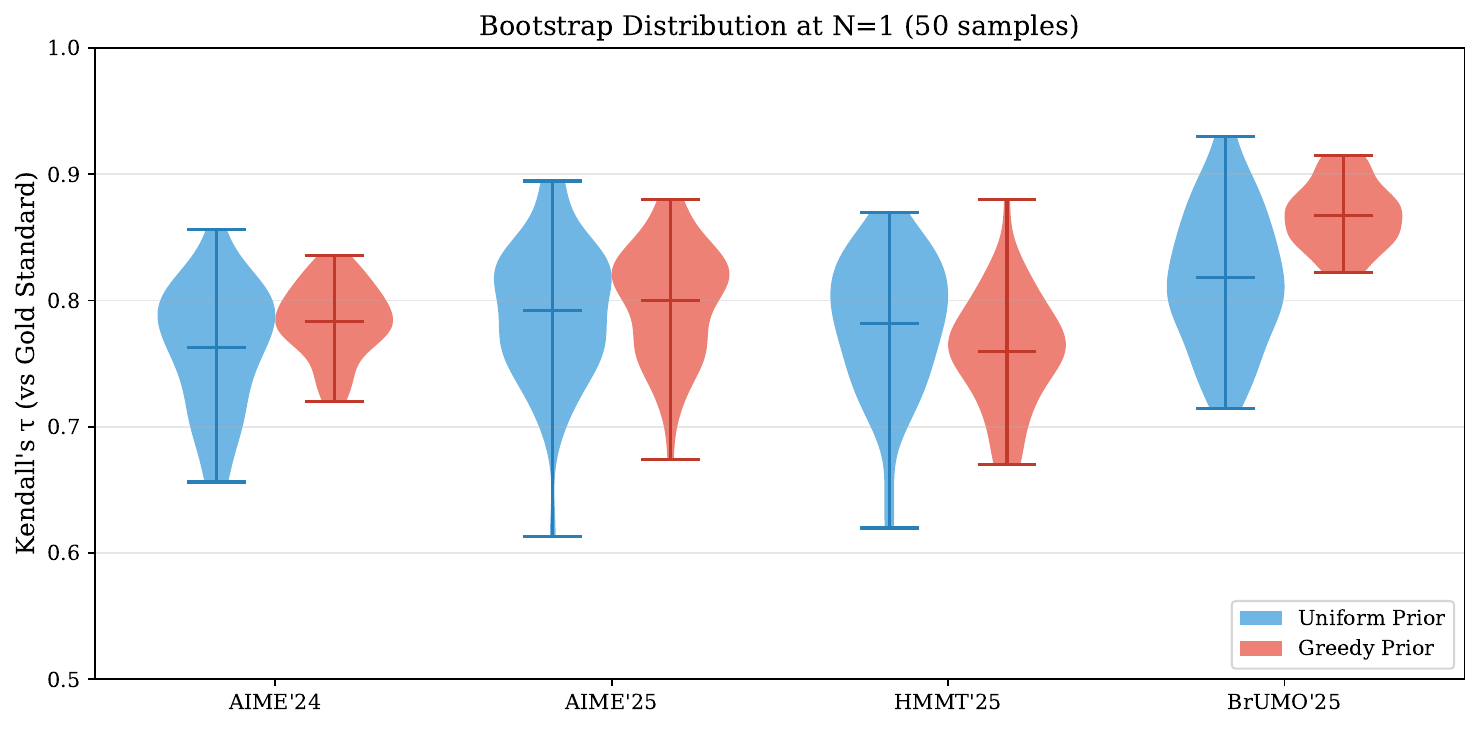}
    \caption{Bootstrap distributions of Kendall's $\tau_b$ at $N=1$ ($50$ samples). Violin plots show the full distribution; the greedy prior (red) yields narrower distributions but can shift the mean negatively (\hmmt{}) or positively (\brumo{}).}
    \label{fig:bootstrap_violin}
\end{figure}

%% file: appendix/cat.tex
\section{Categorical Ranking}\label{sec:app:cat}

This appendix gives the experimental setup and per-dataset results for the categorical-ranking experiments summarized in \cref{sec:exp:cat}.

\subsection{Setup}\label{app:ssub:cat:framework}

The binary Bayesian estimator (\cref{ssec:bayesian_ranking}) models each trial outcome as $R_{lmn}\in\{0,1\}$ and places a Beta prior on the per-question solve rate. We generalize this to categorical outcomes $R_{lmn}\in\{0,\dots,C\}$, where each completion is mapped to one of $C+1$ categories based on auxiliary signals extracted during generation. A categorical \emph{scheme} $s$ specifies:
\begin{enumerate}[nosep]
\item a categorical mapping $\phi_s\colon \text{completion features}\to\{0,\dots,C_s\}$, which assigns each completion to a category based on predicates over the base signals (\cref{tab:cat_signals}), and
\item a utility weight vector $\mathbf{w}_s=(w_0,\dots,w_{C_s})\in\mathbb{R}^{C_s+1}$, encoding the relative value of each category.
\end{enumerate}
Bayesian estimation replaces the Beta--binomial model with a Dirichlet--multinomial model: for each model--question pair, we place a symmetric Dirichlet prior on the $C+1$ category probabilities $\boldsymbol{\theta}=(\theta_0,\dots,\theta_C)$ and compute the posterior mean of the weighted utility $\sum_{k=0}^{C}w_k\hat{\theta}_k$. Model-level scores are then aggregated across questions, as in the binary case.

\subsection{Base Signals}\label{app:ssub:cat:signals}

For each of the $L=11$ models in the categorical cohort, we extract $9$ features per completion (\cref{tab:cat_signals}). These features span five domains: answer format (\texttt{has\_box}), correctness (\texttt{is\_correct}), generation cost (\texttt{token\_ratio}, \texttt{repeated\_pattern}), decoding confidence (\texttt{prompt\_bpt}, \texttt{completion\_bpt}), and external verification via CompassVerifier (\texttt{compass\_A/B/C}). The feature tensors have shape $(N, M, 9)$ per model, with $N=80$ trials and $M=30$ questions per benchmark.

We use \compassicon~\texttt{CompassVerifier-3B} as a reward model for the external verification signals. During evaluation, we use its scores on completions generated by the other models to define the categorical schemes. The implementation uses \texttt{Transformers}~\cite{wolf2019huggingface} and \texttt{Accelerate}~\cite{accelerate}, with FlashAttention kernels~\cite{Dao2022FlashAttention} and the \texttt{DFloat11} format~\cite{Zhang2025DFloat11} for throughput.

\input{Tables/categorical_signals.tex}

\subsection{Derived Predicates and Thresholds}\label{app:ssub:cat:thresholds}

Several predicates are shared across schemes. All thresholds are computed per-model from the available samples:
\begin{itemize}[nosep]
\item \textbf{Invalid}: $\texttt{repeated\_pattern}=1$ or $\texttt{compass\_C}\ge 0.5$.
\item \textbf{Confidence}: High confidence $:=$ $\texttt{completion\_bpt}\le P_{40}(\texttt{completion\_bpt})$; wrong--high-confidence $:=$ wrong and $\texttt{completion\_bpt}\le P_{60}(\texttt{completion\_bpt}\mid\text{wrong})$.
\item \textbf{Prompt OOD}: $\texttt{prompt\_bpt}\ge P_{90}(\texttt{prompt\_bpt})$.
\item \textbf{Efficiency bands}: Economical/moderate/verbose based on $P_{33}$ and $P_{66}$ of $\texttt{token\_ratio}$.
\item \textbf{Verifier}: CompassVerifier dominant label is $\arg\max(A,B,C)$; verifier-high $:=$ $A\ge 0.6$.
\end{itemize}

\subsection{Scheme Definitions}\label{app:ssub:cat:schemes}

We design $25$ categorical schemes spanning diverse design axes: correctness-only baselines (A, H, S, Y), confidence-aware (C, I, J, V), format-aware (B, P, T), efficiency-aware (F, G, M), verifier-based (D, K, O, U, Z), OOD-aware (E, N, W), abstention-aware (L, Q), and composite (R). Many schemes are metric-level near-duplicates (e.g., A $\equiv$ S, H $\equiv$ Y, L $\equiv$ Q); we therefore select $8$ non-redundant representative schemes covering distinct design axes (\cref{tab:cat_schemes}).

\input{Tables/categorical_schemes.tex}

\subsection{Evaluation Protocol}\label{app:ssub:cat:protocol}

For each scheme, we apply the same $N=1$ subsampling protocol as in \cref{ssec:exp:n1_stability}: we subsample one of the $N=80$ trials per question, compute the scheme's categorical ranking, and measure Kendall's $\tau_b$ against three references:
\begin{enumerate}[nosep]
\item \textbf{Gold-standard} ($\tau_{\text{GS}}$): agreement with the binary \bayesu{80} ranking, which treats outcomes as correct/wrong with a uniform Dirichlet prior.
\item \textbf{Self-consistency} ($\tau_{\text{Self}}$): agreement with the scheme's own all-$80$-trial ranking (Scheme@$80$).
\item \textbf{Greedy-prior} ($\tau_{\text{Greedy}}$): agreement with \bayesg{80}, the binary Bayes ranking incorporating a greedy-decoding empirical prior.
\end{enumerate}
Statistics (mean and standard deviation) are computed over the $80$ single-trial draws. Combined results aggregate the four benchmarks ($M=120$ questions) and are reported in \cref{tab:cat_combined}; per-dataset results are reported below.

\subsection{Per-Dataset Results}\label{app:ssub:cat:per_dataset}

\cref{tab:cat_per_dataset} reports gold-standard agreement and self-consistency for each benchmark separately. Three patterns emerge.

\input{Tables/categorical_per_dataset.tex}

\paragraph{Narrow spread on individual benchmarks.}
On each benchmark individually, all eight schemes achieve $\tau_{\text{GS}}$ between $0.73$ and $0.83$, with inter-scheme variation much smaller than on the combined benchmark. On \aimefour{}, the range across the $8$ schemes is only $0.007$ ($0.813$--$0.820$). This is because, with $M=30$ questions and $L=11$ models, a single trial provides limited information for distinguishing among category structures; the combined benchmark ($M=120$) offers finer discrimination.

\paragraph{Verifier-only degrades on hard benchmarks.}
The Verifier-only scheme exhibits the largest performance drop on the harder benchmarks: $\tau_{\text{GS}}$ falls from $0.813$ (\aimefour{}) to $0.753$ (\hmmt{}) and $0.734$ (\brumo{}), a decline of $0.06$--$0.08$. In contrast, correctness-driven schemes (Conservative, Efficiency-adjusted, Format-aware) remain above $0.80$ on all benchmarks. This suggests that CompassVerifier judgments are less reliable proxies for correctness on more challenging problems.

\paragraph{Self-consistency converges to gold-standard on individual benchmarks.}
On \aimefour{}, the self-consistency column is nearly identical to the gold-standard column for most schemes, indicating that the all-$80$ scheme ranking coincides with the binary \bayesu{80} ranking when the number of questions is small. On the combined benchmark (\cref{tab:cat_combined}), self-consistency consistently exceeds gold-standard agreement, reflecting the fact that each scheme converges to its own distinct ordering when given enough questions.

%% file: Tables/categorical_signals.tex
\begin{table}[t]
\centering
\small
\setlength{\tabcolsep}{4pt}
\caption{Nine base signals extracted per completion for the categorical ranking experiments. Each model--question--trial entry produces a vector in $\mathbb{R}^9$.}
\label{tab:cat_signals}
\begin{tabular}{@{}r l l@{}}
\toprule
\# & Signal & Description \\
\midrule
1 & \texttt{has\_box} & Boxed final answer present (0/1) \\
2 & \texttt{is\_correct} & Exact-match correctness (0/1) \\
3 & \texttt{token\_ratio} & Completion tokens / 32768 \\
4 & \texttt{repeated\_pattern} & Non-\texttt{stop} finish reason (0/1) \\
5 & \texttt{prompt\_bpt} & Prompt bits-per-token \\
6 & \texttt{completion\_bpt} & Completion bits-per-token \\
7 & \texttt{compass\_A} & Verifier $P(\text{correct})$ \\
8 & \texttt{compass\_B} & Verifier $P(\text{wrong})$ \\
9 & \texttt{compass\_C} & Verifier $P(\text{irrelevant})$ \\
\bottomrule
\end{tabular}
\end{table}

%% file: Tables/categorical_schemes.tex
\begin{table*}[t]
\centering
\small
\setlength{\tabcolsep}{4pt}
\caption{Eight representative categorical schemes used in \cref{sec:exp:cat}. Each scheme maps a completion to one of $C+1$ categories using the base signals in \cref{tab:cat_signals} and scores the result with a utility weight vector $\mathbf{w}$. Category $0$ is always \emph{Invalid} ($w_0=0$) unless otherwise noted.}
\label{tab:cat_schemes}
\resizebox{\textwidth}{!}{
\begin{tabular}{@{}l l l l@{}}
\toprule
Scheme & Intent & Categories ($k$) & Weights $\mathbf{w}$ \\
\midrule
Conservative &
Penalize confidently-wrong &
\begin{tabular}[t]{@{}l@{}}1: Wrong $\wedge$ HighConf \\ 2: Wrong $\wedge$ LowConf \\ 3: Correct\end{tabular} &
$(0,\, {-}0.10,\, 0.05,\, 1.00)$ \\[6pt]
Efficiency-adj. &
Discount verbose correct &
\begin{tabular}[t]{@{}l@{}}1--3: Wrong $\times$ \{Econ., Mod., Verb.\} \\ 4--6: Correct $\times$ \{Econ., Mod., Verb.\}\end{tabular} &
$(0,\, 0.10,\, 0.07,\, 0.03,\, 1,\, 0.92,\, 0.85)$ \\[6pt]
Format-aware &
Reward boxed correct &
\begin{tabular}[t]{@{}l@{}}1: Wrong $\wedge$ Unboxed; 2: Wrong $\wedge$ Boxed \\ 3: Correct; 4: Correct $\wedge$ Boxed\end{tabular} &
$(0,\, 0.10,\, 0.05,\, 0.90,\, 1)$ \\[6pt]
Balanced comp. &
Format $\times$ confidence &
\begin{tabular}[t]{@{}l@{}}1: Wrong $\wedge$ Unboxed; 2--3: Wrong $\wedge$ Boxed $\times$ Conf \\ 4--7: Correct $\times$ \{Un/Boxed\} $\times$ Conf\end{tabular} &
$(0,\, 0.10,\, 0.06,\, {-}0.02,\, 0.90,\, 0.95,\, 0.97,\, 1)$ \\[6pt]
OOD-robust &
Reward in-distribution &
\begin{tabular}[t]{@{}l@{}}1: OOD $\wedge$ Wrong; 2: InDist $\wedge$ Wrong \\ 3: OOD $\wedge$ Correct; 4: InDist $\wedge$ Correct\end{tabular} &
$(0,\, 0.05,\, 0.10,\, 0.95,\, 1)$ \\[6pt]
Rare-event &
OOD + abstention &
\begin{tabular}[t]{@{}l@{}}1: OOD $\wedge$ Wrong; 2: OOD $\wedge$ Correct \\ 3: InDist $\wedge$ Wrong; 4: InDist $\wedge$ Correct; 5: Abstain\end{tabular} &
$(0,\, 0.05,\, 1,\, 0.08,\, 0.95,\, 0.20)$ \\[6pt]
Verifier-calib. &
Penalize false-positive &
\begin{tabular}[t]{@{}l@{}}1: Wrong $\wedge$ $A{\ge}0.6$; 2: Wrong $\wedge$ $A{<}0.6$ \\ 3--5: Correct $\times$ \{$A_\text{low}$, $A_\text{mid}$, $A_\text{high}$\}\end{tabular} &
$(0,\, {-}0.05,\, 0.05,\, 0.88,\, 0.94,\, 1)$ \\[6pt]
Verifier-only &
No ground truth &
\begin{tabular}[t]{@{}l@{}}0: Repeated; 1: Dominant $=C$ \\ 2: Dominant $=B$; 3: Dominant $=A$\end{tabular} &
$(0,\, 0,\, 0.1,\, 1)$ \\
\bottomrule
\end{tabular}
}
\end{table*}

%% file: Tables/categorical_per_dataset.tex
\begin{table*}[t]
\centering
\small
\setlength{\tabcolsep}{4pt}
\caption{Per-dataset categorical ranking at $N\!=\!1$. Gold-standard agreement ($\tau_{\text{GS}}$: vs.\ \bayesu{80}) and self-consistency ($\tau_{\text{Self}}$: vs.\ Scheme@$80$) for the $8$ representative schemes. Values are mean Kendall's $\tau_b$ over $80$ single-trial draws.}
\label{tab:cat_per_dataset}
\begin{tabular}{@{}l cccc cccc@{}}
\toprule
& \multicolumn{4}{c}{Gold-standard agreement ($\tau_{\text{GS}}$)} & \multicolumn{4}{c}{Self-consistency ($\tau_{\text{Self}}$)} \\
\cmidrule(lr){2-5} \cmidrule(lr){6-9}
Scheme & \aimefour & \aimefive & \hmmt & \brumo & \aimefour & \aimefive & \hmmt & \brumo \\
\midrule
Conservative     & 0.814 & 0.813 & 0.801 & 0.815 & 0.814 & 0.813 & 0.801 & 0.820 \\
Efficiency-adj.  & 0.814 & 0.821 & 0.812 & 0.817 & 0.814 & 0.814 & 0.814 & 0.828 \\
Format-aware     & 0.820 & 0.808 & 0.812 & 0.819 & 0.820 & 0.811 & 0.813 & 0.830 \\
Balanced comp.   & 0.816 & 0.810 & 0.804 & 0.806 & 0.816 & 0.813 & 0.805 & 0.816 \\
OOD-robust       & 0.819 & 0.806 & 0.788 & 0.810 & 0.819 & 0.803 & 0.802 & 0.824 \\
Rare-event       & 0.816 & 0.804 & 0.793 & 0.816 & 0.816 & 0.801 & 0.807 & 0.828 \\
Verifier-calib.  & 0.817 & 0.802 & 0.786 & 0.796 & 0.810 & 0.801 & 0.800 & 0.807 \\
Verifier-only    & 0.813 & 0.805 & 0.753 & 0.734 & 0.806 & 0.809 & 0.795 & 0.810 \\
\bottomrule
\end{tabular}
\end{table*}

%% file: appendix/related_work.tex
\section{Extended Related Work}\label{sec:app:extended_related_work}

Test-time scaling produces repeated stochastic outcomes per item, making LLM benchmarking closer to classical repeated-measurement settings than to single-run leaderboards. This appendix summarizes the main ranking families we use and their typical applications.

\paragraph{Paired-comparison and rating models.}
Paired-comparison models represent comparisons through win/tie counts and infer latent strengths, with Bradley--Terry as a canonical likelihood-based model \cite{BradleyTerry1952}. Practical systems often use online rating updates such as Elo and its extensions (e.g., Glicko) or fully Bayesian skill ratings such as TrueSkill \cite{Elo1978,Glickman1999,herbrich2006trueskill}. For data with ties, common generalizations include Rao--Kupper and Davidson models \cite{rao1967ties,davidson1970bties}. These models are widely used for preference aggregation in LLM leaderboards \cite{chiang2024chatbotarena,ameli2024statisticalframework}, but are also natural in dense benchmarks once per-item outcomes are reduced to pairwise wins.

\paragraph{Listwise, setwise choice models.}
When each trial yields an ordering over many items, listwise choice models such as Plackett--Luce provide a likelihood over permutations \cite{plackett1975permutations,luce1959choice}. Davidson--Luce extends setwise choice to allow ties within selected sets \cite{firth2019davidsonluce}. In our binary benchmark setting, each trial induces a two-level partition (solved vs.\ unsolved), so these models reduce to structured forms of pairwise likelihoods while still providing a principled view of aggregation.

\paragraph{IRT and difficulty-aware benchmarking.}
Item response theory models couple model ``ability'' with item difficulty (and sometimes discrimination), with the Rasch and Birnbaum formulations as classic examples \cite{rasch1960rasch,birnbaum1968latent}. IRT has recently been proposed as a way to disentangle model skill from benchmark composition in LLM evaluation \cite{zhou2025lost}. When multiple trials per item are available, repeated-measures extensions and binomial-response formulations are natural \cite{deboeckwilson2004explanatory,verhelst1993dynamicrasch,wang2019longitudinalirt}, and difficulty reweighting has also been explored in NLP evaluation contexts \cite{gotou2020correctiondifficulty}.

\paragraph{Graph, spectral, and social-choice methods.}
Beyond likelihood-based models, ranking from comparisons has a long tradition in social choice and graph-based aggregation. Voting rules such as Borda and Condorcet-style methods satisfy different axioms and can behave differently under noise and ties \cite{Borda1781,Condorcet1785,Arrow1951,brandt2016compsocchoice}. Spectral and Markov-chain approaches derive scores from transition graphs, including PageRank and Rank Centrality \cite{page1999pagerank,negahban2017rankcentrality}; HodgeRank and related spectral methods interpret comparisons as edge flows and decompose them into global and cyclic components \cite{jiang2009hodgerank,fogel2016serialrank}. AlphaRank was introduced for multi-agent evaluation with potentially non-transitive interactions \cite{omidshafiei2019alpharank}, and related work studies open-ended evaluation dynamics \cite{balduzzi2019openended}. Our study brings these families into a common test-time-scaling benchmark setting and compares them under controlled increases in the number of repeated trials.

%% file: appendix/experiment_setup.tex
\section{Experiment Setup and Reproducibility}\label{sec:app:setup}

\subsection{Models and Datasets}\label{app:exp}

\paragraph{Datasets.}
We evaluate on four Olympiad-style math benchmarks: \aimefour~\cite{MAA_AIME2024}, \aimefive~\cite{MAA_AIME2025}, \brumo~\cite{BrUMO_2025}, and \hmmt~\cite{HMMT_Feb2025}. For \aimefour{} and \aimefive{}, we combine AIME~I and AIME~II from the corresponding year, yielding $30$ integer-answer problems per benchmark. For \hmmt{}, we use the official February 2025 contest set, which spans algebra, geometry, number theory, and combinatorics. For \brumo{}, we use the published 2025 problem sets from the tournament archive.

\paragraph{Models.}
To reduce prompt-format confounds, we use provider-recommended chat templates (defaulting to DeepSeek/Qwen-style templates when no model-specific template is given) and shared decoding settings across models unless noted otherwise. Our base cohort comprises $11$ configurations: eight distinct models plus three reasoning-effort modes (low, medium, and high) of gpt-oss. These are:
\skyicon~Sky-T1-32B-Flash~\cite{reduce_overthinking_2025} (Sky-T1 Flash release),
\qwenicon~Qwen3-30B-A3B-Thinking-2507~\cite{qwen3technicalreport} (Qwen3 thinking model),
\dsicon~DeepSeek-R1-Distill-Qwen-1.5B~\cite{guo2025deepseek} (1.5B distilled reasoning model),
\gpticon~\mbox{gpt-oss-20b}~\cite{openai2025gptoss120bgptoss20bmodel} (OpenAI open-weight model; we use the default MXFP4 quantization and the Harmony reasoning-effort controls),
\gairicon~LIMO-v2~\cite{ye2025limoreasoning} (reasoning model),
\lgicon~EXAONE-4.0-1.2B~\cite{exaone-4.0} (hybrid reasoning/non-reasoning model),
\nvidiaicon~OpenReasoning-Nemotron-1.5B~\cite{openreasoning_nemotron_15b} (NVIDIA reasoning model), and
\openicon~OpenThinker2-32B~\cite{guha2025openthoughtsdatarecipesreasoning} and \openicon~OpenThinker3-1.5B~\cite{guha2025openthoughtsdatarecipesreasoning} (models trained from the OpenThoughts data recipes).

In addition to this base cohort, we evaluate nine more reasoning-capable models to study how the conclusions change with cohort size:
\microsoft~Phi-4-reasoning and \microsoft~Phi-4-reasoning-plus~\cite{abdin2025phi4reasoning},
\openr~OpenR1-Distill-7B~\cite{openr1},
\fuse~FuseO1-DeepSeekR1-QwQ-SkyT1-Flash-32B-Preview~\cite{fuseo1_flash_preview},
\qihoo~Light-R1-14B-DS~\cite{wen2025lightr1curriculumsftdpo},
\nvidiaicon~AceReason-Nemotron-1.1-7B~\cite{liu2025acereason},
\nvidiaicon~NVIDIA-Nemotron-Nano-9B-v2~\cite{nvidia2025nemotronnano2},
\qwenicon~Qwen3-4B-Thinking-2507~\cite{qwen3technicalreport}, and
\bespoke~Bespoke-Stratos-7B~\cite{bespoke_stratos}.

\begin{table}[t]
\centering
\resizebox{\columnwidth}{!}{
\begin{tabular}{c p{6.5cm} l}
\toprule
ID & Model & Short name \\
\midrule
1  & \dsicon~DeepSeek-R1-Distill-Qwen-1.5B                         & DS-R1-Qwen              \\
2  & \gairicon~LIMO-v2                                             & LIMO-v2                 \\
3  & \openicon~OpenThinker2-32B                                    & OpenThinker2            \\
4  & \openicon~OpenThinker3-1.5B                                   & OpenThinker3            \\
5  & \qwenicon~Qwen3-30B-A3B-Thinking-2507                         & Qwen3-Thinking          \\
6  & \skyicon~Sky-T1-32B-Flash                                     & Sky-T1-Flash            \\
7  & \gpticon~gpt-oss-20b\_high                                    & gpt-oss-high            \\
8  & \gpticon~gpt-oss-20b\_low                                     & gpt-oss-low             \\
9  & \gpticon~gpt-oss-20b\_medium                                  & gpt-oss-medium          \\
10 & \lgicon~EXAONE-4.0-1.2B                                       & EXAONE-4.0              \\
11 & \nvidiaicon~OpenReasoning-Nemotron-1.5B                       & OR-Nemotron             \\
12 & \microsoft~Phi-4-reasoning                                    & Phi-4                   \\
13 & \microsoft~Phi-4-reasoning-plus                               & Phi-4-plus              \\
14 & \openr~OpenR1-Distill-7B                                      & OR1-Distill             \\
15 & \fuse~FuseO1-DeepSeekR1-QwQ-\newline SkyT1-Flash-32B-Preview & FuseO1-DS-QwQ-SkyT1     \\
16 & \qihoo~Light-R1-14B-DS                                        & Light-R1-DS             \\
17 & \nvidiaicon~AceReason-Nemotron-1.1-7B                         & AR-Nemotron             \\
18 & \nvidiaicon~NVIDIA-Nemotron-Nano-9B-v2                        & NVIDIA-Nemotron         \\
19 & \qwenicon~Qwen3-4B-Thinking-2507                              & Qwen3-4B                \\
20 & \bespoke~Bespoke-Stratos-7B                                   & Bespoke                 \\
\bottomrule
\end{tabular}
}
\caption{Mapping between model IDs, full model names, and the shortened names used in figures and legends.}
\label{app:tab:modelid}
\end{table}

\paragraph{Prompting.}
We use provider-recommended prompt templates for each model. For most models, we adopt the standard DeepSeek/Qwen-style prompt, ``Please reason step by step, and put your final answer within \texttt{\textbackslash boxed\{\}}.'' For \gpticon~\texttt{gpt-oss-20b}, we use the OpenAI Harmony prompt template, which specifies three discrete levels of reasoning effort. For \nvidiaicon~\texttt{OpenReasoning-Nemotron-1.5B}, we use the task-specific prompt, ``Solve the following math problem. Make sure to put the answer (and only the answer) inside \texttt{\textbackslash boxed\{\}}.''

\subsection{Reproducibility}

For stochastic runs, we use top-$p$ sampling with temperature $0.6$, $p=0.95$, batch size $1$, and random seeds $1234$ through $1313$, yielding $N=80$ trials per dataset--model pair. All models are served with \texttt{vLLM} (PagedAttention)~\cite{kwon2023efficient} in \texttt{bf16} precision, except releases that require MXFP4 quantization (e.g., \texttt{gpt-oss}). We record log-probabilities for both input prompts and generated tokens, with \texttt{max\_tokens} set to $32{,}768$. All experiments run on clusters equipped with $8\times$ NVIDIA H200 GPUs (141\,GB per GPU).

\subsection{Computational Cost and Token Statistics}\label{sec:compute_stats}

We evaluate 20 models across four benchmarks, with 80 trials per model and 30 questions per benchmark, for a total of 192{,}000 independent inference runs. The full evaluation requires 7{,}445 GPU-hours (approximately 310 GPU-days) and generates 2.96B tokens (2{,}963{,}318{,}176 total); \cref{tab:computational_cost} reports the task-level totals. Of these tokens, 37M (1.2\%) are prompt tokens and 2.93B (98.8\%) are completion tokens, for an average of 15{,}434 tokens per query. Among the four benchmarks, \hmmt{} is the most computationally expensive at 2{,}217 GPU-hours, whereas \brumo{} is the least expensive at 1{,}651 GPU-hours. Across model configurations, \texttt{gpt-oss-20b-low} is the most efficient (48.4 GPU-hours for 9{,}600 queries) and \texttt{LIMO-v2} the least efficient (894.3 GPU-hours for the same workload), with a corpus-wide average of 139.6 seconds per query.

\begin{table}[t]
\centering
\resizebox{\columnwidth}{!}{
\begin{tabular}{lrr}
\toprule
\textbf{Task} & \textbf{Inference Time (hours)} & \textbf{Completion Tokens (M)} \\
\midrule
AIME'24  & 1{,}699.4 & 680.0 \\
AIME'25  & 1{,}878.4 & 728.3 \\
HMMT'25  & 2{,}216.5 & 851.2 \\
BrUMO'25 & 1{,}650.9 & 666.9 \\
\midrule
\textbf{TOTAL} & \textbf{7{,}445.2} & \textbf{2{,}926.4} \\
\bottomrule
\end{tabular}
}
\caption{Task-level computational cost aggregated over 20 models, 80 trials, four tasks, and 30 questions per task. Token counts correspond to completion tokens only.}
\label{tab:computational_cost}
\end{table}

\subsection{Ranking-Method Identifiers}\label{ssec:app:setup:ranking_apis}
Tables that report individual ranking methods use the corresponding \scorio{} identifiers (e.g., \texttt{avg}, \texttt{bayes}, \texttt{rasch\_mml}) for compactness. We print these identifiers verbatim to match the implementation used in the experiments and to make the reported rankings directly reproducible.

\subsection{Rank Correlation Metrics}\label{sec:app:metrics}

\paragraph{Kendall's tau}\label{app:kendall_tau}
Kendall's tau ($\tau$) \cite{kendall} measures ordinal agreement between two rankings through pairwise concordance and discordance. For rankings of $n$ items, let $n_c$ and $n_d$ denote the numbers of concordant and discordant pairs, let $n_0 = n(n-1)/2$ be the total number of pairs, and let $n_1$ and $n_2$ be the numbers of tied pairs in the two rankings. The two common variants are
\begin{align}
    \text{Tau-a:}\quad \tau_a &= \frac{n_c - n_d}{n_0}, \\
    \text{Tau-b:}\quad \tau_b &= \frac{n_c - n_d}{\sqrt{(n_0 - n_1)(n_0 - n_2)}}.
\end{align}
Tau-a ignores ties, whereas Tau-b corrects for them. Because ties are common in our setting, we use $\tau_b$ throughout.

%% file: appendix/scorio.tex
\section{\scorio{}, Open-Source Library for LLM Ranking}\label{sec:app:scorio}

\scorio{} is a Python library for ranking LLMs from repeated-trial benchmark evaluations under test-time scaling. It provides a unified interface for mapping the response tensor $\mathbf{R}\in\{0,1\}^{L\times M\times N}$ (and, where relevant, optional prior outcomes) to model scores and rankings across evaluation metrics, probabilistic paired-comparison and rating systems, voting rules, listwise choice models, item response theory, and graph- or spectral-based methods. The package can be installed with \texttt{pip install scorio}.

All ranking methods in \scorio{} operate on the response tensor $\mathbf{R}$, where $L$ is the number of models, $M$ the number of questions, and $N$ the number of trials per question. In Python, this is represented as a NumPy array of shape \texttt{(L, M, N)}. \Cref{lst:scorio:input} shows how to construct $\mathbf{R}$ and call a basic ranking method.

\begin{lstlisting}[
  caption={Constructing the response tensor and computing rankings with \scorio{}.},
  label={lst:scorio:input},
  float=t,
]
import numpy as np
from scorio import rank

# Binary response tensor: L=3 models, M=4 questions, N=5 trials
R = np.random.randint(0, 2, size=(3, 4, 5))

# Rank by mean accuracy
rankings = rank.avg(R)

# Return both rankings and scores
rankings, scores = rank.avg(R, return_scores=True)
\end{lstlisting}

Every function in the \texttt{rank} module follows the same interface: it takes the tensor $\mathbf{R}$ as the first argument, returns a ranking array of shape \texttt{(L,)}, and accepts an optional \texttt{return\_scores=True} flag to additionally return the underlying scores. Rankings are $1$-indexed, with lower values indicating better models.

\scorio{} provides a broad collection of ranking methods. \Cref{lst:scorio:eval} illustrates evaluation-based methods, including the Pass@$k$ family that quantifies how reliably models solve questions within $k$ sampled trials.

\begin{lstlisting}[
  caption={Evaluation-based ranking methods.},
  label={lst:scorio:eval},
  float=t,
]
# Pass@k: probability at least 1 of k draws succeeds
rankings, scores = rank.pass_at_k(R, k=3, return_scores=True)

# G-Pass@k with threshold tau
rankings = rank.g_pass_at_k_tau(R, k=5, tau=0.6)

# Bayesian posterior ranking with optional prior outcomes
R0 = np.random.randint(0, 2, size=(3, 4, 2))  # prior data
rankings = rank.bayes(R, R0=R0)
\end{lstlisting}

The \texttt{bayes} method generalizes beyond binary correctness to categorical outcomes $R_{lmn}\in\{0,\dots,C\}$ via a weight vector $\mathbf{w}\in\mathbb{R}^{C+1}$ that maps each category to a score. It also accepts an optional prior tensor $\mathbf{R}_0$ that incorporates outcomes from a different evaluation setting (e.g., greedy decoding) as a Bayesian prior. \Cref{lst:scorio:bayes} demonstrates both use cases.

\begin{lstlisting}[
  caption={\bayes{N} with categorical outcomes and greedy prior.},
  label={lst:scorio:bayes},
  float=t,
]
# Categorical outcomes: 0=wrong, 1=partial, 2=correct
# L=3 models, M=4 questions, N=5 trials
R_cat = np.random.randint(0, 3, size=(3, 4, 5))

# Weight vector mapping categories to scores
w = np.array([0.0, 0.5, 1.0])

rankings, scores = rank.bayes(R_cat, w=w,
                              return_scores=True)

# Using greedy decoding results as Bayesian prior
# R0 shape (M, D): shared prior across all models
R0_greedy = np.random.randint(0, 3, size=(4, 2))
rankings = rank.bayes(R_cat, w=w, R0=R0_greedy)

# Conservative ranking via posterior quantile
rankings = rank.bayes(R_cat, w=w, R0=R0_greedy,
                      quantile=0.05)
\end{lstlisting}

For probabilistic paired-comparison models, \scorio{} implements the Bradley--Terry model and its extensions, as well as Elo and TrueSkill rating systems (\cref{lst:scorio:pairwise}). These methods construct pairwise comparisons from $\mathbf{R}$ and estimate latent strength parameters.

\begin{lstlisting}[
  caption={Paired-comparison and rating system methods.},
  label={lst:scorio:pairwise},
  float=t,
]
# Bradley-Terry maximum likelihood
rankings, scores = rank.bradley_terry(R, return_scores=True)

# Bradley-Terry with MAP regularization
rankings = rank.bradley_terry_map(R, prior=1.0)

# Elo rating system
rankings, scores = rank.elo(R, K=32.0, return_scores=True)

# TrueSkill Bayesian rating
rankings = rank.trueskill(R)
\end{lstlisting}

Graph-based and spectral methods rank models by analyzing the structure of a pairwise comparison graph derived from $\mathbf{R}$, as shown in \cref{lst:scorio:graph}.

\begin{lstlisting}[
  caption={Graph-based ranking methods.},
  label={lst:scorio:graph},
  float=t,
]
# PageRank on the pairwise win-probability graph
rankings, scores = rank.pagerank(R, damping=0.85,
                                 return_scores=True)

# Spectral ranking (principal eigenvector)
rankings = rank.spectral(R)

# Rank centrality via Markov chain stationary distribution
rankings = rank.rank_centrality(R)
\end{lstlisting}

The full list of ranking methods, organized by family, is given in \cref{sec:app:apis}, and the exact method configurations used in our experiments are reported in \cref{ssec:app:setup:ranking_apis}.

\subsection{Ranking Methods}\label{sec:app:apis}

\subsubsection{Pointwise Methods}\label{ssec:app:pointwise_methods}

\paragraph{Mean accuracy.}
The simplest pointwise score is the mean accuracy
\begin{equation}
s_l^{\mathrm{mean}} := \frac{1}{M}\sum_{m=1}^M \widehat{p}_{lm},
\end{equation}
which corresponds to \texttt{avg} in \scorio.

\paragraph{Inverse-difficulty weighting.}
To emphasize hard questions, \texttt{inverse\_difficulty} weights each question
by the inverse of its global solve rate $p_m := \frac{1}{LN}\sum_{l,n} R_{lmn}$:
\begin{equation}
\begin{aligned}
w_m &\propto \frac{1}{\operatorname{clip}(p_m,\epsilon,1-\epsilon)},\\
s_l^{\mathrm{inv\text{-}diff}} &:= \sum_{m=1}^M w_m\,\widehat{p}_{lm},
\end{aligned}
\end{equation}
with weights normalized to $\sum_m w_m = 1$.

\begin{algorithm}[t]
\caption{Pointwise scoring (mean and inverse-difficulty)}
\label{alg:app:pointwise}
\footnotesize
\begin{algorithmic}[1]
\Require $R\in\{0,1\}^{L\times M\times N}$, $\epsilon>0$
\Ensure Scores $s\in\mathbb{R}^L$
\State Compute $\widehat{p}_{lm}\gets \frac{1}{N}\sum_{n=1}^N R_{lmn}$
\State \textbf{Mean:} $s_l \gets \frac{1}{M}\sum_{m=1}^M \widehat{p}_{lm}$
\State \textbf{Inv-diff:} compute $p_m\gets \frac{1}{LN}\sum_{l,n}R_{lmn}$
\State Set $w_m\propto 1/\operatorname{clip}(p_m,\epsilon,1-\epsilon)$ and normalize to $\sum_m w_m=1$
\State $s_l \gets \sum_{m} w_m\,\widehat{p}_{lm}$
\end{algorithmic}
\end{algorithm}

\subsubsection{Evaluation-metric Methods}\label{ssec:app:eval_ranking_methods}

These methods rank models by \emph{evaluation metrics} computed from per-question trial outcomes. The simplest baseline is mean accuracy (\texttt{avg}; \cref{ssec:app:pointwise_methods}); below we detail \pass{k}-family metrics and \bayes{N}. For a fixed model $l$, define the per-question success counts $\nu_{lm}:=\sum_{n=1}^N R_{lmn}$. Each metric defines a per-question score $f(\nu_{lm};N)$ (or $f(\nu_{lm};N,k,\tau)$) and then averages across questions.

\paragraph{\pass{k} (\texttt{pass\_at\_k}).}
\pass{k} \cite{chen2021evaluating} is the probability that at least one of $k$ samples is correct.
For each question $m$,
\begin{equation}
\mathrm{Pass@}k_{lm} := 1 - \frac{\binom{N-\nu_{lm}}{k}}{\binom{N}{k}},
\end{equation}
and the model-level score is $s_l^{\mathrm{Pass@}k} := \frac{1}{M}\sum_{m=1}^M \mathrm{Pass@}k_{lm}$.

\paragraph{Pass-hat@k / G-Pass@k (\texttt{pass\_hat\_k}).}
This metric (also called G-Pass@k in parts of the recent LLM evaluation literature \cite{yao2024taubench}) is the probability that \emph{all} $k$ selected samples are correct:
\begin{equation}
\widehat{\mathrm{Pass@}k}_{lm} := \frac{\binom{\nu_{lm}}{k}}{\binom{N}{k}},
\end{equation}
with $s_l^{\widehat{\mathrm{Pass@}k}} := \frac{1}{M}\sum_{m=1}^M \widehat{\mathrm{Pass@}k}_{lm}$.

\paragraph{G-Pass@k$_\tau$ (\texttt{g\_pass\_at\_k\_tau}).}
G-Pass@k$_\tau$ \cite{liu2024stable_reasoning} generalizes the above by requiring at least $j_0:=\lceil\tau k\rceil$ successes among the $k$ selected samples.
Let $X_{lm}\sim\mathrm{Hypergeom}(N,\nu_{lm},k)$ be the number of successes in a draw of size $k$ without replacement; then
\begin{equation}
\begin{aligned}
\mathrm{G\text{-}Pass@}k_{\tau,lm}
&:= \Pr(X_{lm}\ge j_0)\\
&= \sum_{j=j_0}^{k}\frac{\binom{\nu_{lm}}{j}\binom{N-\nu_{lm}}{k-j}}{\binom{N}{k}},
\end{aligned}
\end{equation}
and $s_l^{\mathrm{G\text{-}Pass@}k_\tau}:=\frac{1}{M}\sum_{m=1}^M \mathrm{G\text{-}Pass@}k_{\tau,lm}$.
\scorio{} defines the endpoint $\tau=0$ to recover \pass{k} (and for any $\tau\in(0,1/k]$ the threshold $j_0=\lceil\tau k\rceil$ equals $1$, so the expression matches \pass{k}), while $\tau=1$ recovers Pass-hat@k.

\paragraph{mG-Pass@k (\texttt{mg\_pass\_at\_k}).}
mG-Pass@k \cite{liu2024stable_reasoning} aggregates G-Pass@k$_\tau$ over $\tau\in[0.5,1]$.
In \scorio{}, we use the equivalent expectation form
\begin{equation}
\begin{aligned}
\mathrm{mG\text{-}Pass@}k_{lm}
&:= \frac{2}{k}\,\mathbb{E}\!\left[(X_{lm}-m_0)_+\right],\\
m_0 &:=\left\lceil \tfrac{k}{2}\right\rceil,
\end{aligned}
\end{equation}
where $(x)_+:=\max(x,0)$ and $X_{lm}\sim\mathrm{Hypergeom}(N,\nu_{lm},k)$ as above. The model-level score is $s_l^{\mathrm{mG\text{-}Pass@}k}:=\frac{1}{M}\sum_{m=1}^M \mathrm{mG\text{-}Pass@}k_{lm}$.

\paragraph{\bayes{N} (\texttt{bayes}).}
\bayes{N} \cite{hariri2025dontpasskbayesianframework} applies to multi-category outcomes $R_{lmn}\in\{0,\dots,C\}$ with a weight vector $w\in\mathbb{R}^{C+1}$.
For a fixed model $l$ and question $m$, let $n_{mk}:=\sum_{n=1}^N \mathbf{1}\{R_{lmn}=k\}$ be category counts.
Optionally, a prior outcome matrix $R_0\in\{0,\dots,C\}^{M\times D}$ contributes pseudo-counts $n^0_{mk}:=1+\sum_{d=1}^D \mathbf{1}\{(R_0)_{md}=k\}$ (a Dirichlet$(1,\dots,1)$ prior), giving $\nu_{mk}:=n_{mk}+n^0_{mk}$ and $T:=1+C+D+N$.
\bayes{N} returns a posterior mean $\mu_l$ and uncertainty $\sigma_l$ of the weighted score:
\begin{equation}
\mu_l = w_0 + \frac{1}{M\,T}\sum_{m=1}^M\sum_{k=0}^{C}\nu_{mk}(w_k-w_0),
\label{eq:bayesu}
\end{equation}
\begin{equation}
\begin{aligned}
\sigma_l
&= \Bigl(\frac{1}{M^2(T+1)}\sum_{m=1}^M\Bigl[
\sum_k \frac{\nu_{mk}}{T}(w_k-w_0)^2\\
&\qquad\qquad
-\Bigl(\sum_k \frac{\nu_{mk}}{T}(w_k-w_0)\Bigr)^2
\Bigr]\Bigr)^{1/2}.
\end{aligned}
\end{equation}
\scorio{} ranks by $\mu_l$ (default) or by a conservative normal-quantile score $\mu_l+\Phi^{-1}(q)\sigma_l$ for a chosen $q\in[0,1]$.

\subsubsection{Bayesian Methods}\label{ssec:app:bayesian_methods}

\paragraph{Thompson sampling ranking (\texttt{thompson}).}
Thompson sampling \cite{Thompson1933,Russo2018TS} ranks by Monte Carlo samples from a conjugate Beta--Binomial posterior over each model's aggregate success probability.
We model $p_l\sim\mathrm{Beta}(\alpha,\beta)$ and treat all $MN$ trials as i.i.d.\ Bernoulli outcomes \cite{Gelman2013BDA}.
Let $S_l:=\sum_{m=1}^M\sum_{n=1}^N R_{lmn}$ be the total number of successes for model $l$; then
\begin{equation}
p_l \mid \mathbf{R} \sim \mathrm{Beta}\!\left(\alpha + S_l,\;\beta + MN - S_l\right).
\end{equation}
For $t=1,\dots,T$ we draw $p_l^{(t)}\sim p_l\mid\mathbf{R}$ independently for each model, compute the induced rank
$r_l^{(t)}\in\{1,\dots,L\}$ (smaller is better), and score by the negative average rank
\begin{equation}
s_l^{\mathrm{TS}} := -\frac{1}{T}\sum_{t=1}^T r_l^{(t)}.
\end{equation}

\paragraph{Bayesian Bradley--Terry via MCMC (\texttt{bayesian\_mcmc}).}
To obtain a full Bayesian posterior over paired-comparison strengths, we combine the Bradley--Terry likelihood \cite{BradleyTerry1952}
with a Gaussian prior and approximate the posterior with Metropolis--Hastings sampling \cite{Metropolis1953,Hastings1970}.
We first form decisive win counts
\begin{equation}
W_{ij} := \sum_{m=1}^M\sum_{n=1}^N \mathbf{1}\{R_{imn}=1,\;R_{jmn}=0\},
\end{equation}
ignoring ties (both correct or both incorrect). Parameterizing $\pi_i=\exp(\theta_i)$, the BT likelihood is
\begin{equation}
\begin{aligned}
\Pr(i\succ j \mid \theta)
&= \frac{\exp(\theta_i)}{\exp(\theta_i)+\exp(\theta_j)},\\
\log p(\mathbf{W}\mid \theta)
&= \sum_{i\neq j} W_{ij}\log \Pr(i\succ j \mid \theta),
\end{aligned}
\end{equation}
with an independent prior $\theta_i\sim\mathcal{N}(0,\sigma^2)$ \cite{CaronDoucet2012}.
We sample from $p(\theta\mid \mathbf{W})$ and rank models by the posterior mean score
$s_i^{\mathrm{MCMC}}:=\mathbb{E}[\theta_i\mid \mathbf{W}]$.

\subsubsection{Voting-based Methods}\label{ssec:app:voting_methods}

Voting rules aggregate per-question preferences into a global ranking. To adapt them to our test-time-scaling setting, we treat each question $m$ as a
``voter'' that ranks models by their per-question solve frequency across trials:
\begin{equation}
k_{lm} := \sum_{n=1}^N R_{lmn}\in\{0,1,\dots,N\}.
\end{equation}
When $N=1$, each question induces only a two-level ranking (correct vs.\ incorrect),
so Borda/Copeland reduce to (ties of) accuracy-based ordering; when $N>1$ these
rules exploit the additional resolution from $k_{lm}$.

\paragraph{Borda count.}
For each question $m$, let $r_{lm}\in\{1,\dots,L\}$ be the (tie-averaged) rank of
model $l$ when sorting $k_{\cdot m}$ in descending order (smaller rank is better).
The Borda score is
\begin{equation}
s_l^{\mathrm{Borda}} := \sum_{m=1}^M (L - r_{lm}),
\end{equation}
which assigns $(L-1)$ points for a unique first place and $0$ for a unique last
place, with ties receiving the average of the tied positions \cite{Borda1781,brandt2016compsocchoice}.

\paragraph{Copeland.}
For each pair $(i,j)$, define the number of questions that prefer $i$ to $j$ as
$W^{(q)}_{ij}:=\sum_m \mathbb{I}[k_{im}>k_{jm}]$.
Copeland declares $i$ to beat $j$ if $W^{(q)}_{ij}>W^{(q)}_{ji}$ and scores each
model by net pairwise dominance:
\begin{equation}
s_i^{\mathrm{Copeland}} := \sum_{j\neq i}\mathrm{sign}\!\left(W^{(q)}_{ij}-W^{(q)}_{ji}\right),
\end{equation}
where $\mathrm{sign}(0)=0$ \cite{Copeland1951,brandt2016compsocchoice}.

\paragraph{Win rate.}
Using the same question-level win counts $W^{(q)}$, define a model's win rate as
the fraction of decisive pairwise outcomes it wins:
\begin{equation}
s_i^{\mathrm{winrate}} :=
\frac{\sum_{j\neq i} W^{(q)}_{ij}}{\sum_{j\neq i} \left(W^{(q)}_{ij}+W^{(q)}_{ji}\right)},
\end{equation}
with the convention $s_i^{\mathrm{winrate}}=0.5$ if the denominator is zero.

\paragraph{Condorcet-style pairwise-majority rules.}
Many voting rules are defined from an aggregated pairwise preference matrix.
To incorporate per-question ties when $k_{im}=k_{jm}$, we define
\begin{equation}
P^{(q)}_{ij} := \sum_{m=1}^M \Big(\mathbb{I}[k_{im}>k_{jm}] + \tfrac{1}{2}\mathbb{I}[k_{im}=k_{jm}]\Big),
\end{equation}
so that $P^{(q)}_{ij}+P^{(q)}_{ji}=M$. Let margins be $\Delta_{ij}:=P^{(q)}_{ij}-P^{(q)}_{ji}$.

\paragraph{Minimax (Simpson--Kramer).}
The minimax score is based on a model's worst pairwise defeat:
\begin{equation}
s_i^{\mathrm{minimax}} := -\max_{j\neq i}\max(0,\Delta_{ji}),
\end{equation}
and ranks models by the size of their worst defeat (closer to $0$ is better) \cite{brandt2016compsocchoice}.

\paragraph{Schulze (beatpath).}
Schulze computes strongest-path strengths $p_{ij}$ in the directed graph of pairwise victories and ranks $i$ above $j$ if $p_{ij}>p_{ji}$ \cite{Schulze2010,brandt2016compsocchoice}.

\paragraph{Ranked Pairs (Tideman).}
Ranked Pairs sorts pairwise victories by strength (e.g., margin $\Delta_{ij}$), then locks them in that order whenever doing so does not introduce a cycle; the resulting acyclic dominance graph induces a ranking \cite{Tideman1987,brandt2016compsocchoice}.

\paragraph{Kemeny--Young.}
Kemeny--Young returns an ordering $\pi$ that maximizes agreement with the pairwise preferences:
\begin{equation}
\pi \in \arg\max_{\text{total orders } \pi}\;\sum_{i\prec_\pi j} P^{(q)}_{ij},
\end{equation}
which is equivalent to a maximum-likelihood ranking under certain noise models and is a classic Condorcet extension \cite{Kemeny1959,Young1977,brandt2016compsocchoice}. (Exact optimization is NP-hard in general; we solve the induced linear ordering problem via MILP for the problem sizes in this paper.)

\paragraph{Borda elimination rules (Nanson and Baldwin).}
Nanson's method iteratively recomputes Borda scores over remaining candidates and removes those below the mean, while Baldwin's method removes the lowest Borda scorer(s) each round \cite{Nanson1883,Baldwin1926,brandt2016compsocchoice}.

\paragraph{Majority Judgment.}
Majority Judgment treats $k_{lm}\in\{0,\dots,N\}$ as discrete grades and ranks models by their median grade, breaking ties using the majority-gauge rule \cite{BalinskiLaraki2011}.

\begin{algorithm}[t]
\caption{Voting rules on per-question trial counts}
\label{alg:app:voting}
\footnotesize
\begin{algorithmic}[1]
\Require $R\in\{0,1\}^{L\times M\times N}$
\Ensure Borda scores $s^{\mathrm{Borda}}$, Copeland scores $s^{\mathrm{Copeland}}$, win-rate scores $s^{\mathrm{winrate}}$
\State Compute $k_{lm}\gets \sum_{n=1}^N R_{lmn}$
\State $s^{\mathrm{Borda}}\gets 0$; $s^{\mathrm{Copeland}}\gets 0$; initialize $W^{(q)}\gets 0$
\For{$m=1$ \textbf{to} $M$}
  \State Rank models by $k_{\cdot m}$ (descending) with average-tie ranks $r_{\cdot m}$
  \State $s_l^{\mathrm{Borda}} \mathrel{+}= L - r_{lm}$ for all $l$
\EndFor
\For{$1\le i<j\le L$}
  \State $W^{(q)}_{ij}\gets \sum_m \mathbb{I}[k_{im}>k_{jm}]$
  \State $W^{(q)}_{ji}\gets \sum_m \mathbb{I}[k_{jm}>k_{im}]$
  \If{$W^{(q)}_{ij}>W^{(q)}_{ji}$} $s_i^{\mathrm{Copeland}}\mathrel{+}=1$; $s_j^{\mathrm{Copeland}}\mathrel{-}=1$
  \ElsIf{$W^{(q)}_{ji}>W^{(q)}_{ij}$} $s_i^{\mathrm{Copeland}}\mathrel{-}=1$; $s_j^{\mathrm{Copeland}}\mathrel{+}=1$
  \EndIf
\EndFor
\State $s_i^{\mathrm{winrate}}\gets \frac{\sum_{j\neq i} W^{(q)}_{ij}}{\sum_{j\neq i}(W^{(q)}_{ij}+W^{(q)}_{ji})}$ (or $0.5$ if denominator is $0$)
\end{algorithmic}
\end{algorithm}

\subsubsection{Paired-comparison Probabilistic Models}\label{ssec:app:paired_models}

These methods first reduce $\mathbf{R}$ to pairwise win/tie counts between
models, then fit a parametric paired-comparison model.
For each ordered pair $(i,j)$, define wins $W_{ij}$ and ties $T_{ij}$ as in
\cref{ssec:representation} (pairwise representation).

\paragraph{Bradley--Terry (BT).}
The BT model \cite{BradleyTerry1952} assigns each model a positive strength
$\pi_i>0$ and assumes
\begin{equation}
\Pr(i\succ j) = \frac{\pi_i}{\pi_i+\pi_j}.
\end{equation}
Given win counts $W_{ij}$, the log-likelihood is
\begin{equation}
\log p(\mathbf{W}\mid \pi) = \sum_{i\neq j} W_{ij}\Big[\log \pi_i - \log(\pi_i+\pi_j)\Big],
\label{eq:bt-loglik}
\end{equation}
with identifiability enforced by centering log-strengths.
\scorio{} provides ML (\texttt{bradley\_terry}) and MAP (\texttt{bradley\_terry\_map}) estimation; MAP adds a prior penalty on log-strengths (e.g., Gaussian) \cite{CaronDoucet2012}.

\paragraph{Tie extensions.}
In our binary setting, a pairwise tie occurs when both models are correct or
both are incorrect on the same question--trial. \scorio{} implements two classic
tie models:
\begin{itemize}[leftmargin=*]
  \item \textbf{Davidson} \cite{davidson1970bties}: adds a tie parameter and models $(i\succ j)$, $(j\succ i)$, and $(i\sim j)$ explicitly (\texttt{bradley\_terry\_davidson}, \texttt{bradley\_terry\_davidson\_map}).
  \item \textbf{Rao--Kupper} \cite{rao1967ties}: alternative tie parameterization via $\kappa\ge 1$ (\texttt{rao\_kupper}, \texttt{rao\_kupper\_map}).
\end{itemize}
For Davidson, with tie parameter $\nu>0$,
\begin{equation}
\begin{aligned}
\Pr(i\succ j) &= \frac{\pi_i}{\pi_i+\pi_j+\nu\sqrt{\pi_i\pi_j}},\\
\Pr(j\succ i) &= \frac{\pi_j}{\pi_i+\pi_j+\nu\sqrt{\pi_i\pi_j}},\\
\Pr(i\sim j) &= \frac{\nu\sqrt{\pi_i\pi_j}}{\pi_i+\pi_j+\nu\sqrt{\pi_i\pi_j}}.
\end{aligned}
\end{equation}
For Rao--Kupper, with $\kappa\ge 1$,
\begin{equation}
\begin{aligned}
\Pr(i\succ j) &= \frac{\pi_i}{\pi_i+\kappa\pi_j},\\
\Pr(j\succ i) &= \frac{\pi_j}{\kappa\pi_i+\pi_j},\\
\Pr(i\sim j) &= \frac{(\kappa^2-1)\pi_i\pi_j}{(\pi_i+\kappa\pi_j)(\kappa\pi_i+\pi_j)}.
\end{aligned}
\end{equation}

\begin{algorithm}[t]
\caption{Paired-comparison models (BT, Davidson, Rao--Kupper) via ML/MAP}
\label{alg:app:paired_models}
\footnotesize
\begin{algorithmic}[1]
\Require $R\in\{0,1\}^{L\times M\times N}$; model family; optional prior penalty on log-strengths; max iterations $T$
\Ensure Scores (strengths) $\hat{\pi}\in\mathbb{R}_+^L$
\State Compute pairwise win/tie counts $(W_{ij},T_{ij})$ from $R$
\State Parameterize strengths by log-strengths $\theta_i=\log \pi_i$ and enforce identifiability by centering: $\theta\leftarrow \theta-\frac{1}{L}\sum_i\theta_i$
\State Define the family-specific log-likelihood $\log p(W,T\mid \theta,\text{tie-params})$
\State Define objective $\mathcal{L}=-\log p(\cdot)+\text{prior}(\theta)$ (prior term is $0$ for ML)
\State Optimize $\mathcal{L}$ with L-BFGS for up to $T$ iterations
\State Return $\hat{\pi}_i=\exp(\hat{\theta}_i)$ as scores (larger is better)
\end{algorithmic}
\end{algorithm}

\subsubsection{Sequential Rating Systems}\label{ssec:app:rating_systems}

Sequential rating systems process a stream of head-to-head ``matches'' rather than aggregating all pairwise outcomes into a single count matrix. In our benchmark setting, the natural match stream is induced by each question--trial $(m,n)$: for every pair of models $(i,j)$, we observe a binary outcome pair $(R_{imn},R_{jmn})\in\{0,1\}^2$ and declare $i$ to beat $j$ if $(1,0)$ and $j$ to beat $i$ if $(0,1)$. When $(R_{imn},R_{jmn})\in\{(1,1),(0,0)\}$, the comparison is a tie; \scorio{} exposes tie-handling policies (e.g., treat ties as draws or ignore certain ties) for these methods.

\paragraph{Elo.}
Elo \cite{Elo1978} maintains a scalar rating $r_i$ for each model. For a match between $i$ and $j$, define the expected score
\begin{equation}
E_{ij} := \frac{1}{1 + 10^{(r_j-r_i)/400}},
\end{equation}
and let $S_{ij}\in\{0,\tfrac{1}{2},1\}$ be the realized match score for $i$ against $j$ (win/draw/loss, depending on the tie-handling rule). The sequential Elo update is
\begin{equation}
\begin{aligned}
r_i &\leftarrow r_i + K(S_{ij}-E_{ij}),\\
r_j &\leftarrow r_j + K((1-S_{ij})-(1-E_{ij})),
\end{aligned}
\end{equation}
with learning rate $K>0$ (\texttt{elo} in \scorio). Because the updates are sequential, the final ratings can depend on the order in which the match stream is processed.

\paragraph{Glicko.}
Glicko \cite{Glickman1999} augments Elo with an uncertainty parameter (rating deviation) $\mathrm{RD}_i$ and updates ratings using batches of matches within rating periods. In our implementation, each question--trial $(m,n)$ constitutes one rating period containing all pairwise matches on that $(m,n)$. Define $q:=\ln(10)/400$ and
\begin{equation}
g(\mathrm{RD}) := \frac{1}{\sqrt{1+\frac{3q^2\mathrm{RD}^2}{\pi^2}}}.
\end{equation}
For a player $i$ in a rating period with opponents $j\in\mathcal{O}_i$ and outcomes $S_{ij}$, define expected scores
\begin{equation}
E_{ij} := \frac{1}{1 + 10^{-g(\mathrm{RD}_j)(r_i-r_j)/400}},
\end{equation}
and
\begin{equation}
d_i^2 := \left(q^2\sum_{j\in\mathcal{O}_i} g(\mathrm{RD}_j)^2 E_{ij}(1-E_{ij})\right)^{-1}.
\end{equation}
The Glicko updates are
\begin{equation}
\begin{aligned}
\mathrm{RD}_i' &:= \left(\frac{1}{\mathrm{RD}_i^2} + \frac{1}{d_i^2}\right)^{-1/2},\\
r_i' &:= r_i + \frac{q}{\frac{1}{\mathrm{RD}_i^2}+\frac{1}{d_i^2}}
\sum_{j\in\mathcal{O}_i} g(\mathrm{RD}_j)(S_{ij}-E_{ij}),
\end{aligned}
\end{equation}
with optional $\mathrm{RD}$ inflation between rating periods and a maximum $\mathrm{RD}$ cap (as in the original Glicko specification). This corresponds to \texttt{glicko} in \scorio; we rank by $r_i'$ (larger is better), and $\mathrm{RD}_i'$ can be used as an uncertainty summary.

\paragraph{TrueSkill.}
TrueSkill \cite{herbrich2006trueskill} is a Bayesian rating system that models each model's latent skill as a Gaussian $\mathcal{N}(\mu_i,\sigma_i^2)$ and updates $(\mu_i,\sigma_i)$ after each match using approximate inference. In \scorio, we apply a two-player TrueSkill update to each decisive $(1,0)$ or $(0,1)$ pairwise match in the induced stream (ties are ignored) and return the final $\mu_i$ as the score (\texttt{trueskill}); a per-round dynamics parameter $\tau$ inflates $\sigma$ between rounds to model drift.

\subsubsection{Listwise / Setwise Choice Models (Luce Family)}\label{ssec:app:luce_models}

Unlike pairwise models, these methods operate on \emph{setwise} events induced
by each question--trial $(m,n)$. Define the winner and loser sets
\begin{equation}
\begin{aligned}
U_{mn} &:= \{l: R_{lmn}=1\},\\
V_{mn} &:= \{l: R_{lmn}=0\}.
\end{aligned}
\end{equation}
If $U_{mn}=\emptyset$ or $U_{mn}=\mathcal{L}$, the event contains no ranking
information and is discarded.

\paragraph{Plackett--Luce (PL).}
The PL model \cite{plackett1975permutations,luce1959choice} is a listwise
generalization of BT for full rankings. In our binary setting we apply PL to
the pairwise win matrix (equivalently BT) and estimate strengths using the
MM update from \citet{hunter2004mmbradleyterry} (\texttt{plackett\_luce},
\texttt{plackett\_luce\_map}).

\paragraph{Davidson--Luce (setwise ties).}
Davidson--Luce \cite{firth2019davidsonluce} models the probability of a tied
winner set $U_{mn}$ emerging from the full set $U_{mn}\cup V_{mn}$, explicitly
accounting for ties within $U_{mn}$ and $V_{mn}$ (\texttt{davidson\_luce},
\texttt{davidson\_luce\_map}).
Let $\pi_i>0$ be strengths and $\delta_t>0$ be tie-prevalence parameters with
\(\delta_1\equiv 1\). For a comparison set $S$ and tie order $t$, define
$g_t(T):=\left(\prod_{i\in T}\pi_i\right)^{1/t}$ and
\begin{equation}
\begin{aligned}
Z(S) := {}&
\sum_{t'=1}^{\min(D,|S|)} \delta_{t'} \\
&\quad \cdot \sum_{\substack{T\subseteq S\\|T|=t'}} g_{t'}(T),
\end{aligned}
\end{equation}
where $D$ is the maximum tie order considered.
Then, for an event $(U,V)$ with $S=U\cup V$ and $t=|U|$,
\begin{equation}
\Pr(U\succ V\mid S) = \frac{\delta_t\,g_t(U)}{Z(S)}.
\end{equation}

\paragraph{Bradley--Terry--Luce (BTL) setwise-choice construction.}
BTL converts each winner $i\in U_{mn}$ into a Luce choice event from
$\{i\}\cup V_{mn}$, with choice probability
$\Pr(i\mid \{i\}\cup V)=\pi_i/(\pi_i+\sum_{j\in V}\pi_j)$ (\texttt{bradley\_terry\_luce}, \texttt{bradley\_terry\_luce\_map}).
Equivalently, for an event $(U,V)$ the BTL likelihood factorizes as
\begin{equation}
\Pr(U\succ V) = \prod_{i\in U}\frac{\pi_i}{\pi_i+\sum_{j\in V}\pi_j}.
\end{equation}

\begin{algorithm}[t]
\caption{MM algorithm for PL/BT on the pairwise win matrix}
\label{alg:app:pl_mm}
\begin{algorithmic}[1]
\Require Pairwise win matrix $W\in\mathbb{R}_+^{L\times L}$, iterations $T$
\Ensure Strengths $\hat{\pi}\in\mathbb{R}_+^L$ (normalized)
\State $w_i\gets \sum_j W_{ij}$ (total wins); $n_{ij}\gets W_{ij}+W_{ji}$ (total comparisons)
\State Initialize $\pi_i\propto w_i$ and normalize $\sum_i\pi_i=1$
\For{$t=1$ \textbf{to} $T$}
  \For{$i=1$ \textbf{to} $L$}
    \State $d_i \gets \sum_{j\neq i:\,n_{ij}>0} \frac{n_{ij}}{\pi_i+\pi_j}$
    \State $\pi_i \gets w_i / d_i$
  \EndFor
  \State Normalize $\pi$ to sum to $1$
\EndFor
\State Return $\pi$
\end{algorithmic}
\end{algorithm}

\begin{algorithm}[t]
\caption{Setwise event extraction and Luce-family estimation (Davidson--Luce / BTL)}
\label{alg:app:setwise_models}
\begin{algorithmic}[1]
\Require $R\in\{0,1\}^{L\times M\times N}$; model type $\in\{\text{Davidson--Luce},\text{BTL}\}$; optional prior on log-strengths; max iterations $T$
\Ensure Strength scores $\hat{\pi}\in\mathbb{R}_+^L$
\State Build events $\mathcal{E}\gets\{(U_{mn},V_{mn}) : 0<|U_{mn}|<L\}$
\State Parameterize $\pi_i=\exp(\theta_i)$ with centered $\theta$ for identifiability
\State Define the event log-likelihood $\sum_{(U,V)\in\mathcal{E}}\log p(U\succ V\mid \theta)$ for the chosen model
\State Add prior penalty on $\theta$ for MAP (or $0$ for ML)
\State Optimize with L-BFGS for up to $T$ iterations and return $\hat{\pi}$
\end{algorithmic}
\end{algorithm}

\subsubsection{Item Response Theory (IRT) Methods}\label{ssec:app:irt_methods}

\scorio{} includes several IRT-inspired ranking methods that treat each model as an ``examinee'' with a latent ability and each question as an ``item'' with latent parameters (e.g., difficulty). We use IRT primarily as a \emph{ranking model}: we estimate abilities $\{\theta_l\}_{l=1}^L$ and rank models by $\theta_l$ (larger is better), using $\texttt{rank\_scores}$ for tie-aware rank variants.

\paragraph{Data and binomial reduction.}
Our raw observations are binary trial outcomes $R_{lmn}\in\{0,1\}$ for model $l\in\{1,\dots,L\}$, question $m\in\{1,\dots,M\}$, and trial $n\in\{1,\dots,N\}$. When trials are i.i.d.\ conditional on parameters, the sufficient statistic for an item-model pair is the correct-count
\begin{equation}
k_{lm} := \sum_{n=1}^N R_{lmn} \in \{0,1,\dots,N\},
\end{equation}
so that likelihood-based IRT estimation can be written as a binomial-response model \cite{mccullaghnelder1989glm,deboeckwilson2004explanatory}.

\paragraph{Rasch (1PL).}
The Rasch model \cite{rasch1960rasch} assumes a single item parameter (difficulty $b_m$):
\begin{equation}
k_{lm} \sim \mathrm{Binomial}\!\left(N,\;\sigma(\theta_l - b_m)\right),
\end{equation}
where $\sigma(x)=1/(1+e^{-x})$. The model is invariant to global shifts $(\theta,b)\mapsto(\theta+c,b+c)$, so we impose an identifiability constraint by centering item difficulties (e.g., $\sum_m b_m = 0$).

\paragraph{2PL and 3PL.}
The 2PL model \cite{birnbaum1968latent} adds an item discrimination parameter $a_m>0$:
\begin{equation}
k_{lm} \sim \mathrm{Binomial}\!\left(N,\;\sigma\!\big(a_m(\theta_l - b_m)\big)\right).
\end{equation}
The 3PL model further adds a pseudo-guessing parameter $c_m$:
\begin{equation}
\begin{split}
k_{lm} &\sim \mathrm{Binomial}\!\left(N,\;p_{lm}\right),\\
p_{lm} &:= c_m + (1-c_m)\sigma\!\big(a_m(\theta_l - b_m)\big).
\end{split}
\end{equation}
In our implementation, we constrain $a_m$ via a log-parameterization and keep $c_m$ in a bounded range (or optionally fix $c_m$ to a known chance level).

\paragraph{Estimation variants used in \scorio.}
\begin{itemize}[leftmargin=*]
  \item \textbf{JMLE / MLE} (\texttt{rasch}, \texttt{rasch\_2pl}, \texttt{rasch\_3pl}): optimize the joint log-likelihood over $\theta$ and item parameters.
  \item \textbf{MAP} (\texttt{rasch\_map}, \texttt{rasch\_2pl\_map}, \texttt{rasch\_3pl\_map}): add a prior penalty on abilities, typically Gaussian, as in Bayes modal estimation \cite{Mislevy1986}.
  \item \textbf{MML + EAP} (\texttt{rasch\_mml}): integrate out abilities under a population model (we use a standard normal prior), fit item parameters by EM, then compute EAP ability estimates \cite{BockAitkin1981,ChenHouDodd1998}.
  \item \textbf{Credible/LB scoring} (\texttt{rasch\_mml\_credible}): rank by a posterior quantile of $\theta_l$ (e.g., a lower bound), which yields a conservative, uncertainty-aware ranking.
  \item \textbf{Dynamic IRT} (\texttt{dynamic\_irt}): a longitudinal extension that allows per-model trends across trials \cite{verhelst1993dynamicrasch,wang2019longitudinalirt}.
\end{itemize}

\begin{algorithm}[t]
\caption{Binomial xPL IRT (JMLE/MAP) for ranking}
\label{alg:app:irt_xpl}
\begin{algorithmic}[1]
\Require Response tensor $R\in\{0,1\}^{L\times M\times N}$; model type $\in\{\text{1PL},\text{2PL},\text{3PL}\}$; optional ability prior $p(\theta)$; max iterations $T$
\Ensure Ability scores $\hat{\theta}\in\mathbb{R}^{L}$ and optional item parameters
\State Compute counts $k_{lm}\gets \sum_{n=1}^N R_{lmn}$ and set $n\gets N$
\State Initialize $\theta$ from per-model accuracy; initialize $b$ from per-item solve rate; set $a_m\gets 1$ (2PL/3PL); set $c_m\gets 0.25$ (3PL)
\State Define $p_{lm}(\theta,b,a,c)$ according to the chosen xPL link
\State Define the binomial log-likelihood $\ell(k;n,p)\gets k\log p + (n-k)\log(1-p)$
\State Define objective (negative log posterior)
\[
\begin{aligned}
\mathcal{L}(\theta,b,a,c) ={}& -\sum_{l,m} \ell(k_{lm}; n, p_{lm}) \\
&- \log p(\theta).
\end{aligned}
\]
\Statex Set $\log p(\theta)=0$ for pure MLE.
\State Impose identifiability at each iteration by centering item difficulties: $b \leftarrow b - \frac{1}{M}\sum_m b_m$
\State Optimize $\mathcal{L}$ with a quasi-Newton method (e.g., L-BFGS) for up to $T$ iterations
\State Return $\hat{\theta}$ as scores (larger is better) and optionally $\hat{b}$, $\hat{a}$, $\hat{c}$
\end{algorithmic}
\end{algorithm}

\begin{algorithm}[t]
\caption{Rasch MML (EM + quadrature) with EAP and posterior-quantile scoring}
\label{alg:app:irt_mml}
\begin{algorithmic}[1]
\Require Counts $k\in\{0,\dots,N\}^{L\times M}$; trials $N$; quadrature points $\{\theta_q,w_q\}_{q=1}^Q$; EM iterations $S$
\Ensure EAP scores $\hat{\theta}^{\mathrm{EAP}}$ (or quantile scores) and item difficulties $\hat{b}$
\State Initialize item difficulties $b$ from per-item solve rates and center $b$
\For{$s=1$ \textbf{to} $S$}
  \State \textbf{E-step:} compute $\log p(k_l \mid \theta_q,b)$ for each model $l$ and quadrature point $q$
  \State Compute posterior weights $w_{lq}\propto \exp(\log p(k_l\mid \theta_q,b))\,w_q$ and normalize over $q$
  \State Define $\ell(k;n,p)\gets k\log p + (n-k)\log(1-p)$
  \State \textbf{M-step:} for each item $m$, update $b_m$ by minimizing
  \[
  -\sum_{l,q} w_{lq}\,\ell\!\left(k_{lm};N,\sigma(\theta_q-b_m)\right).
  \]
  \State Center $b$
\EndFor
\State Recompute posterior weights $w_{lq}$ under final $b$
\State Compute EAP scores: $\hat{\theta}^{\mathrm{EAP}}_l \gets \sum_q w_{lq}\theta_q$
\State (Optional) Compute quantile score $Q_\alpha(\theta_l\mid k)$ from the discrete posterior CDF (used by \texttt{rasch\_mml\_credible})
\State Return scores and $\hat{b}$
\end{algorithmic}
\end{algorithm}

\begin{algorithm}[t]
\caption{Dynamic IRT growth model (logistic longitudinal Rasch)}
\label{alg:app:irt_dynamic_growth}
\begin{algorithmic}[1]
\Require Response tensor $R\in\{0,1\}^{L\times M\times N}$; normalized time grid $t_n\in[0,1]$; max iterations $T$
\Ensure Baseline abilities $\hat{\theta}_0\in\mathbb{R}^L$, slopes $\hat{\theta}_1\in\mathbb{R}^L$, and item difficulties $\hat{b}\in\mathbb{R}^M$
\State Fit the longitudinal model $P(R_{lmn}=1)=\sigma(\theta_{0,l}+\theta_{1,l}t_n-b_m)$ by maximizing the Bernoulli likelihood over all $(l,m,n)$
\State Add weak regularization on slopes (e.g., $\|\theta_1\|_2^2$) to avoid overfitting i.i.d.\ sampling noise
\State Center $b$ for identifiability
\State Optimize with a quasi-Newton method (e.g., L-BFGS) for up to $T$ iterations
\State Return $\hat{\theta}_0$ as ranking scores and optionally $\hat{\theta}_1,\hat{b}$
\end{algorithmic}
\end{algorithm}

\subsubsection{Graph and Spectral Methods}\label{ssec:app:graph_methods}

These methods operate on the pairwise comparison graph derived from the win/tie counts $(W_{ij},T_{ij})$ defined in \cref{ssec:representation}. A common derived quantity is the empirical tied-split win probability
\begin{equation}
\widehat{P}_{i\succ j} := \frac{W_{ij}+\tfrac{1}{2}T_{ij}}{W_{ij}+W_{ji}+T_{ij}},\qquad \widehat{P}_{i\succ i}:=\tfrac{1}{2}.
\end{equation}
In our fully observed benchmark setting, $W_{ij}+W_{ji}+T_{ij}=MN$ for all $i\neq j$ (\cref{ssec:representation}), so $\widehat{P}_{i\succ j}$ is a simple rescaling of aggregated counts.

\paragraph{PageRank.}
We build a directed weighted graph where an edge from $j$ to $i$ has weight $\widehat{P}_{i\succ j}$ (interpreting ``losers link to winners''), then form a column-stochastic transition matrix $P$ by normalizing each column:
\begin{equation}
P_{ij} := \frac{\widehat{P}_{i\succ j}}{\sum_{k\neq j}\widehat{P}_{k\succ j}}\quad (i\neq j),
\end{equation}
with the standard dangling-node convention of a uniform column if the denominator is zero. PageRank scores $r\in\Delta^{L-1}$ solve
\begin{equation}
r = d\,P\,r + (1-d)\,\tfrac{1}{L}\mathbf{1},
\end{equation}
where $d\in(0,1)$ is the damping factor and $\mathbf{1}$ is the all-ones vector \cite{page1999pagerank}. This corresponds to \texttt{pagerank} in \scorio.

\paragraph{Spectral (eigenvector centrality).}
We form the nonnegative matrix $W$ with off-diagonal entries $W_{ij}:=\widehat{P}_{i\succ j}$ and set the diagonal to the row sum $W_{ii}:=\sum_{j\neq i}W_{ij}$ (a self-loop that makes the matrix diagonally dominant). The spectral score vector is the principal right eigenvector $v\ge 0$ of $W$, normalized to $\sum_i v_i=1$. This corresponds to \texttt{spectral} in \scorio.

\paragraph{Rank Centrality.}
Rank Centrality \cite{negahban2017rankcentrality} constructs a random walk on the comparison graph whose transition probabilities prefer moving from a model to those that beat it. Let $d_{\max}$ be the maximum (undirected) degree of the comparison graph (in our benchmark setting $d_{\max}=L-1$). Define a row-stochastic matrix
\begin{equation}
\begin{aligned}
P_{ij} &:= \frac{1}{d_{\max}}\widehat{P}_{j\succ i}\quad (i\neq j),\\
P_{ii} &:= 1-\sum_{j\neq i}P_{ij}.
\end{aligned}
\end{equation}
The stationary distribution $\pi$ of $P$ is used as the score vector (larger $\pi_i$ is better). This corresponds to \texttt{rank\_centrality} in \scorio.

\paragraph{$\alpha$-Rank.}
$\alpha$-Rank \cite{omidshafiei2019alpharank} ranks strategies via evolutionary dynamics by constructing a Markov chain over models using fixation probabilities in a finite population. In our constant-sum binary evaluation setting, we treat $\widehat{P}_{i\succ j}$ as the payoff to strategy $i$ against $j$ (so the per-match payoff sum is $1$ when ties are split as $\tfrac{1}{2}$). For population size $m\ge 2$ and selection intensity $\alpha\ge 0$, the (constant-sum) fixation probability of a mutant $r$ in a resident population $s$ is
\begin{equation}
\rho_{r,s} :=
\begin{cases}
\frac{1-\exp(-u)}{1-\exp(-m u)} & u\neq 0,\\
\frac{1}{m} & u=0,
\end{cases}
\end{equation}
\begin{equation}
u := \alpha\frac{m}{m-1}\left(\widehat{P}_{r\succ s}-\tfrac{1}{2}\right).
\end{equation}
The induced Markov chain on models has off-diagonal transitions $C_{s r}:=\frac{1}{L-1}\rho_{r,s}$ and diagonal $C_{ss}:=1-\sum_{r\neq s}C_{sr}$; the stationary distribution of $C$ is the $\alpha$-Rank score vector. This corresponds to \texttt{alpharank} in \scorio.

\paragraph{Nash equilibrium mixture.}
Following the use of Nash equilibria as evaluation summaries in symmetric zero-sum games \cite{balduzzi2019openended}, we define a zero-sum payoff matrix
\begin{equation}
A_{ij} := 2\widehat{P}_{i\succ j}-1,\qquad A_{ii}:=0,
\end{equation}
which is antisymmetric when $\widehat{P}$ is derived from tied-split win rates. We compute a maximin mixed strategy $x\in\Delta^{L-1}$ (a Nash equilibrium strategy for the row player)
\begin{equation}
x \in \arg\max_{x\in\Delta^{L-1}} \min_{y\in\Delta^{L-1}} x^\top A y,
\end{equation}
via a standard linear program. To obtain a per-model evaluation score (``Nash averaging''), we then score each model by its expected performance against the equilibrium mixture opponent:
\begin{equation}
s_i := \sum_{j=1}^L \widehat{P}_{i\succ j}\,x_j \in [0,1],
\end{equation}
and rank models by $s$ (higher is better). We additionally report the equilibrium mixture $x$ as a strategic summary of the meta-game when needed. This corresponds to \texttt{nash} in \scorio.

\subsubsection{Seriation-based Methods}\label{ssec:app:seriation_methods}

\paragraph{SerialRank.}
SerialRank \cite{fogel2016serialrank} is a spectral seriation method that constructs a similarity graph from a skew-symmetric comparison matrix. From pairwise counts $(W,T)$, define
\begin{equation}
\begin{aligned}
C_{ij}&:=\frac{W_{ij}-W_{ji}}{W_{ij}+W_{ji}+T_{ij}}\in[-1,1],\\
C_{ii}&:=0,
\end{aligned}
\end{equation}
so that $C_{ij}>0$ indicates $i$ tends to beat $j$ (and $C$ is skew-symmetric). SerialRank forms the similarity matrix
\begin{equation}
S := \tfrac{1}{2}\left(L\,\mathbf{1}\mathbf{1}^\top + C C^\top\right),
\end{equation}
then computes the graph Laplacian $L_S:=\mathrm{diag}(S\mathbf{1})-S$. The ordering is given by sorting a Fiedler vector (the eigenvector associated with the second-smallest eigenvalue of $L_S$), with the sign chosen to best agree with the observed comparisons. This corresponds to \texttt{serial\_rank} in \scorio.

\subsubsection{Hodge-theoretic Methods}\label{ssec:app:hodge_methods}

\paragraph{HodgeRank.}
HodgeRank \cite{jiang2009hodgerank} interprets pairwise comparisons as a skew-symmetric edge flow on a graph and recovers global scores by least squares. Using the same tied-split probabilities as above, define the observed edge flow
\begin{equation}
\begin{aligned}
\overline{Y}_{ij}
&:= \widehat{P}_{j\succ i}-\widehat{P}_{i\succ j}
 = \frac{W_{ji}-W_{ij}}{W_{ij}+W_{ji}+T_{ij}},\\
\overline{Y}_{ii} &:=0,
\end{aligned}
\end{equation}
and choose symmetric edge weights $w_{ij}$ (e.g., the total number of comparisons on edge $(i,j)$). HodgeRank solves the weighted least-squares problem
\begin{equation}
\begin{aligned}
s^\star
&\in \arg\min_{s\in\mathbb{R}^L}\;\sum_{i<j} w_{ij}\left((s_j-s_i)-\overline{Y}_{ij}\right)^2\\
&= \arg\min_s \|\mathrm{grad}(s)-\overline{Y}\|_{2,w}^2,
\end{aligned}
\end{equation}
which reduces to a weighted graph Laplacian system; we compute the minimum-norm solution via the Moore--Penrose pseudoinverse and rank by $s^\star$ (higher is better). This corresponds to \texttt{hodge\_rank} in \scorio.

\subsection{Ranking Method APIs and Hyperparameters}\label{ssec:app:setup:ranking_apis}
We evaluate the ranking methods described in \cref{sec:app:apis}. Each method maps the trial outcome tensor
$R\in\{0,1\}^{L\times M\times N}$ (and, where applicable, an optional prior tensor $R_0$) to a ranking over the
$L$ models. For reproducibility, we list the exact API identifiers and argument values used in our experiments;
\texttt{None} denotes an unset optional argument.

\paragraph{Metrics.}
\begin{itemize}[leftmargin=*,noitemsep,topsep=0pt]
\item \texttt{avg}
\item \texttt{pass\_at\_k\_2 (k=2)}
\item \texttt{pass\_hat\_k\_2 (k=2)}
\item \texttt{mg\_pass\_at\_k\_2 (k=2)}
\item \texttt{bayes (R0=None, quantile=None)}
\item \texttt{bayes\_greedy (R0=R0, quantile=None)}
\item \texttt{bayes\_ci (R0=None, quantile=0.05)}
\item \texttt{inverse\_difficulty (return\_scores=false, clip\_range=[0.01, 0.99])}
\end{itemize}

\paragraph{Pairwise rating.}
\begin{itemize}[leftmargin=*,noitemsep,topsep=0pt]
\item \texttt{elo\_tie\_skip (K=0.05, initial\_rating=1500.0, tie\_handling=skip)}
\item \texttt{elo\_tie\_draw (K=0.05, initial\_rating=1500.0, tie\_handling=draw)}
\item \texttt{elo\_tie\_correct\_draw\_only (K=0.05, initial\_rating=1500.0, tie\_handling=correct\_draw\_only)}
\item \texttt{glicko\_tie\_skip (initial\_rating=1500.0, initial\_rd=350.0, c=0.0, rd\_max=350.0, tie\_handling=skip, return\_deviation=false)}
\item \texttt{glicko\_tie\_draw (initial\_rating=1500.0, initial\_rd=350.0, c=0.0, rd\_max=350.0, tie\_handling=draw, return\_deviation=false)}
\item \texttt{glicko\_tie\_correct\_draw\_only (initial\_rating=1500.0, initial\_rd=350.0, c=0.0, rd\_max=350.0, tie\_handling=correct\_draw\_only, return\_deviation=false)}
\item \texttt{trueskill (mu\_initial=25.0, sigma\_initial=8.333333333333334, beta=4.166666666666667, tau=0.00333333333)}
\end{itemize}

\paragraph{Probabilistic comparisons.}
\begin{itemize}[leftmargin=*,noitemsep,topsep=0pt]
\item \texttt{bradley\_terry (return\_scores=false, max\_iter=500)}
\item \texttt{bradley\_terry\_map (prior=1.0, max\_iter=500)}
\item \texttt{bradley\_terry\_davidson (return\_scores=false, max\_iter=500)}
\item \texttt{bradley\_terry\_davidson\_map (prior=1.0, max\_iter=500)}
\item \texttt{rao\_kupper (tie\_strength=1.1, max\_iter=500)}
\item \texttt{rao\_kupper\_map (tie\_strength=1.1, prior=1.0, max\_iter=500)}
\item \texttt{thompson (n\_samples=10000, prior\_alpha=1.0, prior\_beta=1.0, seed=42)}
\item \texttt{bayesian\_mcmc (n\_samples=5000, burnin=1000, prior\_var=1.0, seed=42)}
\item \texttt{plackett\_luce (return\_scores=false, max\_iter=500, tol=1e-08)}
\item \texttt{plackett\_luce\_map (prior=1.0, max\_iter=500)}
\item \texttt{bradley\_terry\_luce (return\_scores=false, max\_iter=500)}
\item \texttt{bradley\_terry\_luce\_map (prior=1.0, max\_iter=500)}
\end{itemize}

\paragraph{Voting rules.}
\begin{itemize}[leftmargin=*,noitemsep,topsep=0pt]
\item \texttt{borda (return\_scores=false)}
\item \texttt{copeland (return\_scores=false)}
\item \texttt{win\_rate (return\_scores=false)}
\item \texttt{minimax\_variant\_margin\_tie\_ignore (variant=margin, tie\_policy=ignore)}
\item \texttt{minimax\_variant\_margin\_tie\_half (variant=margin, tie\_policy=half)}
\item \texttt{minimax\_\allowbreak variant\_\allowbreak winning\_\allowbreak votes\_\allowbreak tie\_\allowbreak ignore (variant=winning\_\allowbreak votes, tie\_\allowbreak policy=ignore)}
\item \texttt{minimax\_\allowbreak variant\_\allowbreak winning\_\allowbreak votes\_\allowbreak tie\_\allowbreak half (variant=winning\_\allowbreak votes, tie\_\allowbreak policy=half)}
\item \texttt{schulze\_tie\_ignore (tie\_policy=ignore)}
\item \texttt{schulze\_tie\_half (tie\_policy=half)}
\item \texttt{ranked\_\allowbreak pairs\_\allowbreak strength\_\allowbreak margin\_\allowbreak tie\_\allowbreak ignore (strength=margin, tie\_\allowbreak policy=ignore)}
\item \texttt{ranked\_\allowbreak pairs\_\allowbreak strength\_\allowbreak margin\_\allowbreak tie\_\allowbreak half (strength=margin, tie\_\allowbreak policy=half)}
\item \texttt{ranked\_\allowbreak pairs\_\allowbreak strength\_\allowbreak winning\_\allowbreak votes\_\allowbreak tie\_\allowbreak ignore (strength=winning\_\allowbreak votes, tie\_\allowbreak policy=ignore)}
\item \texttt{ranked\_\allowbreak pairs\_\allowbreak strength\_\allowbreak winning\_\allowbreak votes\_\allowbreak tie\_\allowbreak half (strength=winning\_\allowbreak votes, tie\_\allowbreak policy=half)}
\item \texttt{kemeny\_young\_tie\_ignore (tie\_policy=ignore, time\_limit=None)}
\item \texttt{kemeny\_young\_tie\_half (tie\_policy=half, time\_limit=None)}
\item \texttt{nanson\_rank\_ties\_average (rank\_ties=average)}
\item \texttt{nanson\_rank\_ties\_max (rank\_ties=max)}
\item \texttt{baldwin\_rank\_ties\_average (rank\_ties=average)}
\item \texttt{baldwin\_rank\_ties\_max (rank\_ties=max)}
\item \texttt{majority\_judgment (return\_scores=false)}
\end{itemize}

\paragraph{IRT.}
\begin{itemize}[leftmargin=*,noitemsep,topsep=0pt]
\item \texttt{rasch (return\_scores=false, max\_iter=500, return\_item\_params=false)}
\item \texttt{rasch\_map (prior=1.0, max\_iter=500, return\_item\_params=false)}
\item \texttt{rasch\_2pl (return\_scores=false, max\_iter=500, return\_item\_params=false)}
\item \texttt{rasch\_2pl\_map (prior=1.0, max\_iter=500, return\_item\_params=false)}
\item \texttt{rasch\_3pl (return\_scores=false, max\_iter=500, fix\_guessing=None, return\_item\_params=false)}
\item \texttt{rasch\_3pl\_map (prior=1.0, max\_iter=500, fix\_guessing=None, return\_item\_params=false)}
\item \texttt{rasch\_mml (return\_scores=false, max\_iter=100, em\_iter=20, n\_quadrature=21, return\_item\_params=false)}
\item \texttt{rasch\_mml\_credible (quantile=0.05, max\_iter=100, em\_iter=20, n\_quadrature=21)}
\item \texttt{dynamic\_irt\_linear (variant=linear, max\_iter=500, return\_item\_params=false)}
\item \texttt{dynamic\_irt\_growth (variant=growth, max\_iter=500, return\_item\_params=false)}
\end{itemize}

\paragraph{Graph/game.}
\begin{itemize}[leftmargin=*,noitemsep,topsep=0pt]
\item \texttt{pagerank (damping=0.85, max\_iter=100, tol=1e-12)}
\item \texttt{spectral (max\_iter=10000, tol=1e-12)}
\item \texttt{alpharank (alpha=1.0, population\_size=50, max\_iter=100000, tol=1e-12)}
\item \texttt{nash\_vs\_equilibrium (n\_iter=100, temperature=0.1, solver=lp, score\_type=vs\_equilibrium, return\_equilibrium=false)}
\item \texttt{nash\_\allowbreak advantage\_\allowbreak vs\_\allowbreak equilibrium (n\_\allowbreak iter=100, temperature=0.1, solver=lp, score\_\allowbreak type=advantage\_\allowbreak vs\_\allowbreak equilibrium, return\_\allowbreak equilibrium=false)}
\item \texttt{rank\_centrality\_tie\_ignore (tie\_handling=ignore, smoothing=0.0, teleport=0.0, max\_iter=10000, tol=1e-12)}
\item \texttt{rank\_centrality\_tie\_half (tie\_handling=half, smoothing=0.0, teleport=0.0, max\_iter=10000, tol=1e-12)}
\item \texttt{serial\_rank\_prob\_diff (comparison=prob\_diff)}
\item \texttt{serial\_rank\_sign (comparison=sign)}
\item \texttt{hodge\_rank\_binary\_total (pairwise\_stat=binary, weight\_method=total, return\_diagnostics=false)}
\item \texttt{hodge\_rank\_binary\_decisive (pairwise\_stat=binary, weight\_method=decisive, return\_diagnostics=false)}
\item \texttt{hodge\_rank\_binary\_uniform (pairwise\_stat=binary, weight\_method=uniform, return\_diagnostics=false)}
\item \texttt{hodge\_rank\_log\_odds\_total (pairwise\_stat=log\_odds, weight\_method=total, epsilon=0.5, return\_diagnostics=false)}
\item \texttt{hodge\_rank\_log\_odds\_decisive (pairwise\_stat=log\_odds, weight\_method=decisive, epsilon=0.5, return\_diagnostics=false)}
\item \texttt{hodge\_rank\_log\_odds\_uniform (pairwise\_stat=log\_odds, weight\_method=uniform, epsilon=0.5, return\_diagnostics=false)}
\end{itemize}

%% file: custom.bib
@article{kendall,
  title = {A New Measure of Rank Correlation},
  author = {Kendall, M. G.},
  journal = {Biometrika},
  year = {1938},
  volume = {30},
  number = {1-2},
  pages = {81--93},
  doi = {10.1093/biomet/30.1-2.81},
  url = {https://doi.org/10.1093/biomet/30.1-2.81},
  month = {jun}
}

@book{Arrow1951,
  title = {Social Choice and Individual Values},
  author = {Arrow, Kenneth J.},
  year = {1951},
  publisher = {John Wiley \& Sons},
  address = {New York}
}

@book{Condorcet1785,
  title = {Essai sur l'application de l'analyse `a la probabilit{\'e} des d{\'e}cisions rendues `a la pluralit{\'e} des voix},
  author = {Condorcet, Marie Jean Antoine Nicolas Caritat, Marquis de},
  year = {1785},
  publisher = {Imprimerie Royale},
  address = {Paris},
  url = {https://archive.org/details/bub_gb_RzAVAAAAQAAJ}
}

@inproceedings{Wang2022SelfConsistency,
  title = {Self-Consistency Improves Chain of Thought Reasoning in Language Models},
  author = {Xuezhi Wang and Jason Wei and Dale Schuurmans and Quoc V. Le and Ed H. Chi and Sharan Narang and Aakanksha Chowdhery and Denny Zhou},
  booktitle = {International Conference on Learning Representations},
  year = {2023},
  doi = {10.48550/arXiv.2203.11171},
  url = {https://openreview.net/forum?id=1PL1NIMMrw},
  eprint = {2203.11171},
  archivePrefix = {arXiv},
  primaryClass = {cs.CL}
}

@misc{Snell2024TestTimeCompute,
  title = {Scaling {LLM} Test-Time Compute Optimally can be More Effective than Scaling Model Parameters},
  author = {Charlie Snell and Jaehoon Lee and Kelvin Xu and Aviral Kumar},
  year = {2024},
  doi = {10.48550/arXiv.2408.03314},
  url = {https://arxiv.org/abs/2408.03314},
  eprint = {2408.03314},
  archivePrefix = {arXiv},
  primaryClass = {cs.LG}
}

@inproceedings{Zeng2025Revisiting,
  title = {Revisiting the Test-Time Scaling of o1-like Models: Do they Truly Possess Test-Time Scaling Capabilities?},
  author = {Zeng, Zhiyuan and Chen, Qingyuan and Yin, Zhangyue and Zhou, Yunhua and Qiu, Xipeng},
  booktitle = {Proceedings of the 63rd Annual Meeting of the Association for Computational Linguistics (Volume 1: Long Papers)},
  year = {2025},
  pages = {4651--4665},
  doi = {10.18653/v1/2025.acl-long.232},
  url = {https://aclanthology.org/2025.acl-long.232/}
}

@inproceedings{christiano2017deep,
  title = {Deep Reinforcement Learning from Human Preferences},
  author = {Paul F. Christiano and Jan Leike and Tom B. Brown and Miljan Martic and Shane Legg and Dario Amodei},
  booktitle = {Advances in Neural Information Processing Systems},
  year = {2017},
  volume = {30},
  pages = {4299--4307},
  doi = {10.5555/3294996.3295184},
  url = {https://papers.nips.cc/paper/7017-deep-reinforcement-learning-from-human-preferences}
}

@inproceedings{rafailov2023direct,
  title = {Direct Preference Optimization: Your Language Model is Secretly a Reward Model},
  author = {Rafael Rafailov and Archit Sharma and Eric Mitchell and Christopher D. Manning and Stefano Ermon and Chelsea Finn},
  booktitle = {Advances in Neural Information Processing Systems},
  year = {2023},
  volume = {36},
  pages = {53728--53741},
  doi = {10.5555/3666122.3668460},
  url = {https://papers.nips.cc/paper_files/paper/2023/hash/a85b405ed65c6477a4fe8302b5e06ce7-Abstract-Conference.html}
}

@inproceedings{chiang2024chatbotarena,
  title = {Chatbot Arena: An Open Platform for Evaluating {LLM}s by Human Preference},
  author = {Wei-Lin Chiang and Lianmin Zheng and Ying Sheng and Anastasios Nikolas Angelopoulos and Tianle Li and Dacheng Li and Banghua Zhu and Hao Zhang and Michael Jordan and Joseph E. Gonzalez and Ion Stoica},
  booktitle = {Proceedings of the 41st International Conference on Machine Learning},
  year = {2024},
  volume = {235},
  pages = {8359--8388},
  publisher = {PMLR},
  series = {Proceedings of Machine Learning Research},
  url = {https://proceedings.mlr.press/v235/chiang24b.html},
  eprint = {2403.04132},
  archivePrefix = {arXiv},
  primaryClass = {cs.AI}
}

@inproceedings{ameli2024statisticalframework,
  title = {A Statistical Framework for Ranking {LLM}-based Chatbots},
  author = {Siavash Ameli and Siyuan Zhuang and Ion Stoica and Michael W. Mahoney},
  booktitle = {International Conference on Learning Representations},
  year = {2025},
  doi = {10.48550/arXiv.2412.18407},
  url = {https://openreview.net/forum?id=rAoEub6Nw2},
  eprint = {2412.18407},
  archivePrefix = {arXiv},
  primaryClass = {stat.ML}
}

@misc{zhou2025lost,
  title = {Lost in Benchmarks? Rethinking Large Language Model Benchmarking with Item Response Theory},
  author = {Hongli Zhou and Hui Huang and Ziqing Zhao and Lvyuan Han and Huicheng Wang and Kehai Chen and Muyun Yang and Wei Bao and Jian Dong and Bing Xu and Conghui Zhu and Hailong Cao and Tiejun Zhao},
  year = {2025},
  doi = {10.48550/arXiv.2505.15055},
  url = {https://arxiv.org/abs/2505.15055},
  eprint = {2505.15055},
  archivePrefix = {arXiv},
  primaryClass = {cs.CL}
}

@inproceedings{gotou2020correctiondifficulty,
  title = {Taking the Correction Difficulty into Account in Grammatical Error Correction Evaluation},
  author = {Gotou, Takumi and Nagata, Ryo and Mita, Masato and Hanawa, Kazuaki},
  booktitle = {Proceedings of the 28th International Conference on Computational Linguistics},
  year = {2020},
  pages = {2085--2095},
  doi = {10.18653/v1/2020.coling-main.188},
  url = {https://aclanthology.org/2020.coling-main.188/}
}

@misc{abdin2025phi4reasoning,
  title = {{Phi-4-reasoning} Technical Report},
  author = {Marah Abdin and Sahaj Agarwal and Ahmed Awadallah and Vidhisha Balachandran and Harkirat Behl and Lingjiao Chen and Gustavo de Rosa and Suriya Gunasekar and Mojan Javaheripi and Neel Joshi and Piero Kauffmann and Yash Lara and Caio César Teodoro Mendes and Arindam Mitra and Besmira Nushi and Dimitris Papailiopoulos and Olli Saarikivi and Shital Shah and Vaishnavi Shrivastava and Vibhav Vineet and Yue Wu and Safoora Yousefi and Guoqing Zheng},
  year = {2025},
  doi = {10.48550/arXiv.2504.21318},
  url = {https://arxiv.org/abs/2504.21318},
  eprint = {2504.21318},
  archivePrefix = {arXiv}
}

@misc{bespoke_stratos,
  title = {Bespoke-Stratos: The Unreasonable Effectiveness of Reasoning Distillation},
  author = {{{Bespoke Labs}}},
  year = {2025},
  url = {https://www.bespokelabs.ai/blog/bespoke-stratos-the-unreasonable-effectiveness-of-reasoning-distillation},
  note = {Accessed: 2025-01-22}
}

@misc{BrUMO_2025,
  title = {Brown University Math Olympiad ({BrUMO})},
  author = {{{Brown University Math Olympiad Organizers}}},
  year = {2025},
  url = {https://www.brumo.org/tournament-info},
  note = {Official BrUMO website with tournament information (Apr 4--5, 2025); accessed 2025-09-25}
}

@misc{exaone-4.0,
  title = {{EXAONE 4.0}: Unified Large Language Models Integrating Non-reasoning and Reasoning Modes},
  author = {{{LG AI Research}}},
  year = {2025},
  doi = {10.48550/arXiv.2507.11407},
  url = {https://arxiv.org/abs/2507.11407},
  eprint = {2507.11407},
  archivePrefix = {arXiv}
}

@misc{guha2025openthoughtsdatarecipesreasoning,
  title = {{OpenThoughts: Data Recipes for Reasoning Models}},
  author = {Etash Guha and Ryan Marten and Sedrick Keh and Negin Raoof and Georgios Smyrnis and Hritik Bansal and Marianna Nezhurina and Jean Mercat and Trung Vu and Zayne Sprague and Ashima Suvarna and Benjamin Feuer and Liangyu Chen and Zaid Khan and Eric Frankel and Sachin Grover and Caroline Choi and Niklas Muennighoff and Shiye Su and Wanjia Zhao and John Yang and Shreyas Pimpalgaonkar and Kartik Sharma and Charlie Cheng-Jie Ji and Yichuan Deng and Sarah Pratt and Vivek Ramanujan and Jon Saad-Falcon and Jeffrey Li and Achal Dave and Alon Albalak and Kushal Arora and Blake Wulfe and Chinmay Hegde and Greg Durrett and Sewoong Oh and Mohit Bansal and Saadia Gabriel and Aditya Grover and Kai-Wei Chang and Vaishaal Shankar and Aaron Gokaslan and Mike A. Merrill and Tatsunori Hashimoto and Yejin Choi and Jenia Jitsev and Reinhard Heckel and Maheswaran Sathiamoorthy and Alexandros G. Dimakis and Ludwig Schmidt},
  year = {2025},
  doi = {10.48550/arXiv.2506.04178},
  url = {https://arxiv.org/abs/2506.04178},
  eprint = {2506.04178},
  archivePrefix = {arXiv},
  primaryClass = {cs.LG}
}

@article{guo2025deepseek,
  title = {{DeepSeek-R1} incentivizes reasoning in {LLMs} through reinforcement learning},
  author = {Guo, D. and Yang, D. and Zhang, H. and Song, J. and Wang, P. and Zhu, Q. and Xu, R. and Zhang, R. and Ma, S. and Bi, X. and Zhang, X. and Yu, X. and Wu, Y. and Wu, Z. F. and Gou, Z. and Shao, Z. and Li, Z. and Gao, Z. and Liu, A. and Xue, B. and Wang, B. and Wu, B. and Feng, B. and Lu, C. and Zhao, C. and Deng, C. and Ruan, C. and Dai, D. and Chen, D. and Ji, D. and Li, E. and Lin, F. and Dai, F. and Luo, F. and Hao, G. and Chen, G. and Li, G. and Zhang, H. and Xu, H. and Ding, H. and Gao, H. and Qu, H. and Li, H. and Guo, J. and Li, J. and Chen, J. and Yuan, J. and Tu, J. and Qiu, J. and Li, J. and Cai, J. L. and Ni, J. and Liang, J. and Chen, J. and Dong, K. and Hu, K. and You, K. and Gao, K. and Guan, K. and Huang, K. and Yu, K. and Wang, L. and Zhang, L. and Zhao, L. and Wang, L. and Zhang, L. and Xu, L. and Xia, L. and Zhang, M. and Zhang, M. and Tang, M. and Zhou, M. and Li, M. and Wang, M. and Li, M. and Tian, N. and Huang, P. and Zhang, P. and Wang, Q. and Chen, Q. and Du, Q. and Ge, R. and Zhang, R. and Pan, R. and Wang, R. and Chen, R. J. and Jin, R. L. and Chen, R. and Lu, S. and Zhou, S. and Chen, S. and Ye, S. and Wang, S. and Yu, S. and Zhou, S. and Pan, S. and Li, S. S. and Zhou, S. and Wu, S. and Yun, T. and Pei, T. and Sun, T. and Wang, T. and Zeng, W. and Liu, W. and Liang, W. and Gao, W. and Yu, W. and Zhang, W. and Xiao, W. L. and An, W. and Liu, X. and Wang, X. and Chen, X. and Nie, X. and Cheng, X. and Liu, X. and Xie, X. and Liu, X. and Yang, X. and Li, X. and Su, X. and Lin, X. and Li, X. Q. and Jin, X. and Shen, X. and Chen, X. and Sun, X. and Wang, X. and Song, X. and Zhou, X. and Wang, X. and Shan, X. and Li, Y. K. and Wang, Y. Q. and Wei, Y. X. and Zhang, Y. and Xu, Y. and Li, Y. and Zhao, Y. and Sun, Y. and Wang, Y. and Yu, Y. and Zhang, Y. and Shi, Y. and Xiong, Y. and He, Y. and Piao, Y. and Wang, Y. and Tan, Y. and Ma, Y. and Liu, Y. and Guo, Y. and Ou, Y. and Wang, Y. and Gong, Y. and Zou, Y. and He, Y. and Xiong, Y. and Luo, Y. and You, Y. and Liu, Y. and Zhou, Y. and Zhu, Y. X. and Huang, Y. and Li, Y. and Zheng, Y. and Zhu, Y. and Ma, Y. and Tang, Y. and Zha, Y. and Yan, Y. and Ren, Z. Z. and Ren, Z. and Sha, Z. and Fu, Z. and Xu, Z. and Xie, Z. and Zhang, Z. and Hao, Z. and Ma, Z. and Yan, Z. and Wu, Z. and Gu, Z. and Zhu, Z. and Liu, Z. and Li, Z. and Xie, Z. and Song, Z. and Pan, Z. and Huang, Z. and Xu, Z. and Zhang, Z. and Zhang, Z.},
  journal = {Nature},
  year = {2025},
  volume = {645},
  number = {8081},
  pages = {633--638},
  doi = {10.1038/s41586-025-09422-z},
  url = {https://doi.org/10.1038/s41586-025-09422-z}
}

@misc{HMMT_Feb2025,
  title = {HMMT February 2025 Archive (Problems and Solutions)},
  author = {{{Harvard--MIT Mathematics Tournament}}},
  year = {2025},
  url = {https://www.hmmt.org/www/archive/282},
  note = {Official HMMT archive page for February 2025 competition; accessed 2025-09-25}
}

@inproceedings{kwon2023efficient,
  title = {Efficient Memory Management for Large Language Model Serving with {PagedAttention}},
  author = {Kwon, Woosuk and Li, Zhuohan and Zhuang, Siyuan and Sheng, Ying and Zheng, Lianmin and Yu, Cody Hao and Gonzalez, Joseph and Zhang, Hao and Stoica, Ion},
  booktitle = {Proceedings of the 29th Symposium on Operating Systems Principles},
  year = {2023},
  pages = {611--626},
  publisher = {ACM},
  doi = {10.1145/3600006.3613165},
  url = {https://arxiv.org/abs/2309.06180},
  eprint = {2309.06180},
  archivePrefix = {arXiv}
}

@inproceedings{liu2025acereason,
  title = {AceReason-Nemotron 1.1: Advancing Math and Code Reasoning through {SFT} and {RL} Synergy},
  author = {Zihan Liu and Zhuolin Yang and Yang Chen and Chankyu Lee and Mohammad Shoeybi and Bryan Catanzaro and Wei Ping},
  booktitle = {International Conference on Learning Representations},
  year = {2026},
  doi = {10.48550/arXiv.2506.13284},
  url = {https://openreview.net/forum?id=IaEqjWXd1d},
  eprint = {2506.13284},
  archivePrefix = {arXiv},
  primaryClass = {cs.CL}
}

@misc{MAA_AIME2024,
  title = {American Invitational Mathematics Examination ({AIME})},
  author = {{{Mathematical Association of America}}},
  year = {2024},
  url = {https://maa.org/maa-invitational-competitions/},
  note = {Official MAA page for the AIME competition (covers AIME 2024); accessed 2025-09-25}
}

@misc{MAA_AIME2025,
  title = {American Invitational Mathematics Examination ({AIME})},
  author = {{{Mathematical Association of America}}},
  year = {2025},
  url = {https://maa.org/maa-invitational-competitions/},
  note = {Official MAA page for the AIME competition (covers AIME 2025); accessed 2025-09-25}
}

@misc{nvidia2025nemotronnano2,
  title = {{NVIDIA} Nemotron Nano 2: An Accurate and Efficient Hybrid Mamba-Transformer Reasoning Model},
  author = {{{NVIDIA}}},
  year = {2025},
  doi = {10.48550/arXiv.2508.14444},
  url = {https://arxiv.org/abs/2508.14444},
  eprint = {2508.14444},
  archivePrefix = {arXiv},
  primaryClass = {cs.CL}
}

@misc{openai2025gptoss120bgptoss20bmodel,
  title = {{{gpt-oss-120b} \& {gpt-oss-20b} Model Card}},
  author = {{{OpenAI}}},
  year = {2025},
  doi = {10.48550/arXiv.2508.10925},
  url = {https://arxiv.org/abs/2508.10925},
  eprint = {2508.10925},
  archivePrefix = {arXiv},
  primaryClass = {cs.CL}
}

@misc{openr1,
  title = {{Open-R1}: A Fully Open Reproduction of {DeepSeek-R1}},
  author = {{{Hugging Face}}},
  year = {2025},
  url = {https://github.com/huggingface/open-r1},
  month = {jan}
}

@misc{qwen3technicalreport,
  title = {{Qwen3} Technical Report},
  author = {{{Qwen Team}}},
  year = {2025},
  doi = {10.48550/arXiv.2505.09388},
  url = {https://arxiv.org/abs/2505.09388},
  eprint = {2505.09388},
  archivePrefix = {arXiv},
  primaryClass = {cs.CL}
}

@misc{reduce_overthinking_2025,
  title = {Think Less, Achieve More: Cut Reasoning Costs by 50\% Without Sacrificing Accuracy},
  author = {{{NovaSky Team}}},
  year = {2025},
  url = {https://novasky-ai.github.io/posts/reduce-overthinking/},
  note = {Accessed: 2025-01-23}
}

@inproceedings{wen2025lightr1curriculumsftdpo,
  title = {Light-R1: Curriculum {SFT}, {DPO} and {RL} for Long {COT} from Scratch and Beyond},
  author = {Liang Wen and Yunke Cai and Fenrui Xiao and Xin He and Qi An and Zhenyu Duan and Yimin Du and Junchen Liu and Lifu Tang and Xiaowei Lv and Haosheng Zou and Yongchao Deng and Shousheng Jia and Xiangzheng Zhang},
  booktitle = {Proceedings of the 63rd Annual Meeting of the Association for Computational Linguistics (Volume 6: Industry Track)},
  year = {2025},
  pages = {318--327},
  publisher = {Association for Computational Linguistics},
  doi = {10.18653/v1/2025.acl-industry.24},
  url = {https://aclanthology.org/2025.acl-industry.24/}
}

@inproceedings{ye2025limoreasoning,
  title = {{LIMO}: Less is More for Reasoning},
  author = {Yixin Ye and Zhen Huang and Yang Xiao and Ethan Chern and Shijie Xia and Pengfei Liu},
  booktitle = {Second Conference on Language Modeling},
  year = {2025},
  doi = {10.48550/arXiv.2502.03387},
  url = {https://openreview.net/forum?id=T2TZ0RY4Zk},
  eprint = {2502.03387},
  archivePrefix = {arXiv},
  primaryClass = {cs.CL}
}

@misc{chen2021evaluating,
  title = {Evaluating Large Language Models Trained on Code},
  author = {Mark Chen and Jerry Tworek and Heewoo Jun and Qiming Yuan and Henrique Ponde de Oliveira Pinto and Jared Kaplan and Harri Edwards and Yuri Burda and Nicholas Joseph and Greg Brockman and Alex Ray and Raul Puri and Gretchen Krueger and Michael Petrov and Heidy Khlaaf and Girish Sastry and Pamela Mishkin and Brooke Chan and Scott Gray and Nick Ryder and Mikhail Pavlov and Alethea Power and Lukasz Kaiser and Mohammad Bavarian and Clemens Winter and Philippe Tillet and Felipe Petroski Such and Dave Cummings and Matthias Plappert and Fotios Chantzis and Elizabeth Barnes and Ariel Herbert-Voss and William Hebgen Guss and Alex Nichol and Alex Paino and Nikolas Tezak and Jie Tang and Igor Babuschkin and Suchir Balaji and Shantanu Jain and William Saunders and Christopher Hesse and Andrew N. Carr and Jan Leike and Josh Achiam and Vedant Misra and Evan Morikawa and Alec Radford and Matthew Knight and Miles Brundage and Mira Murati and Katie Mayer and Peter Welinder and Bob McGrew and Dario Amodei and Sam McCandlish and Ilya Sutskever and Wojciech Zaremba},
  year = {2021},
  doi = {10.48550/arXiv.2107.03374},
  url = {https://arxiv.org/abs/2107.03374},
  eprint = {2107.03374},
  archivePrefix = {arXiv},
  primaryClass = {cs.LG}
}

@inproceedings{yao2024taubench,
  title = {{$\tau$}-bench: A Benchmark for Tool-Agent-User Interaction in Real-World Domains},
  author = {Shunyu Yao and Noah Shinn and Pedram Razavi and Karthik Narasimhan},
  booktitle = {International Conference on Learning Representations},
  year = {2025},
  doi = {10.48550/arXiv.2406.12045},
  url = {https://openreview.net/forum?id=roNSXZpUDN},
  eprint = {2406.12045},
  archivePrefix = {arXiv},
  primaryClass = {cs.AI}
}

@inproceedings{liu2024stable_reasoning,
  title = {Are Your {LLM}s Capable of Stable Reasoning?},
  author = {Junnan Liu and Hongwei Liu and Linchen Xiao and Ziyi Wang and Kuikun Liu and Songyang Gao and Wenwei Zhang and Songyang Zhang and Kai Chen},
  booktitle = {Findings of the Association for Computational Linguistics: ACL 2025},
  year = {2025},
  pages = {17594--17632},
  publisher = {Association for Computational Linguistics},
  doi = {10.18653/v1/2025.findings-acl.905},
  url = {https://aclanthology.org/2025.findings-acl.905/}
}

@inproceedings{hariri2025dontpasskbayesianframework,
  title = {Don't Pass@k: A Bayesian Framework for Large Language Model Evaluation},
  author = {Hariri, Mohsen and Samandar, Amirhossein and Hinczewski, Michael and Chaudhary, Vipin},
  booktitle = {Proceedings of the 14th International Conference on Learning Representations (ICLR 2026)},
  year = {2026},
  url = {https://openreview.net/forum?id=PTXi3Ef4sT},
  eprint = {2510.04265},
  archivePrefix = {arXiv}
}

@article{Thompson1933,
  title = {On the Likelihood that One Unknown Probability Exceeds Another in View of the Evidence of Two Samples},
  author = {Thompson, William R.},
  journal = {Biometrika},
  year = {1933},
  volume = {25},
  number = {3-4},
  pages = {285--294},
  doi = {10.1093/biomet/25.3-4.285},
  url = {https://doi.org/10.1093/biomet/25.3-4.285}
}

@article{Russo2018TS,
  title = {A Tutorial on Thompson Sampling},
  author = {Russo, Daniel J. and Van Roy, Benjamin and Kazerouni, Abbas and Osband, Ian and Wen, Zheng},
  journal = {Foundations and Trends in Machine Learning},
  year = {2018},
  volume = {11},
  number = {1},
  pages = {1--96},
  doi = {10.1561/2200000070},
  url = {https://doi.org/10.1561/2200000070}
}

@book{Gelman2013BDA,
  title = {Bayesian Data Analysis},
  author = {Gelman, Andrew and Carlin, John B. and Stern, Hal S. and Dunson, David B. and Vehtari, Aki and Rubin, Donald B.},
  year = {2013},
  publisher = {CRC Press},
  edition = {3},
  doi = {10.1201/b16018},
  url = {https://doi.org/10.1201/b16018}
}

@article{BradleyTerry1952,
  title = {Rank Analysis of Incomplete Block Designs: The Method of Paired Comparisons},
  author = {Bradley, Ralph Allan and Terry, Milton E.},
  journal = {Biometrika},
  year = {1952},
  volume = {39},
  number = {3-4},
  pages = {324--345},
  doi = {10.1093/biomet/39.3-4.324},
  url = {https://doi.org/10.1093/biomet/39.3-4.324}
}

@article{Metropolis1953,
  title = {Equation of State Calculations by Fast Computing Machines},
  author = {Metropolis, Nicholas and Rosenbluth, Arianna W. and Rosenbluth, Marshall N. and Teller, Augusta H. and Teller, Edward},
  journal = {The Journal of Chemical Physics},
  year = {1953},
  volume = {21},
  number = {6},
  pages = {1087--1092},
  doi = {10.1063/1.1699114},
  url = {https://doi.org/10.1063/1.1699114}
}

@article{Hastings1970,
  title = {Monte Carlo Sampling Methods Using Markov Chains and Their Applications},
  author = {Hastings, W. K.},
  journal = {Biometrika},
  year = {1970},
  volume = {57},
  number = {1},
  pages = {97--109},
  doi = {10.1093/biomet/57.1.97},
  url = {https://doi.org/10.1093/biomet/57.1.97}
}

@article{CaronDoucet2012,
  title = {Efficient Bayesian Inference for Generalized Bradley--Terry Models},
  author = {Caron, Fran{\c{c}}ois and Doucet, Arnaud},
  journal = {Journal of Computational and Graphical Statistics},
  year = {2012},
  volume = {21},
  number = {1},
  pages = {174--196},
  doi = {10.1080/10618600.2012.638220},
  url = {https://doi.org/10.1080/10618600.2012.638220}
}

@misc{Borda1781,
  title = {M{\'e}moire sur les {\'e}lections au scrutin},
  author = {de Borda, Jean-Charles},
  year = {1781},
  pages = {657--665},
  url = {https://webusers.imj-prg.fr/~alexandre.guilbaud/LX2U1/Borda_Memoire_sur_les_elections_au_scrutin_MARS_1781_extrait.pdf},
  note = {Often cited as appearing in the 1781 volume (issued in 1784) of the Histoire/M{\'e}moires of the Acad{\'e}mie.},
  howpublished = {Histoire de l'Acad{\'e}mie Royale des Sciences, Paris}
}

@book{brandt2016compsocchoice,
  title = {Handbook of Computational Social Choice},
  editor = {Brandt, Felix and Conitzer, Vincent and Endriss, Ulle and Lang, J{\'e}r{\^o}me and Procaccia, Ariel D.},
  year = {2016},
  publisher = {Cambridge University Press},
  doi = {10.1017/CBO9781107446984},
  url = {https://doi.org/10.1017/CBO9781107446984},
  isbn = {9781107446984}
}

@misc{Copeland1951,
  title = {A Reasonable Social Welfare Function},
  author = {Copeland, Arthur H.},
  year = {1951},
  url = {https://bibbase.org/network/publication/copeland-areasonablesocialwelfarefunction-1951},
  note = {University of Michigan, Ann Arbor. Mimeographed notes.},
  howpublished = {Seminar on Applications of Mathematics to Social Sciences}
}

@article{Schulze2010,
  title = {A new monotonic, clone-independent, reversal symmetric, and {C}ondorcet-consistent single-winner election method},
  author = {Schulze, Markus},
  journal = {Social Choice and Welfare},
  year = {2011},
  volume = {36},
  number = {2},
  pages = {267--303},
  doi = {10.1007/s00355-010-0475-4},
  url = {https://doi.org/10.1007/s00355-010-0475-4}
}

@article{Tideman1987,
  title = {Independence of clones as a criterion for voting rules},
  author = {Tideman, T. N.},
  journal = {Social Choice and Welfare},
  year = {1987},
  volume = {4},
  number = {3},
  pages = {185--206},
  doi = {10.1007/BF00433944},
  url = {https://doi.org/10.1007/BF00433944}
}

@article{Young1977,
  title = {Extending {C}ondorcet's rule},
  author = {Young, H. P.},
  journal = {Journal of Economic Theory},
  year = {1977},
  volume = {16},
  number = {2},
  pages = {335--353},
  doi = {10.1016/0022-0531(77)90012-6},
  url = {https://doi.org/10.1016/0022-0531(77)90012-6}
}

@book{BalinskiLaraki2011,
  title = {Majority Judgment: Measuring, Ranking, and Electing},
  author = {Balinski, Michel and Laraki, Rida},
  year = {2011},
  publisher = {The MIT Press},
  doi = {10.7551/mitpress/9780262015134.001.0001},
  url = {https://doi.org/10.7551/mitpress/9780262015134.001.0001},
  isbn = {9780262295604}
}

@article{davidson1970bties,
  title = {On Extending the Bradley--Terry Model to Accommodate Ties in Paired Comparison Experiments},
  author = {Davidson, Roger R.},
  journal = {Journal of the American Statistical Association},
  year = {1970},
  volume = {65},
  number = {329},
  pages = {317--328},
  doi = {10.1080/01621459.1970.10481082},
  url = {https://doi.org/10.1080/01621459.1970.10481082}
}

@article{rao1967ties,
  title = {Ties in Paired-Comparison Experiments: A Generalization of the Bradley--Terry Model},
  author = {Rao, P. V. and Kupper, L. L.},
  journal = {Journal of the American Statistical Association},
  year = {1967},
  volume = {62},
  number = {317},
  pages = {194--204},
  doi = {10.1080/01621459.1967.10482901},
  url = {https://doi.org/10.1080/01621459.1967.10482901}
}

@book{Elo1978,
  title = {The Rating of Chessplayers, Past and Present},
  author = {Elo, Arpad E.},
  year = {1978},
  publisher = {Arco Publishing},
  url = {https://archive.org/details/ratingofchesspla0000eloa},
  isbn = {0668047216}
}

@article{Glickman1999,
  title = {Parameter Estimation in Large Dynamic Paired Comparison Experiments},
  author = {Glickman, Mark E.},
  journal = {Journal of the Royal Statistical Society: Series C (Applied Statistics)},
  year = {1999},
  volume = {48},
  number = {3},
  pages = {377--394},
  doi = {10.1111/1467-9876.00159},
  url = {https://doi.org/10.1111/1467-9876.00159}
}

@inproceedings{herbrich2006trueskill,
  title = {{TrueSkill}: A Bayesian Skill Rating System},
  author = {Ralf Herbrich and Tom Minka and Thore Graepel},
  booktitle = {Advances in Neural Information Processing Systems},
  year = {2006},
  volume = {19},
  pages = {569--576},
  publisher = {MIT Press},
  url = {https://papers.neurips.cc/paper/3079-trueskilltm-a-bayesian-skill-rating-system.pdf}
}

@article{plackett1975permutations,
  title = {The Analysis of Permutations},
  author = {Plackett, R. L.},
  journal = {Applied Statistics},
  year = {1975},
  volume = {24},
  number = {2},
  pages = {193--202},
  doi = {10.2307/2346567},
  url = {https://doi.org/10.2307/2346567}
}

@book{luce1959choice,
  title = {Individual Choice Behavior: A Theoretical Analysis},
  author = {Luce, R. Duncan},
  year = {1959},
  publisher = {John Wiley \& Sons},
  url = {https://archive.org/details/individualchoice0000luce}
}

@article{hunter2004mmbradleyterry,
  title = {{MM} Algorithms for Generalized Bradley--Terry Models},
  author = {Hunter, David R.},
  journal = {The Annals of Statistics},
  year = {2004},
  volume = {32},
  number = {1},
  pages = {384--406},
  doi = {10.1214/aos/1079120141},
  url = {https://doi.org/10.1214/aos/1079120141}
}

@misc{firth2019davidsonluce,
  title = {Davidson--Luce Model for Multi-item Choice with Ties},
  author = {David Firth and Ioannis Kosmidis and Heather Turner},
  year = {2019},
  doi = {10.48550/arXiv.1909.07123},
  url = {https://arxiv.org/abs/1909.07123},
  eprint = {1909.07123},
  archivePrefix = {arXiv},
  primaryClass = {stat.ME}
}

@book{mccullaghnelder1989glm,
  title = {Generalized Linear Models},
  author = {McCullagh, P. and Nelder, J. A.},
  year = {1989},
  publisher = {Springer},
  doi = {10.1007/978-1-4899-3242-6},
  url = {https://doi.org/10.1007/978-1-4899-3242-6},
  isbn = {9781489932426}
}

@book{deboeckwilson2004explanatory,
  title = {Explanatory Item Response Models},
  editor = {De Boeck, Paul and Wilson, Mark},
  year = {2004},
  publisher = {Springer},
  doi = {10.1007/978-1-4757-3990-9},
  url = {https://doi.org/10.1007/978-1-4757-3990-9},
  isbn = {9781475739909}
}

@book{rasch1960rasch,
  title = {Probabilistic Models for Some Intelligence and Attainment Tests},
  author = {Rasch, Georg},
  year = {1960},
  publisher = {Danish Institute for Educational Research},
  address = {Copenhagen},
  url = {https://archive.org/details/probabilisticmod0000rasc}
}

@incollection{birnbaum1968latent,
  title = {Some Latent Trait Models and Their Use in Inferring an Examinee's Ability},
  author = {Birnbaum, Allan},
  editor = {Lord, Frederic M. and Novick, Melvin R.},
  booktitle = {Statistical Theories of Mental Test Scores},
  year = {1968},
  pages = {396--479},
  publisher = {Addison-Wesley},
  address = {Reading, MA},
  url = {https://faculty.ucmerced.edu/jvevea/classes/290_21/readings/week%209/Birnbaum.pdf}
}

@article{Mislevy1986,
  title = {Bayes Modal Estimation in Item Response Models},
  author = {Mislevy, Robert J.},
  journal = {Psychometrika},
  year = {1986},
  volume = {51},
  number = {2},
  pages = {177--195},
  doi = {10.1007/BF02293979},
  url = {https://doi.org/10.1007/BF02293979}
}

@article{BockAitkin1981,
  title = {Marginal Maximum Likelihood Estimation of Item Parameters: Application of an {EM} Algorithm},
  author = {Bock, R. Darrell and Aitkin, Murray},
  journal = {Psychometrika},
  year = {1981},
  volume = {46},
  number = {4},
  pages = {443--459},
  doi = {10.1007/BF02293801},
  url = {https://doi.org/10.1007/BF02293801}
}

@article{ChenHouDodd1998,
  title = {A Comparison of Maximum Likelihood Estimation and Expected a Posteriori Estimation in {CAT} Using the Partial Credit Model},
  author = {Chen, Ssu-Kuang and Hou, Liling and Dodd, Barbara G.},
  journal = {Educational and Psychological Measurement},
  year = {1998},
  volume = {58},
  number = {4},
  pages = {569--595},
  doi = {10.1177/0013164498058004002},
  url = {https://doi.org/10.1177/0013164498058004002}
}

@article{verhelst1993dynamicrasch,
  title = {A Dynamic Generalization of the Rasch Model},
  author = {Verhelst, Norman D. and Glas, Cees A. W.},
  journal = {Psychometrika},
  year = {1993},
  volume = {58},
  number = {3},
  pages = {395--415},
  doi = {10.1007/BF02294648},
  url = {https://doi.org/10.1007/BF02294648}
}

@article{wang2019longitudinalirt,
  title = {On Longitudinal Item Response Theory Models: A Didactic},
  author = {Wang, Chun and Nydick, Steven W.},
  journal = {Journal of Educational and Behavioral Statistics},
  year = {2020},
  volume = {45},
  number = {3},
  pages = {339--368},
  doi = {10.3102/1076998619882026},
  url = {https://doi.org/10.3102/1076998619882026}
}

@techreport{page1999pagerank,
  title = {The PageRank Citation Ranking: Bringing Order to the Web},
  author = {Lawrence Page and Sergey Brin and Rajeev Motwani and Terry Winograd},
  year = {1999},
  number = {1999-66},
  institution = {Stanford InfoLab},
  url = {https://ilpubs.stanford.edu/422/1/1999-66.pdf},
  month = {nov},
  note = {Previous number: {SIDL-WP-1999-0120}}
}

@article{negahban2017rankcentrality,
  title = {Rank Centrality: Ranking from Pairwise Comparisons},
  author = {Negahban, Sahand and Oh, Sewoong and Shah, Devavrat},
  journal = {Operations Research},
  year = {2017},
  volume = {65},
  number = {1},
  pages = {266--287},
  doi = {10.1287/opre.2016.1534},
  url = {https://doi.org/10.1287/opre.2016.1534}
}

@article{omidshafiei2019alpharank,
  title = {{$\alpha$}-Rank: Multi-Agent Evaluation by Evolution},
  author = {Omidshafiei, Shayegan and Papadimitriou, Christos and Piliouras, Georgios and Tuyls, Karl and Rowland, Mark and Lespiau, Jean-Baptiste and Czarnecki, Wojciech M. and P{\'e}rolat, Julien and Munos, R{\'e}mi},
  journal = {Scientific Reports},
  year = {2019},
  volume = {9},
  number = {1},
  pages = {9937},
  doi = {10.1038/s41598-019-45619-9},
  url = {https://doi.org/10.1038/s41598-019-45619-9}
}

@inproceedings{balduzzi2019openended,
  title = {Open-ended Learning in Symmetric Zero-sum Games},
  author = {Balduzzi, David and Garnelo, Marta and Bachrach, Yoram and Czarnecki, Wojciech and P{\'e}rolat, Julien and Jaderberg, Max and Graepel, Thore},
  booktitle = {Proceedings of the 36th International Conference on Machine Learning},
  year = {2019},
  volume = {97},
  pages = {434--443},
  publisher = {PMLR},
  series = {Proceedings of Machine Learning Research},
  url = {https://proceedings.mlr.press/v97/balduzzi19a.html}
}

@article{fogel2016serialrank,
  title = {Spectral Ranking using Seriation},
  author = {Fogel, Fajwel and d'Aspremont, Alexandre and Vojnovic, Milan},
  journal = {Journal of Machine Learning Research},
  year = {2016},
  volume = {17},
  pages = {88:1--88:45},
  url = {https://jmlr.org/papers/v17/16-035.html}
}

@article{jiang2009hodgerank,
  title = {Statistical Ranking and Combinatorial Hodge Theory},
  author = {Xiaoye Jiang and Lek-Heng Lim and Yuan Yao and Yinyu Ye},
  journal = {Mathematical Programming},
  year = {2011},
  volume = {127},
  number = {1},
  pages = {203--244},
  doi = {10.1007/s10107-010-0419-x},
  url = {https://doi.org/10.1007/s10107-010-0419-x}
}

@article{Kemeny1959,
  title = {Mathematics without Numbers},
  author = {Kemeny, John G.},
  journal = {Daedalus},
  year = {1959},
  volume = {88},
  number = {4},
  pages = {577--591},
  url = {https://www.jstor.org/stable/20026581}
}

@article{Nanson1883,
  title = {Methods of Election},
  author = {Nanson, E. J.},
  journal = {Transactions and Proceedings of the Royal Society of Victoria},
  year = {1883},
  volume = {19},
  pages = {197--240},
  url = {https://www.biodiversitylibrary.org/itemdetails/106382},
  note = {Often cited as 1882 in secondary sources.}
}

@article{Baldwin1926,
  title = {The technique of the Nanson preferential majority system of election},
  author = {Baldwin, J. M.},
  journal = {Proceedings of the Royal Society of Victoria, New Series},
  year = {1926},
  volume = {39},
  number = {1},
  pages = {42--52},
  url = {https://www.biodiversitylibrary.org/itemdetails/236582}
}

@inproceedings{zhang2025dfloat11,
  title = {70\% Size, 100\% Accuracy: Lossless LLM Compression for Efficient GPU Inference via Dynamic-Length Float ({DFloat11})},
  author = {Zhang, Tianyi and Hariri, Mohsen and Zhong, Shaochen and Chaudhary, Vipin and Sui, Yang and Hu, Xia and Shrivastava, Anshumali},
  booktitle = {Advances in Neural Information Processing Systems},
  year = {2025},
  url = {https://arxiv.org/abs/2504.11651},
  eprint = {2504.11651},
  archivePrefix = {arXiv}
}

@misc{openreasoning_nemotron_15b,
  title = {{OpenReasoning-Nemotron-1.5B}},
  author = {{{NVIDIA}}},
  year = {2025},
  url = {https://huggingface.co/nvidia/OpenReasoning-Nemotron-1.5B},
  note = {Model card; accessed 2026-03-09}
}

@misc{fuseo1_flash_preview,
  title = {{FuseO1-DeepSeekR1-QwQ-SkyT1-Flash-32B-Preview}},
  author = {{{FuseAI}}},
  year = {2025},
  url = {https://huggingface.co/FuseAI/FuseO1-DeepSeekR1-QwQ-SkyT1-Flash-32B-Preview},
  note = {Model card; accessed 2026-03-09}
}

@article{wolf2019huggingface,
  author = {Wolf, Thomas and Debut, Lysandre and Sanh, Victor and Chaumond, Julien and Delangue, Clement and Moi, Anthony and Cistac, Pierric and Rault, Tim and Louf, R{\'e}mi and Funtowicz, Morgan and others},
  title = {Huggingface's transformers: State-of-the-art natural language processing},
  journal = {arXiv preprint arXiv:1910.03771},
  year = {2019},
  url = {https://arxiv.org/abs/1910.03771}
}

@misc{accelerate,
  author = {Sylvain Gugger and Lysandre Debut and Thomas Wolf and Philipp Schmid and Zachary Mueller and Sourab Mangrulkar and Marc Sun and Benjamin Bossan},
  title = {Accelerate: Training and inference at scale made simple, efficient and adaptable.},
  year = {2022},
  url = {https://github.com/huggingface/accelerate}
}

@article{Dao2022FlashAttention,
  author = {Dao, Tri},
  title = {Flashattention-2: Faster attention with better parallelism and work partitioning},
  journal = {arXiv preprint arXiv:2307.08691},
  year = {2023},
  url = {https://arxiv.org/abs/2307.08691}
}
